\documentclass[acmtog,nonacm]{acmart}
\acmJournal{TOG}
\title{Raster2Seq: Polygon Sequence Generation for Floorplan Reconstruction}

\author{Hao Phung}
\affiliation{%
  \institution{Cornell University}
  \city{New York}
  \state{NY}
  \country{USA}
}
\email{htp26@cornell.edu}
\orcid{0000-0002-6834-4139}

\author{Hadar Averbuch-Elor}
\affiliation{%
  \institution{Cornell University}
  \city{New York}
  \state{NY}
  \country{USA}
}
\email{hadarelor@cornell.edu}
\orcid{0000-0003-3476-0940}

\usepackage{makecell}
\usepackage{tabularx}
\usepackage{multicol}
\usepackage{multirow}
\usepackage{amsmath}
\usepackage{cleveref}
\usepackage[utf8]{inputenc} %
\usepackage[T1]{fontenc}    %
\usepackage{url}            %
\usepackage{booktabs}       %
\usepackage{amsfonts}       %
\usepackage{nicefrac}       %
\usepackage{microtype}      %
\usepackage{xcolor}         %
\usepackage[table]{xcolor}
\usepackage[dvipsnames]{xcolor}
\usepackage[linesnumbered,ruled,vlined]{algorithm2e}
\usepackage{caption}
\usepackage{subcaption}
\usepackage{tikz}
\usepackage{wrapfig}
\usepackage{verbatim}
\usepackage{fancyvrb}    %

\usepackage{listings}

\lstdefinestyle{promptstyle}{
  basicstyle=\ttfamily\tiny,
  breaklines=true,
  breakatwhitespace=true,
  breakautoindent=false,
  breakindent=0pt,
  postbreak=,
  columns=fullflexible,
  keepspaces=true,
  frame=none,
  xleftmargin=0pt,
  xrightmargin=0pt,
  aboveskip=4pt,
  belowskip=4pt,
}

\newcommand{\change}[1]{{#1}}
\newcommand{\methodname}{\emph{Raster2Seq}\xspace}

\newcommand{\cubi}{CubiCasa5K\xspace}

\newcommand{\ourmodel}{Raster2Seq\xspace}

\usepackage{mwe}

\newcommand{\xcheck}{\checkmark\kern-1.1ex\raisebox{.7ex}{\rotatebox[origin=c]{125}{--}}}
\newcommand{\minihead}[1]{\medskip \noindent \textbf{#1}}

\newcommand{\whitetxt}[1]{{\color{white}#1}\normalfont}

\definecolor{s3d_livingroom}{RGB}{141,211,199}
\definecolor{s3d_kitchen}{RGB}{255,255,179}
\definecolor{s3d_bedroom}{RGB}{190,186,218}
\definecolor{s3d_bathroom}{RGB}{251,128,114}
\definecolor{s3d_balcony}{RGB}{128,177,211}
\definecolor{s3d_corridor}{RGB}{253,180,98}
\definecolor{s3d_diningroom}{RGB}{179,222,105}
\definecolor{s3d_study}{RGB}{252,205,229}
\definecolor{s3d_studio}{RGB}{217,217,217}
\definecolor{s3d_storeroom}{RGB}{188,128,189}
\definecolor{s3d_garden}{RGB}{204,235,197}
\definecolor{s3d_laundryroom}{RGB}{255,237,111}
\definecolor{s3d_office}{RGB}{27,158,119}
\definecolor{s3d_basement}{RGB}{217,95,2}
\definecolor{s3d_garage}{RGB}{117,112,179}
\definecolor{s3d_misc}{RGB}{231,41,138}

\definecolor{outdoor}{RGB}{141, 211, 199}
\definecolor{bedroom}{RGB}{251, 128, 114}
\definecolor{entry}{RGB}{253, 180, 98}
\definecolor{bath}{RGB}{128, 177, 211}
\definecolor{kitchen}{RGB}{255,255,179}
\definecolor{livingroom}{RGB}{190,186,218}
\definecolor{storage}{RGB}{179,222,105}
\definecolor{garage}{RGB}{252,205,229}
\definecolor{undefined}{RGB}{217,217,217}

\definecolor{r2g_unknown}{RGB}{141,211,199}
\definecolor{r2g_living_room}{RGB}{255,255,179}
\definecolor{r2g_kitchen}{RGB}{190,186,218}
\definecolor{r2g_bedroom}{RGB}{251,128,114}
\definecolor{r2g_bathroom}{RGB}{128,177,211}
\definecolor{r2g_restroom}{RGB}{253,180,98}
\definecolor{r2g_balcony}{RGB}{179,222,105}
\definecolor{r2g_closet}{RGB}{252,205,229}
\definecolor{r2g_corridor}{RGB}{217,217,217}
\definecolor{r2g_washing_room}{RGB}{188,128,189}
\definecolor{r2g_PS}{RGB}{204,235,197}
\definecolor{r2g_outside}{RGB}{255,237,111}

\newbox\jsavebox
\newcommand{\jsubfig}[2]{%
	\sbox\jsavebox{#1}%
	\parbox[t]{\wd\jsavebox}{\centering\usebox\jsavebox\\#2}%
	}

\crefname{equation}{Eq.}{Eqs.}
\crefname{figure}{Fig.}{Figs.}
\crefname{table}{Tab.}{Tabs.}

\Crefname{equation}{Equation}{Equations}
\Crefname{figure}{Figure}{Figures}
\Crefname{table}{Table}{Tables}

\copyrightyear{2026}
\acmYear{2026}
\setcopyright{cc}
\setcctype{by}
\acmConference[SIGGRAPH Conference Papers '26]{Special Interest Group on Computer Graphics and Interactive Techniques Conference Conference Papers}{July 19--23, 2026}{Los Angeles, CA, USA}
\acmBooktitle{Special Interest Group on Computer Graphics and Interactive Techniques Conference Conference Papers (SIGGRAPH Conference Papers '26), July 19--23, 2026, Los Angeles, CA, USA}
\acmDOI{10.1145/3799902.3811124}
\acmISBN{979-8-4007-2554-8/2026/07}

\ccsdesc[500]{Computing methodologies~Scene understanding}

\begin{document}

\begin{abstract}
  Reconstructing a structured vector-graphics representation from a rasterized floorplan image is typically an important prerequisite for computational tasks involving floorplans such as automated understanding or CAD workflows. However, existing techniques struggle in faithfully generating the structure and semantics conveyed by complex floorplans that depict large indoor spaces with many rooms and a varying numbers of polygon corners. To this end, we propose \emph{Raster2Seq}, framing floorplan reconstruction as a sequence-to-sequence task in which floorplan elements—such as rooms, windows, and doors—are represented as labeled polygon sequences that jointly encode geometry and semantics. Our approach introduces an autoregressive decoder that learns to predict the next corner conditioned on image features and previously generated corners using guidance from learnable anchors. These anchors represent spatial coordinates in image space, hence allowing for effectively directing the attention mechanism to focus on informative image regions. By embracing the autoregressive mechanism, our method offers flexibility in the output format, enabling for efficiently handling complex floorplans with numerous rooms and diverse polygon structures. Our method achieves state-of-the-art performance on standard benchmarks such as Structure3D, CubiCasa5K, and Raster2Graph, while also demonstrating strong generalization to more challenging datasets like WAFFLE, which contain diverse room structures and complex geometric variations. 
  
  \begin{center}
\fbox{\parbox{0.95\linewidth}{%
  \centering Project page at \url{https://cornell-vailab.github.io/Raster2Seq/}
}}
\end{center}
\end{abstract}

\begin{teaserfigure}
\centering \includegraphics[width=0.999\textwidth]{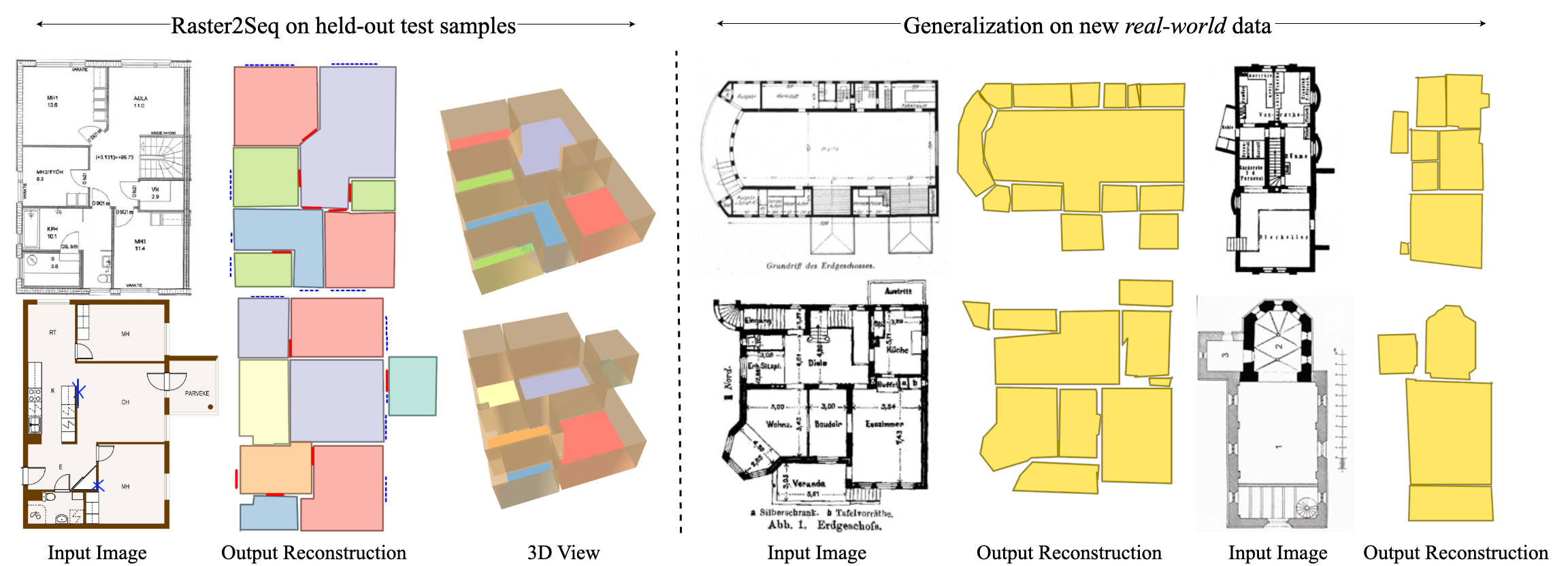}
  \vspace{-10pt}
  \caption{Our approach transforms rasterized floorplan images to vectorized format, reconstructing both its structure and semantics. We illustrate$^*$ results on held-out CubiCasa5K~\cite{kalervo2019cubicasa5k} test samples (left). The colors denote unique semantic categories (\emph{e.g.},  \textcolor{outdoor}{Outdoor}, \textcolor{bedroom}{Bedroom}, \textcolor{bath}{bath}, and \textcolor{entry}{entry}). Additionally, we highlight our model's generalization capabilities over complicated real-world floorplan images from WAFFLE~\cite{ganon2025waffle} (right). $^*${3D visualizations are constructed by extending the 2D boundaries vertically.}
  }
  \label{fig:teaser}
\end{teaserfigure}

\maketitle

\section{Introduction}
\label{sec:intro}

Floorplans are a fundamental element of architectural design that define the structure and semantics of indoor spaces, from the tiny studio apartment in Manhattan to the historic Café Helms in Berlin (depicted in the top right corner of Figure \ref{fig:teaser}).  
While floorplans are typically drawn in a vector-graphics representation using specialized softwares (e.g., AutoCAD), they are usually distributed in rasterized image formats. This rasterization process strips away the structured geometric and semantic information, severely limiting their utility for computational tasks such as automated editing \cite{paschalidou2021atiss,shum2023conditional,zhang2024sceneexpander}, \change{floorplan understanding and generation \cite{wang2015lost,narasimhan2020seeing,shabani2023housediffusion}}, or 3D reconstruction \cite{martin20143d,liu2015rent3d,nguyen2024housecrafter}.

To unlock computational capabilities over rasterized floorplans, several works have explored the \emph{raster-to-vector} conversion task \cite{de2014statistical,liu2017raster,zeng2019deep}, which aims to transform an input floorplan image back to vectorized format. However, despite the significant advancements enabled by Transformer-based architectures~\cite{chen2022heat,yue2023roomformer,hu2024r2g}, existing methods face challenges in capturing the structure and semantics conveyed by complicated real-world floorplans, often depending on pretrained detectors and constructing sub-optimal multi-stage pipelines for performing the conversion.

In this work, we propose \methodname{}, an approach that transforms rasterized floorplan images to vectorized format using a labeled polygon sequence representation. Unlike prior work that simultaneously predict all structural floorplan elements~\cite{stekovic2021montefloor,yue2023roomformer,chen2022heat} \change{and are therefore limited by a fixed-query budget constraint}, our framework autoregressively outputs a polygon sequence, directly modeling both spatial structure and semantic attributes.
Our key observation, motivating our framework design, is that floorplan elements can be effectively modeled as a sequence, leveraging the left-to-right generation bias of masked attention models~\cite{vaswani2017attention}.
This allows us to decompose floorplan reconstruction into interpretable, sequential predictions mirroring the natural CAD design workflow.
We represent each polygon as a sequence of labeled corners, \emph{i.e.}, spatial coordinates labeled with semantic information, and sort the floorplan's polygons using a left-to-right ordering. 
Specifically, we consider rooms, windows and doors, but this representation could easily accommodate additional labeled entities.
At its core, our framework introduces an anchor-based autoregressive decoder that effectively fuses information from image features and the previously generated corners to predict the next labeled corner. In particular, our autoregressive module is guided by learnable anchors that direct the attention mechanism to focus on informative regions, enabling for efficiently handling complex floorplan images. We achieve this without sacrificing semantic fidelity by additionally introducing a token-level semantic classification loss that supervises semantic information over individual corner embeddings.

We show the effectiveness of our framework on multiple benchmarks, conducting experiments in different floorplan reconstruction settings that consider both rasterized RGB images and 2D density maps as input. Our approach consistently surpasses existing methods over a wide range of geometric and semantic metrics. Notably, our results show that more complicated floorplans---containing higher quantities of corners and rooms---yield larger performance gaps. We also show strong generalization capabilities over challenging real-world Internet datasets, demonstrated both qualitatively and quantitatively.

\section{Related Work}
\label{sec:rw}

\subsection{Floorplan Reconstruction}

Raster-to-vector floorplan conversion aims to reconstruct vectorized representations from rasterized floorplan images. Prior to deep learning, multi-step systems~\cite{mace2010system,ahmed2011improved,de2014statistical} relied on handcrafted features to detect floorplan components (e.g. walls). Liu \emph{et al.} \shortcite{liu2017raster} first integrated neural networks for solving this task, predicting corner representations followed by integer programming to recover geometric primitives. Subsequent works utilized pixel-wise segmentation \cite{zeng2019deep} and graph neural networks~\cite{sun2022wallplan} to model hierarchical relationships among floorplan elements. Raster2Graph~\cite{hu2024r2g} employs a transformer~\cite{zhu2021deformable} with image-space augmentation to highlight visible corners for sequential corner prediction. By contrast, our method formulates floorplan conversion as a sequence-to-sequence task, generating polygon coordinates autoregressively. This naturally handles variable-length polygons and dense layouts without requiring image augmentation or corner sampling strategies.

Several works address related floorplan reconstruction tasks using different modalities such as point-cloud density maps~\cite{stekovic2021montefloor,chen2022heat,yue2023roomformer} and RGB panoramas~\cite{cabral2014piecewise,liu2018floornet}, rather than rasterized floorplan images. Early methods like Floor-SP~\cite{chen2019floor} and MonteFloor~\cite{stekovic2021montefloor} frame the task as instance segmentation with additional optimization steps, but these multi-stage pipelines typically generalize poorly to diverse floorplan layouts. More recent end-to-end approaches eliminate post-optimization: HEAT~\cite{chen2022heat} and FRI-Net~\cite{xu2024fri} follow bottom-up strategies—detecting corners then classifying edges, or predicting line primitives then grouping them into rooms. RoomFormer~\cite{yue2023roomformer} and PolyRoom~\cite{liu2024polyroom} formulate floorplan reconstruction as object detection, predicting room coordinates through numerous object queries (e.g., 2800) with Hungarian matching. While these methods were originally designed for 3D-scan-based inputs, we demonstrate that they can be adapted for raster-to-vector conversion. However, as demonstrated in our experiments, when floorplan complexity exceeds this fixed query capacity, performance degrades significantly. Moreover, these methods cannot output a number of predictions beyond a predefined number of corners and rooms per image. By contrast, our method is not limited by a fixed number of predctions, generating ordered, non-redundant outputs sequentially, without additional post-processing steps for extracting semantic predictions.

\minihead{Semantic integration.} Unlike most prior work \change{\cite{chen2023polydiffuse,liu2024polyroom}} that focuses solely on structural prediction, our method also incorporates semantic information. RoomFormer and Raster2Graph also integrate semantics. However, RoomFormer loses fine-grained semantic information by averaging corner embeddings within uniform-length room sequences—inevitably including padding corners—before classification. Raster2Graph introduces unnecessary complexity by predicting four neighbor room classes per corner, causing potential error propagation and additional computational overhead. 
In contrast, we proposed \change{a labeled polygon sequence, employing a granular token-level supervision, where each corner receives direct gradient updates without dilution from padding.} Since rooms are inherently variable-length polygons, our token-level loss naturally aligns with this representation.

\begin{figure*}[t]
\centering
\includegraphics[width=0.999\linewidth]{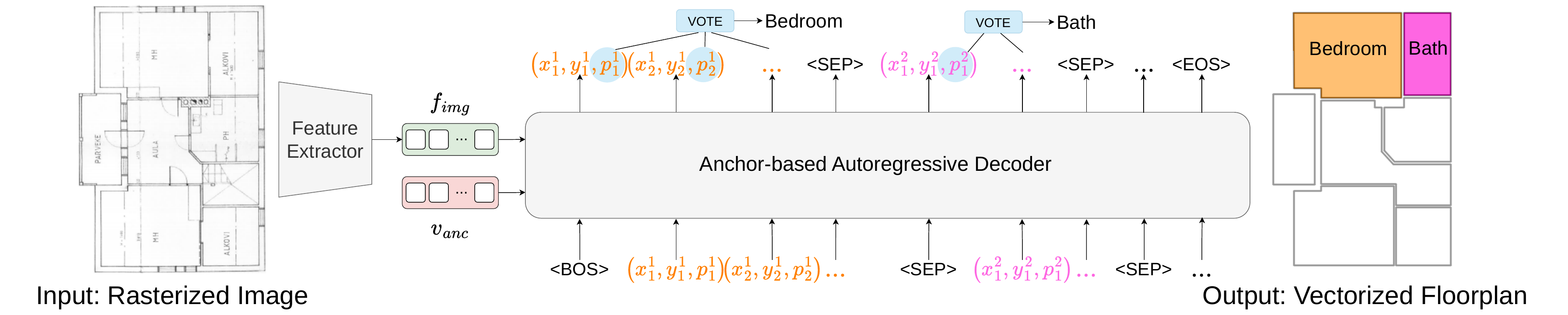}
\caption{\textbf{Method Overview}. Given a rasterized floorplan image (left), our approach converts it into vectorized format, represented as a labeled polygon sequence, separated using special \texttt{<SEP>} tokens. The main architectural component of our framework is an anchor-based autoregressive decoder, which predicts the next token given image features ($f_{img}$), learnable anchors ($v_{anc}$) and the previously generated tokens; see Section \ref{sec:arch} for additional details. Above, we visualize the first two labeled polygons predicted (colored in orange and pink, respectively). }
\label{fig:system}
\end{figure*}

\subsection{Sequence-to-Sequence Modeling for Visual Tasks}

Sequence-to-sequence (seq2seq) modeling~\cite{sutskever2014sequence} was originally proposed for machine translation, with the goal of learning a mapping from a source sequence to a target sequence. This framework was later adapted to a plethora of computer vision tasks by providing image features as input to a decoder (typically an RNN or Transformer) that generates a target sequence. Notable applications include image captioning~\cite{vinyals2015show,xu2015show,cornia2020meshed}, object detection~\cite{chen2021pix2seq}, instance segmentation~\cite{acuna2018efficient,liu2023polyformer,chen2022unified}, and image generation~\cite{ramesh2021zero,yu2022scaling}. The seq2seq paradigm enables end-to-end training and naturally accommodates inputs and outputs of variable lengths, eliminating the need for complex post-processing. This paradigm was adopted by Liu \emph{et al.}~\cite{liu2023polyformer} for representing object segmentations as polygon sequences, which can be utilized for the task of prompt-based segmentation. %
While our method is conceptually similar, our framework introduces several representation and architectural differences for performing floorplan reconstruction. For example, beyond predicting spatial coordinates, we introduce semantic labels into the representation and incorporate a novel semantic training objective for semantic-aware floorplan recognition. This semantic integration improves the utility of vectorized floorplans by producing both structural information and semantic labels.

Prior work has explored the effectiveness of recursive frameworks in modeling complex and structured visual data. For instance, GRASS~\cite{li2017grass} GRAINS~\cite{li2019grains}, READ~\cite{patil2020read}, SceneScript~\cite{avetisyan2024scenescript} demonstrated the utility of recursive prediction for 3D shapes, 3D indoor scene synthesis, 2D document layout generation, and 3D scene reconstruction, respectively.
More closely related to our work, SceneScript formulates 3D scenes as text representations and learns to generate house layouts from input point clouds using predefined text commands for drawing objects (e.g. wall and object box). In our work, we adopt the sequence-to-sequence framework for floorplan transformation, predicting semantic polygon coordinates sequentially based on corner-based representation instead.

\section{Method}
\label{sec:method}

An overview of our proposed method is presented in \Cref{fig:system}. Our goal is to transform a rasterized floorplan image into vectorized format, reconstructing both its structure and semantics. Specifically, we assume that we are provided with an RGB image of a rasterized floorplan \( I \in \mathbb{R}^{H \times W \times 3} \), where \( H \) and \( W \) denote the height and width of the image. The input image $I$ is encoded via a \emph{Feature Extractor} module to produce a feature vector $f_{img}\in \mathbb{R}^{L_I \times D} $ where $L_I$ is the length of the image features and $D$ is the number of channels.

Unlike existing floorplan reconstruction techniques~\cite{zeng2019deep,stekovic2021montefloor,sun2022wallplan, chen2022heat} that extract vectorized floorplans via intermediate geometric elements such as edges, corners, or room segments, we propose to represent vectorized floorplans directly using a sequence of labeled polygons. We introduce this representation in Section \ref{sec:floorplan_representation}. We then describe our \emph{Anchor-based Autoregressive Decoder} module, the main architectural component in our framework, in Section \ref{sec:arch}.  Finally, training and inference details are discussed in Section \ref{sec:train}.

\subsection{Labeled Polygon Sequence Floorplan Representation}
\label{sec:floorplan_representation}
We propose to represent vectorized floorplans using labeled polygon sequences. By labeled, we refer to the polygon's \emph{semantics}. For instance, a room can be labeled as a \emph{kitchen}, \emph{bedroom}, etc.
We parameterize a polygon as a sequence of labeled corner tokens $c$, where $c_i=(x_i,y_i,p_i)$ denotes the $i$-th corner in the polygon, $v_i = (x_i,y_i)$ denotes its spatial position, and ${p}_i \in[0,1]^C$ denotes its semantic probability vector (assuming $C$ unique semantic categories). As we elaborate later in \Cref{sec:train}, room-level semantic predictions are obtained by aggregating semantic information at the token-level. We also consider windows and doors, in addition to rooms. These are simply represented as two additional semantic categories (on top of the room types).

To represent a floorplan that contains multiple rooms (or floorplan \emph{entities}, such as windows)---each represented as a labeled polygon, as detailed above---we concatenate their sequences using a separator  \texttt{<SEP>} token. We also use \texttt{<BOS>} and \texttt{<EOS>} tokens to indicate the beginning and the end of the sequence. Put together, the labeled polygon sequence is structured as follows: %
\begin{equation*}
    [\texttt{<BOS>}, c^1_1, c^1_2, \cdots, \texttt{<SEP>}, c^n_1, c^n_2, \cdots, \texttt{<EOS>} ]
\end{equation*}

As \ourmodel is trained to regress continuous values without relying on a discrete tokenizer, each token is augmented with a token type probability vector $q\in[0,1]^3$, where the three token type categories are \texttt{<CORNER>}, \texttt{<SEP>} or \texttt{<EOS>}; a similar augmentation strategy was recently utilized in \cite{li2024autoregressive}. During training, the \texttt{<CORNER>} type is used as a supervision label for each corner token $c_i$ but is not explicitly included in the sequence. \texttt{<BOS>} is omitted from the token type modeling.
The training objective is to predict the next corner token in the sequence, where the output sequence contains the target tokens to be predicted; see Figure \ref{fig:system}.

\subsection{Anchor-based Autoregressive Decoder}
\label{sec:arch}

\begin{figure}[t]
\centering
\hfill \hfill
\includegraphics[width=0.95\linewidth,clip]{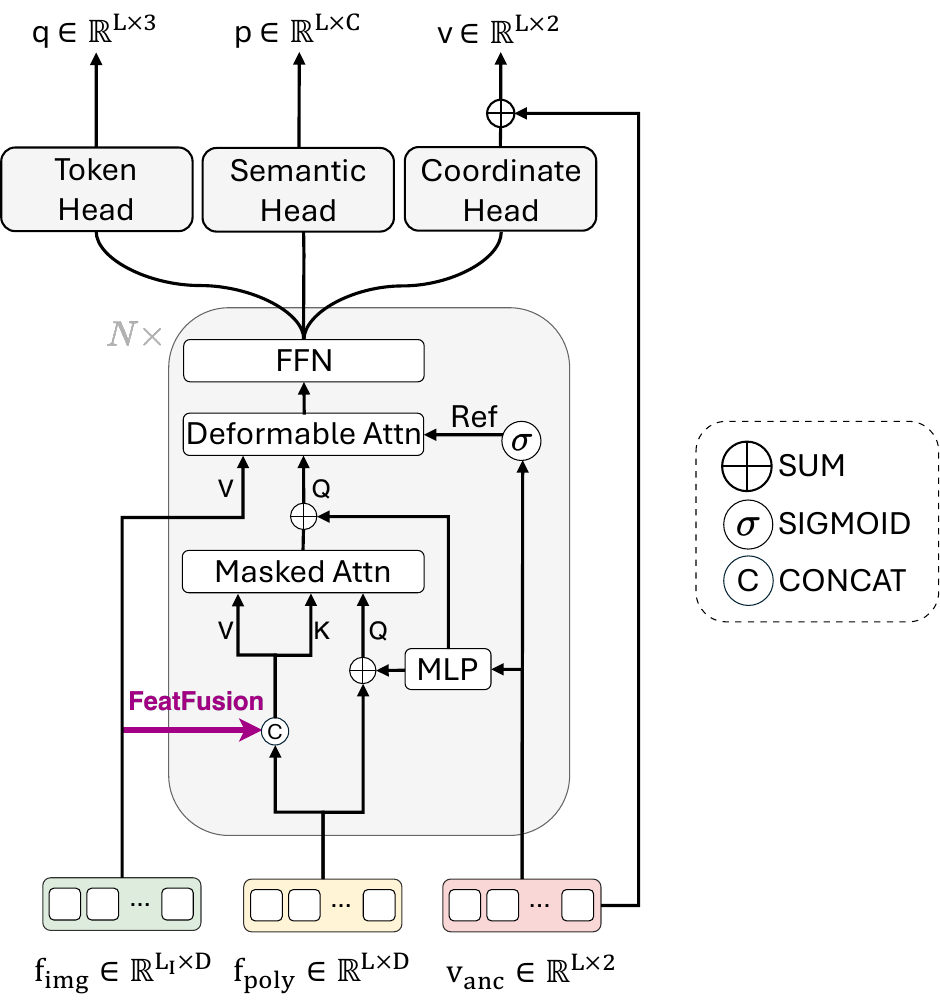}
\caption{Illustration of our anchor-based autoregressive decoder. }
\label{fig:decoder}
\end{figure}

Next, we present our \emph{Anchor-based Autoregressive Decoder} module which predicts labeled polygon sequences; see Figure \ref{fig:decoder} for an illustration. 
Our proposed module is provided with three different inputs: (i) image features extracted with the \emph{Feature Extractor} module, (ii) a sequence of coordinate tokens, and (iii) learnable anchors.

The sequence of coordinate tokens are provided after quantization of the continuous 2D coordinates into a discrete 1D embedding space using a learnable codebook $C\in\mathbb{R}^{H_b \times W_b\times D}$, where $H_b \times W_b$ is number of quantization bins and $D$ is embedding dimension; additional details are provided in the supplementary material. Specifically, the decoder is provided with $L$ coordinate tokens, which are denoted by $f_{poly} \in \mathbb{R}^{L \times D}$.
Learnable anchors, denoted by $v_{anc} \in \mathbb{R}^{L\times2}$, are introduced to avoid direct regression of continuous coordinate values. Instead, the model learns residuals relative to these anchors.  %
\change{The concept of anchors draws inspiration from object detection methods \cite{lin2017focal,zhang2020bridging}, which leverage assigned anchors to produce reliable predictions. As illustrated in our experiments, adopting this concept for our problem setting results in  significant performance gains.}

\minihead{Decoder Architecture.} The decoder contains an autoregressive block that contains three different layers: masked attention, deformable attention, and a feed-forward network layer. In the masked attention layer, a causal mask is applied to ensure that each token can only attend to its preceding tokens, reinforcing a left-to-right generation bias~\cite{vaswani2017attention}. %
As shown in \cref{fig:decoder}, the triplet of query (Q), key (K), and value (V) vectors is derived from the sequence of coordinate tokens. The query vector includes additional positional embeddings from the introduced anchors,
while the key and value vectors are derived from a fused feature vector of shape $[L_I+L, D]$. This fused vector combines image features from the encoder with coordinate-token embeddings through tensor concatenation, referred to as \emph{FeatFusion} (highlighted in \textcolor[RGB]{163,0,133}{purple} in Figure \ref{fig:decoder}). 
We find that this early fusion is crucial for precise coordinate regression. Intuitively, the image features act as a prefix that each token can attend to, providing additional contextual information during decoding. %

Subsequently, the output vectors from the preceding masked attention layer serve as queries in a deformable attention module.
This module, first introduced in \cite{zhu2021deformable}, is an efficient attention-based mechanism that---given a feature map and a set of reference points---for each query, only attends to a small set of sampling points around each reference point, rather than the entire feature map. In our autoregressive decoder, this mechanism allows for attending to a sparse set of relevant spatial positions in the image feature map $f_{img}$.
Specifically, input anchor points are first normalized to [0,1] using a sigmoid function. The deformable attention layer then takes in the query vector and predicts offsets relative to these normalized anchor points using a linear layer. These offsets are added to the anchor points to produce sampling points, allowing the attention mechanism to focus on informative regions of image features. As previously mentioned, the anchor points are learnable parameters that are randomly initialized and learned jointly with the network weights.

Finally, the decoder module contains three lightweight heads on top of the last autoregressive block: a token head for predicting token types, a semantic head for predicting semantic labels, and a coordinate head for predicting 2D corner coordinates. The coordinate head essentially produces residual outputs which are combined with the learnable anchors for producing continuous coordinate values, as illustrated in Figure \ref{fig:decoder}.

\subsection{Training and Inference Details}
\label{sec:train}

Our method is supervised using three different loss functions: a coordinate regression loss, a token-type classification loss, and a semantic classification loss.

\minihead{Coordinate loss.} For the coordinate loss, we use a L1 loss to measure the difference between the predicted coordinates $\hat{v}$ and the ground-truth spatial coordinates $v$, across all $L$ tokens (\emph{i.e.}, corners) in the sequence:
\begin{equation}
    \mathcal{L}_{coord} = \frac{1}{L}\sum_{l=1}^L \mathbf{m}_l \vert  \hat{v}_l - v_l \vert,
\end{equation}
This loss is computed only over non-padded tokens, using an additional mask $\mathbf{m}$ to exclude irrelevant positions. The same masking strategy is applied to the other losses described below.

\minihead{Token-type loss.} As defined as in Section \ref{sec:floorplan_representation}, we consider three token classes: \texttt{<CORNER>}, \texttt{<SEP>}, and \texttt{<EOS>}. The model is trained to classify individual token into one of these categories using a standard cross-entropy loss:

\begin{equation}
    \mathcal{L}_{token} = \frac{1}{L}\sum_{l=1}^L  \mathbf{m}_l \text{CE}(\hat{q}_l, q_l),
\end{equation}
where $\hat{q}_l$ is the predicted probability distribution over three token types, and $q_l$ is the ground-truth one-hot vector for the $l$-th token.

\minihead{Semantic loss.}
We supervise prediction of semantic labels using a cross-entropy loss defined for each token:
\begin{equation}
    \mathcal{L}_{sem} = \frac{1}{L} \sum_{l=1}^L \mathbf{m}_l CE(\hat{p_l}, p_l),
\end{equation}
where $\hat{p}_l$ is the predicted probability distribution over $C$ predefined room classes, and $p_l$ is the one-hot vector representing the ground-truth room class for the $l$-th token in the sequence.

\medskip \noindent 
The total training loss is:
\begin{equation}
    \mathcal{L} = \lambda_{coord} * \mathcal{L}_{coord}  +  \lambda_{token} * \mathcal{L}_{token}  +  \lambda_{sem} * \mathcal{L}_{sem},
\end{equation}
where $\lambda_{coord}$, $\lambda_{token}$ and $\lambda_{sem}$ are  weighting coefficients. 
To induce strong geometric inductive bias, we perform a left-to-right ordering of the polygon sequence during training, where rooms are ordered by top-left coordinates using top-to-bottom, left-to-right scanning priority. As illustrated in our experiments,
the model implicitly captures topological relationships between corners, which results in improved performance.

\begin{figure}[t]
\centering
\vspace{-5pt}
\includegraphics[width=0.999\linewidth,trim={0.1cm 0.1cm 0.1cm 0.1cm},clip]{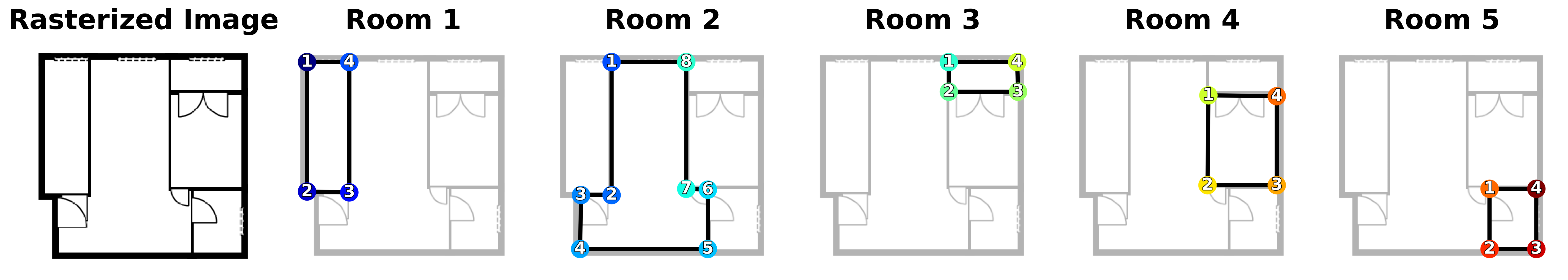}
\vspace{-15pt}
\caption{Given an input rasterized image, our method performs sequential corner prediction. We visualize earlier corners in cooler colors (predictions are enumerated per room). As illustrated above, within each room, corners are predicted in counterclockwise order.}
\label{fig:seq_vis}
\end{figure}

\medskip \noindent 
At inference, \ourmodel predicts tokens sequentially till a \texttt{<EOS>} token is obtained. 
To predict semantic room labels, we aggregate token-level predictions using a majority voting strategy. Specifically, the room label for each polygon sequence is determined by first selecting the class with the highest probability at each token, and then taking the most frequently predicted class across the sequence.
Figure \ref{fig:seq_vis} provides a visualization of the sequential room prediction process, illustrating how the model maintains a left-to-right generation pattern.
Additional details are provided in the supplementary material.

\section{Experiments}
\label{sec:exp}
In this section, we first describe the experimental setup and the baselines we compare our method against (Section \ref{sec:setup}). We then present our main quantitative results (Section \ref{sec:eval}), followed by both a qualitative comparison (Section \ref{sec:res}). Finally, we present an ablation study of our proposed method (Section \ref{sec:ablations}). %
Additional details, experiments, ablations, \change{a runtime comparison,} and a discussion of limitations are provided in the appendix.

\subsection{Experimental Setup} \label{sec:setup}

\minihead{Datasets.} We conduct experiments on four datasets: Structured3D \cite{zheng2020structured3d},  Cubicasa5K~\cite{kalervo2019cubicasa5k}, Raster2Graph~\cite{hu2024r2g}, and WAFFLE~\cite{ganon2025waffle}. Structured3D is a 3D point cloud dataset containing 3,000/250/250 training/val/test samples, annotated with 16 room types. CubiCasa5K is a raster-based floorplan dataset with 4,199/399/399 training/validation/test samples, annotated with 11 classes. Raster2Graph has 9,803/500/499 training/validation/test samples, annotated with 12 classes. WAFFLE contains 20K real-world floorplan images scraped from the Internet. As this dataset only contains approximately 100 annotated samples, we only evaluate zero-shot generalization capabilities on this data. %

For Structured3D, existing work~\cite{yue2023roomformer} use the projection of 3D point clouds along the vertical axis as input images. Since our focus is on raster-to-vector floorplan reconstruction, we convert the Structured3D samples into binary raster images using the ground-truth annotations, yielding images resembling typical floorplans which are used to train our method. We refer to this converted dataset as Structured3D-B for convenience. Some CubiCasa5K images contain multiple floorplans, so we preprocess them into separate images, increasing the dataset size from 5,000 to 6,281 samples (5,267 train / 503 val / 511 test). We use a fixed resolution of $256\times 256$ for all datasets in all experiments.

\begin{table}[t]
\small
\centering
\caption{Quantitative comparison on Structured3D-B, CubiCasa5K, and Raster2Graph datasets, evaluating F1 scores across geometric (Room, Corner, Angle) and semantic (Room, Window \& Door) predictions. Note that not all models include semantic predictions, and the Raster2Graph dataset does not include Window \& Door annotations. Furthermore, the Raster2Graph model can only be evaluated on their dataset, as their approach requires per-corner neighboring room class annotations. }
\label{tab:merged_evaluation}
\resizebox{\linewidth}{!}{%
\begin{tabular}{lccccc}
\toprule
Method & Room & Corner & Angle & Room Semantic & Window \& Door \\
\midrule
\multicolumn{6}{c}{\textbf{Structured3D-B}} \\
\midrule
HEAT & 94.7 & 84.5 & 79.6 & - & - \\
PolyRoom & 98.9 & 96.0 & 91.9 & - & - \\
FRI-Net & 96.5 & 85.4 & 83.3 & - & - \\
RoomFormer & 95.1 & 91.7 & 83.2 & 74.2 & 94.1 \\
Ours & \textbf{99.6} & \textbf{98.3} & \textbf{92.7} & \textbf{76.9} & \textbf{98.5} \\
\midrule
\multicolumn{6}{c}{\textbf{CubiCasa5K}} \\
\midrule
HEAT & 78.2 & 53.7 & 32.3 & - & - \\
PolyRoom & 54.1 & 37.1 & 23.0 & - & - \\
FRI-Net & 77.1 & 50.8 & \textbf{38.0} & - & - \\
RoomFormer & 83.5 & 55.5 & 34.1 & 63.0 & \textbf{78.5} \\
Ours & \textbf{88.7} & \textbf{59.4} & \underline{37.4} & \textbf{63.8} & 77.8 \\
\midrule
\multicolumn{6}{c}{\textbf{Raster2Graph}} \\
\midrule
HEAT & 95.9 & 79.7 & 50.9 & - & - \\
PolyRoom & 56.9 & 42.4 & 23.8 & - & - \\
FRI-Net & 91.5 & 72.3 & 52.8 & - & - \\
RoomFormer & 91.9 & 74.5 & 51.1 & 79.5 & - \\
Raster2Graph & 95.0 & 78.3 & \textbf{67.3} & 83.4 & - \\
Ours & \textbf{97.0} & \textbf{80.3} & \underline{66.6} & \textbf{85.1} & - \\
\bottomrule
\end{tabular}%
}

\end{table}

\minihead{Metrics.} We follow the evaluation protocol used by prior work \cite{stekovic2021montefloor}, focusing on geometric and semantic metrics obtained from matching model predictions with the ground truth annotations. Three evaluation criteria are Room, Corner, and Angle where each criterion is evaluated using Precision, Recall, and F1 score. Specifically, we first match each ground-truth room with the best-predicted room based on Intersection over Union (IoU), and use these matched pairs to compute evaluation metrics at three levels: room, corner, and angle. For room-level evaluation, a match is considered valid if the IoU exceeds 0.5. For corner and angle evaluation, which are point-wise metrics, we follow the protocol of \cite{stekovic2021montefloor} by computing the L2 distance and the oriented angle between predicted and ground-truth corners. A corner is considered correctly recovered if the distance is within 10 pixels and the angle difference is less than 5 degrees. For semantic label evaluation, room type predictions are additionally used for finding matches. By default, we use F1 score for Room, Corner, and Angle for all evaluations. %
For WAFFLE, we report room prediction performance using IoU score to access zero-shot performance on the segmentation task. Additional metrics are reported in \cref{subsec:full_comp}.

\minihead{Baselines.}
We mainly utilize HEAT~\cite{chen2022heat}, RoomFormer~\cite{yue2023roomformer}, FRI-Net~\cite{xu2024fri}--models originally designed for point-cloud density maps—for conducting a quantitative evaluation, finetuning these models to perform floorplan reconstruction from rasterized floorplan inputs. We also compare our method against Raster2Graph~\cite{hu2024r2g} on raster-to-vector conversion task using their provided dataset. Since Raster2Graph requires per-corner neighboring room class annotations, we can only evaluate it on the Raster2Graph dataset proposed by its authors.

\subsection{Quantitative Evaluation}
\label{sec:eval}

\begin{figure}[t]
\centering
\vspace{-5pt}
\includegraphics[width=1.0\linewidth]{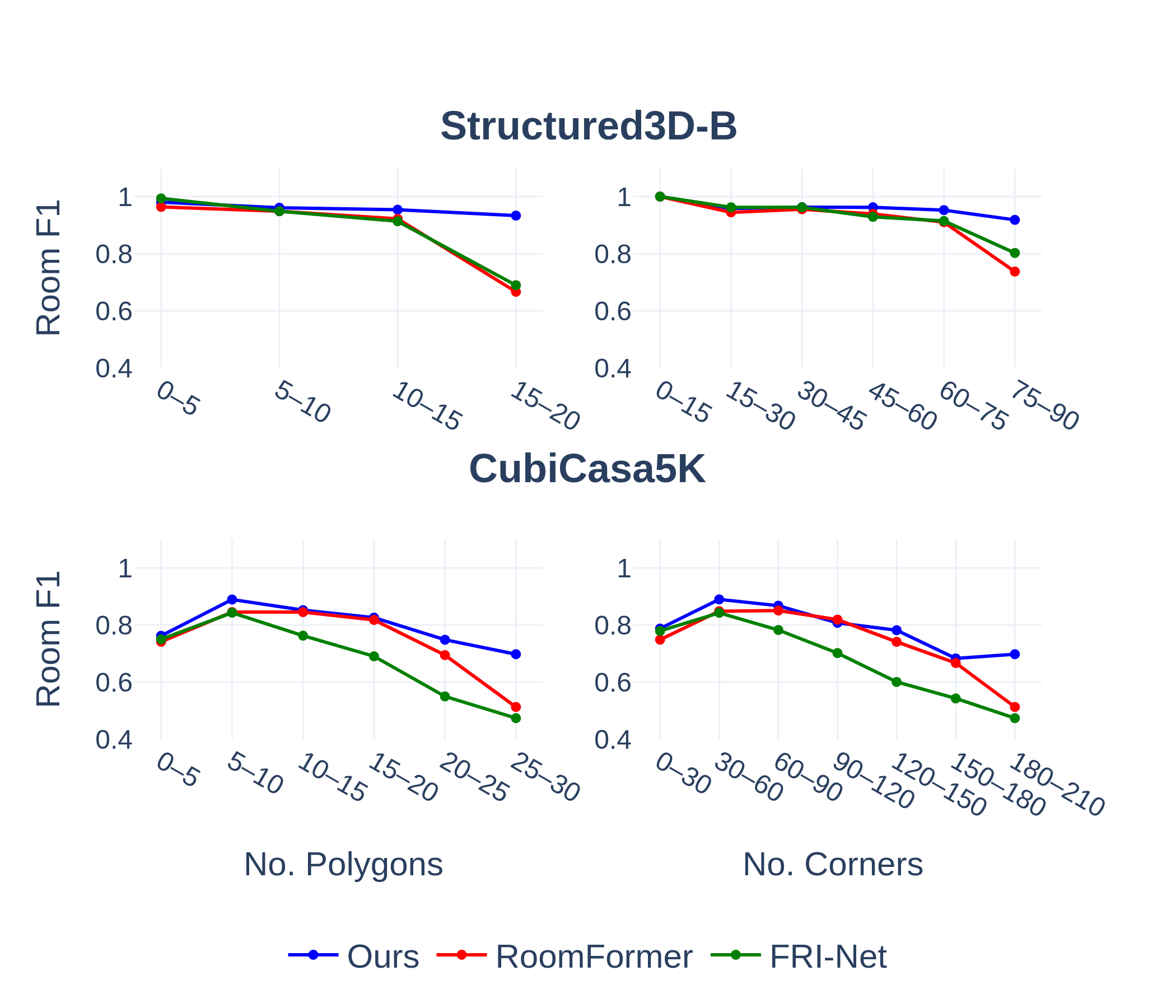}
\vspace{-25pt}
\caption{Performance vs. floorplan complexity---as approximated by the total number of polygons (left) and the total number of corners (right). As illustrated above over Structured3D-B (top) and CubiCasa5K (bottom), our approach yields larger gains as the floorplan complexity increases. }
\label{fig:count_impact}
\end{figure}

We compare performance over the raster-to-vector conversion task across three datasets (see \Cref{tab:merged_evaluation}). Overall, our method achieves state-of-the-art performance on both structural metrics (Room and Corner) and semantic metrics (Room Semantic and Window \& Door). We note that unlike our method that directly optimizes token-level semantic predictions, RoomFormer dilutes semantic information by averaging irrelevant corners within uniform-length sequences, resulting in inferior semantic predictions as evident across nearly all semantic metrics.

Interestingly, several methods exhibit a high variance in performance across different datasets. In particular, both PolyRoom and FRI-Net achieve very high performance performance on the simpler Structured3D-B dataset, while achieving significantly lower scores on more complex datasets like CubiCasa5K and Raster2Graph, where polygon lengths are more diverse and shapes are irregular. We hypothesize that PolyRoom's reliance on segmentation proposals limits its performance to regular and simple floorplans, while FRI-Net's dependence on line assembly to form rooms proves challenging in diverse scenarios. By contrast, our method achieves strong and stable performance across all three datasets.

\minihead{Model Robustness To Floorplan Complexity.} \Cref{fig:count_impact} shows the Room F1 performance of RoomFormer, FRI-Net, and our model across varying numbers of polygons and corners on the Structured3D-B and CubiCasa5K datasets. Our method consistently demonstrates greater robustness as floorplan complexity increases. While both models perform similarly on simpler cases, RoomFormer and FRI-Net exhibit a notable performance drop in complex scenes with over 15 polygons or 150 corners. Importantly, RoomFormer operates with a fixed number of room queries (e.g., 2800). Exceeding this capacity causes out-of-memory errors and increased computation due to quadratic attention costs, thus degrading performance on complex floorplans. By contrast, our recursive approach decomposes floorplan reconstruction into sub-problems, improving interpretability and naturally handling variable-length inputs.

\begin{figure}[t]
\centering
\vspace{-5pt}
\includegraphics[width=1.0\linewidth]{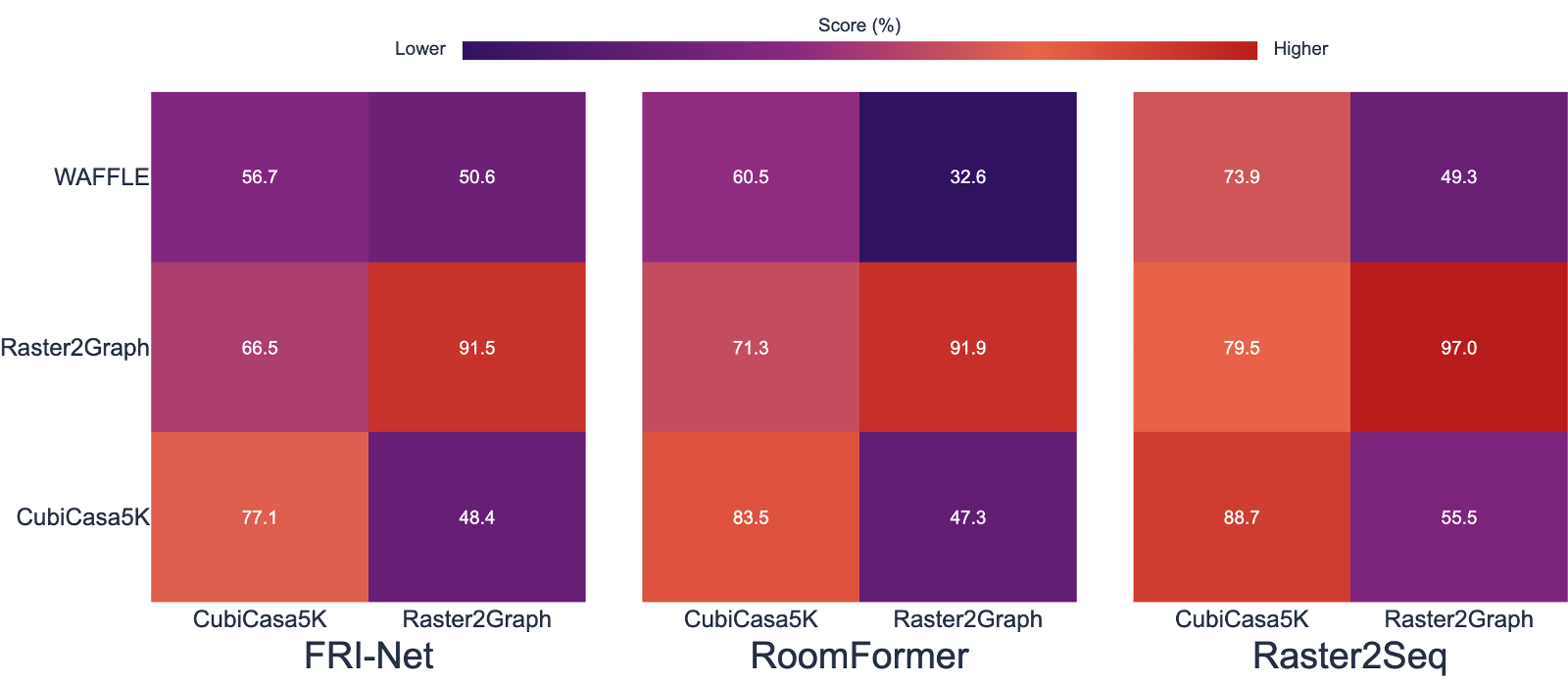}
\vspace{-15pt}
\caption{Cross-evaluation heatmaps showing performance across training (columns) and test (rows) dataset combinations, with lighter colors denoting higher performance. R2G and CC5K denote Raster2Graph and CubiCasa5K datasets, respectively. Our method exhibits strong generalization across different settings, substantially outperforming FRI-Net and RoomFormer.}
\label{fig:heatmap}
\end{figure}

\minihead{Model Generalization.} We perform a cross-evaluation experiment across different train-test dataset configuration.  We evaluate performance using metrics reported previously, using RoomF1 for the CubiCasa5K and Raster2Graph dataset and IoU for WAFFLE. Results are reported in \Cref{fig:heatmap}. As shown, our method demonstrates the strongest generalization performance across various settings, including both same-dataset and cross-dataset evaluations, outperforming other baselines by a large margin. In particular, we observe significant gaps over WAFFLE test set between our method and the counterparts, further demonstrating its robustness on complex and unseen floorplan samples.

\change{
\minihead{Additional Settings and Applications.} In the appendix, we present several additional analyses and extensions of our framework. First, we provide a quantitative comparison on the standard Structured3D benchmark using density map inputs rather than rasterized floorplans (see \cref{sec:density_maps}), showing that our method achieves competitive results with existing baselines under this additional problem setting. We also provide an experiment on testing the robustness of our method in dealing with noisy density map inputs. Additionally, we showcase a downstream application made possible by our vectorized floorplan representation.  Specifically, we use our vectorized floorplan as guidance for generating controllable 3D scenes (see \cref{subsec:trellis}). Finally, we introduce a VLM-based vectorization refinement procedure (see \cref{subsec:VLM_refinement}) that naturally builds on our polygon sequence representation and further improves reconstruction accuracy, highlighting the flexibility of our representation for integrating higher-level reasoning modules.
}

\begin{figure}[t]
    \centering
    \setlength{\tabcolsep}{1pt}
    \begin{tabular}{cccccc}
        \includegraphics[width=0.16\linewidth]{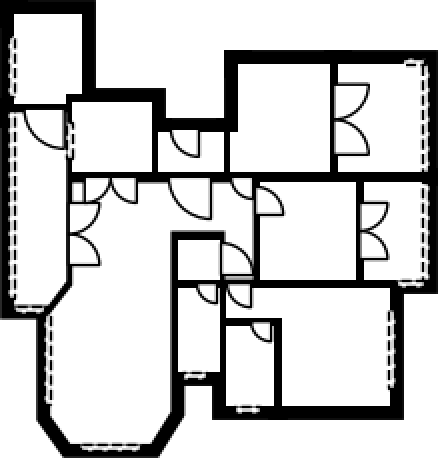} &
        \includegraphics[width=0.16\linewidth]{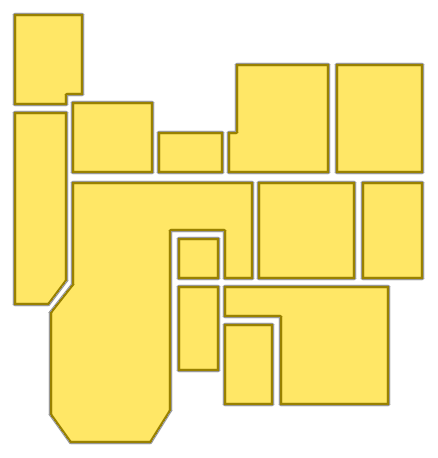} &
        \includegraphics[width=0.16\linewidth]{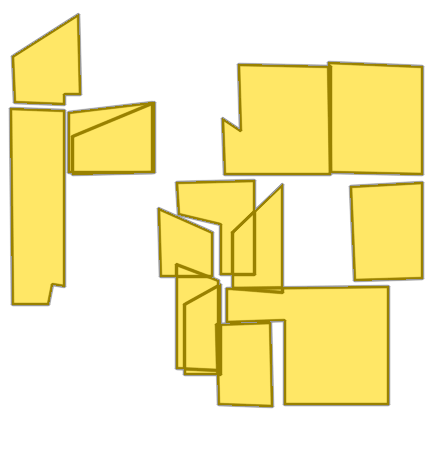} &
        \includegraphics[width=0.16\linewidth]{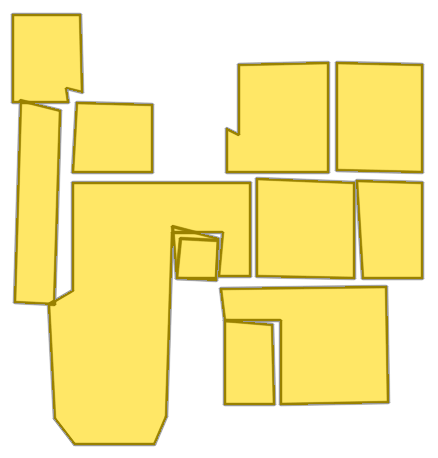} &
        \includegraphics[width=0.16\linewidth]{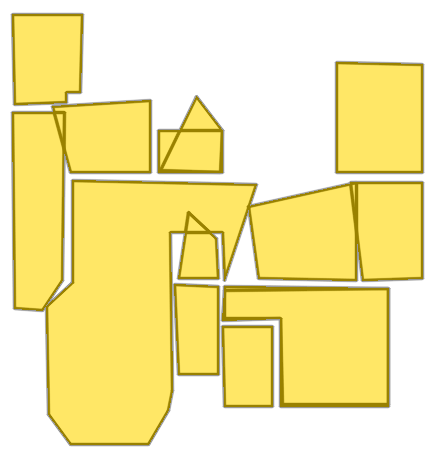} &
        \includegraphics[width=0.16\linewidth]{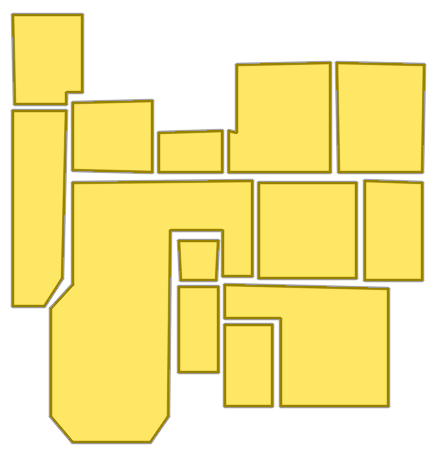} \\
        {\small Input} & {\small GT} & {\small Base} & {\small +FeatFusion} & {\small +Anchor} & {\small +Ordering} \\
    \end{tabular}
    \vspace{-8pt}
    \caption{Ablation results over a sample from the Structure3D-B test set. As illustrated above, incorporating our proposed components significantly improves geometric reconstruction accuracy and alignment with the groundtruth.}
    \label{fig:abl_vis}
\end{figure}

\subsection{Qualitative Results}
\label{sec:res}

\begin{figure*}[!ht]
\centering %
\rotatebox{90}{\whitetxt{sssssssss}Input}
\jsubfig{
\includegraphics[height=2.8cm]{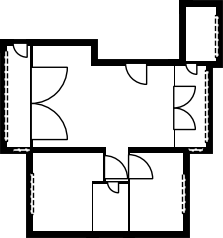}
\includegraphics[height=2.8cm]{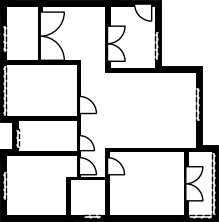}
\includegraphics[height=2.8cm]{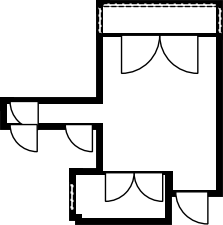}
\includegraphics[height=2.8cm]{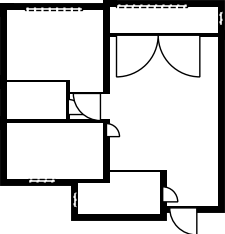}
\includegraphics[height=2.8cm]{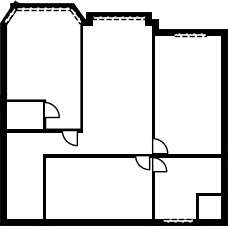}
\includegraphics[height=2.8cm]{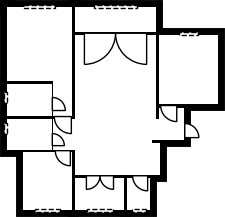}
}{}\\

\rotatebox{90}{\whitetxt{ssssssss}Output}
\jsubfig{
\includegraphics[height=2.8cm]{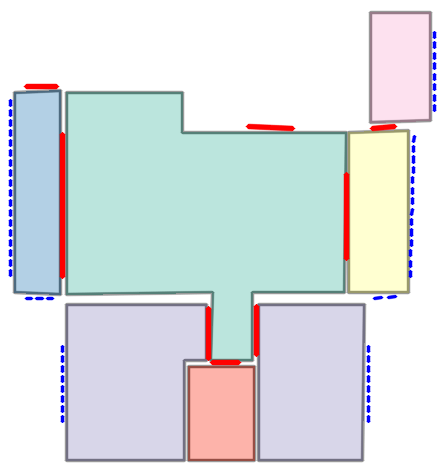}
\includegraphics[height=2.8cm]{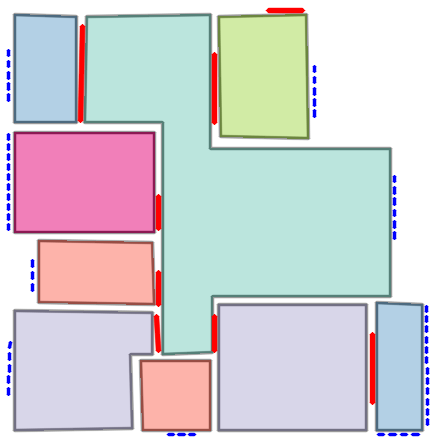}
\includegraphics[height=2.8cm]{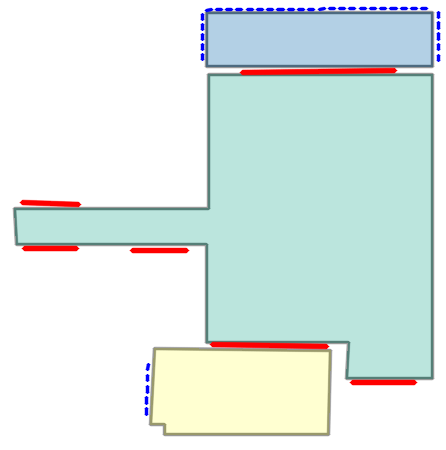}
\includegraphics[height=2.8cm]{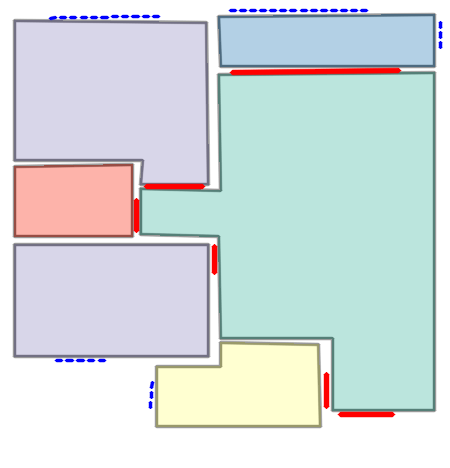}
\includegraphics[height=2.8cm]{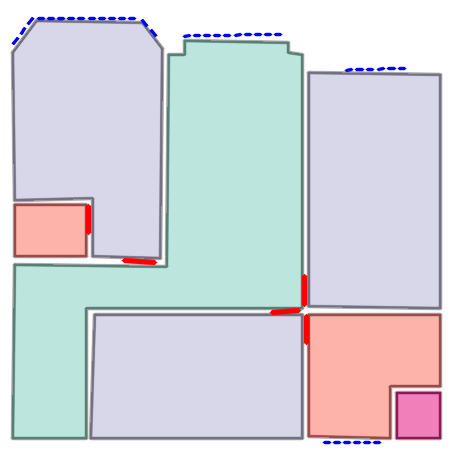}
\includegraphics[height=2.8cm]{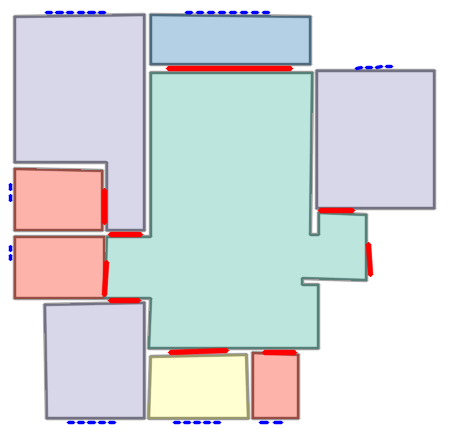}
}{}\\

\vspace{4pt}
\begin{tikzpicture}

\fill[s3d_livingroom, opacity=0.59] (0,1.5) rectangle (1.8,1);
\node[black] at (0.9,1.25) {\small\textbf{Living Room}};

\fill[s3d_kitchen, opacity=0.59] (2.1,1.5) rectangle (3.9,1);
\node[black] at (3,1.25) {\small\textbf{Kitchen}};

\fill[s3d_bedroom, opacity=0.59] (4.2,1.5) rectangle (6.0,1);
\node[black] at (5.1,1.25) {\small\textbf{Bedroom}};

\fill[s3d_bathroom, opacity=0.59] (6.3,1.5) rectangle (8.1,1);
\node[black] at (7.2,1.25) {\small\textbf{Bathroom}};

\fill[s3d_balcony, opacity=0.59] (8.4,1.5) rectangle (10.2,1);
\node[black] at (9.3,1.25) {\small\textbf{Balcony}};

\fill[s3d_corridor, opacity=0.59] (10.5,1.5) rectangle (12.3,1);
\node[black] at (11.4,1.25) {\small\textbf{Corridor}};

\fill[s3d_diningroom, opacity=0.59] (12.6,1.5) rectangle (14.4,1);
\node[black] at (13.5,1.25) {\small\textbf{Dining room}};

\fill[s3d_study, opacity=0.59] (14.7,1.5) rectangle (16.5,1);
\node[black] at (15.6,1.25) {\small\textbf{Study}};

\fill[s3d_studio, opacity=0.59] (0,0.9) rectangle (1.8,0.4);
\node[black] at (0.9,0.65) {\small\textbf{Studio}};

\fill[s3d_storeroom, opacity=0.59] (2.1,0.9) rectangle (3.9,0.4);
\node[black] at (3,0.65) {\small\textbf{Store room}};

\fill[s3d_garden, opacity=0.59] (4.2,0.9) rectangle (6.0,0.4);
\node[black] at (5.1,0.65) {\small\textbf{Garden}};

\fill[s3d_laundryroom, opacity=0.59] (6.3,0.9) rectangle (8.1,0.4);
\node[black] at (7.2,0.65) {\small\textbf{Laundry room}};

\fill[s3d_office, opacity=0.59] (8.4,0.9) rectangle (10.2,0.4);
\node[black] at (9.3,0.65) {\small\textbf{Office}};

\fill[s3d_basement, opacity=0.59] (10.5,0.9) rectangle (12.3,0.4);
\node[black] at (11.4,0.65) {\small\textbf{Basement}};

\fill[s3d_garage, opacity=0.59] (12.6,0.9) rectangle (14.4,0.4);
\node[black] at (13.5,0.65) {\small\textbf{Garage}};

\fill[s3d_misc, opacity=0.59] (14.7,0.9) rectangle (16.5,0.4);
\node[black] at (15.6,0.65) {\small\textbf{Misc}};

\draw[red, very thick, opacity=0.59] (6,0) -- (7,0);
\node[right] at (7.2,0) {\textcolor{red}{Door} };

\draw[blue, dashed, very thick, opacity=0.59] (9,0) -- (10,0);
\node[right] at (10.2,0) {\textcolor{blue}{Window}};

\end{tikzpicture}

\vspace{-4pt}

\caption{\emph{Raster2Seq} reconstruction results on Structured3D-B.}
\label{fig:gallery}
\end{figure*}

\begin{figure*}[!ht]
\centering %

\rotatebox{90}{\whitetxt{sssssssss}Input}
\jsubfig{
\includegraphics[height=2.4cm]{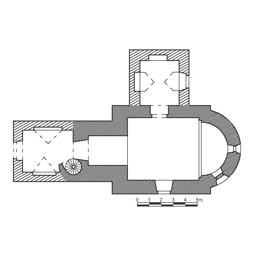}
\includegraphics[height=2.4cm]{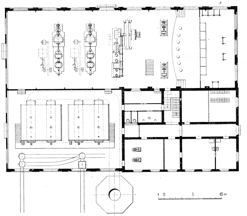}
\includegraphics[height=2.4cm]{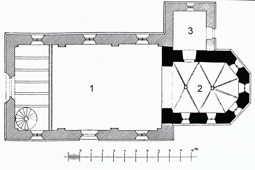}
\includegraphics[height=2.4cm]{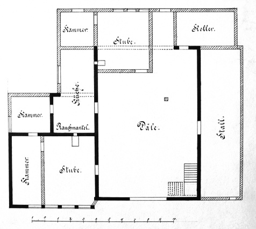}
\includegraphics[height=2.4cm]{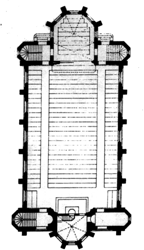}
\includegraphics[height=2.4cm]{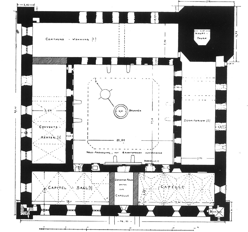}
}{}\\

\rotatebox{90}{\whitetxt{sss}RoomFormer}
\jsubfig{
\includegraphics[height=2.4cm]{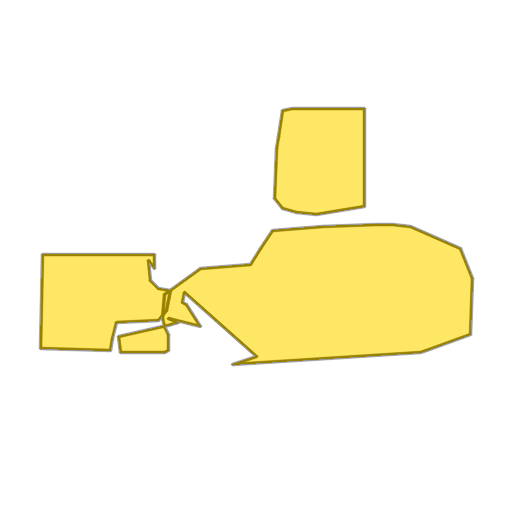}
\includegraphics[height=2.4cm]{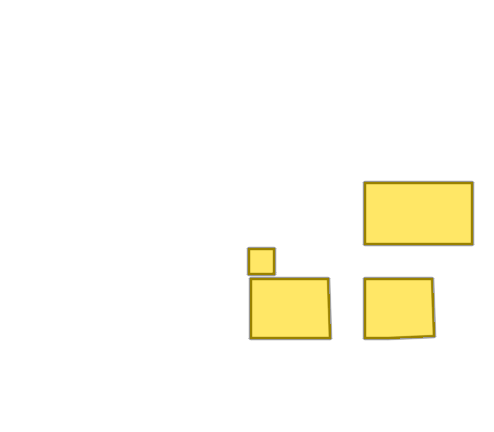}
\includegraphics[height=2.4cm]{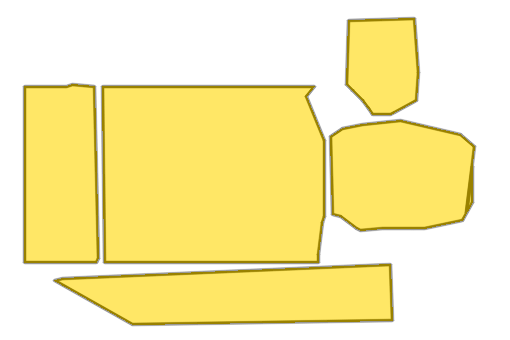}
\includegraphics[height=2.4cm]{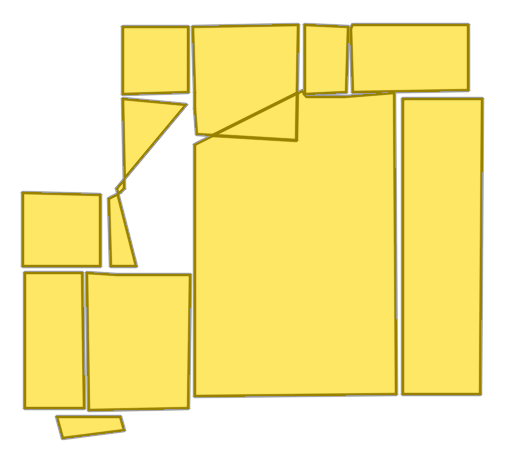}
\includegraphics[height=2.4cm]{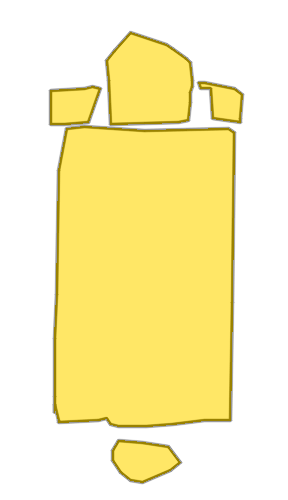}
\includegraphics[height=2.4cm]{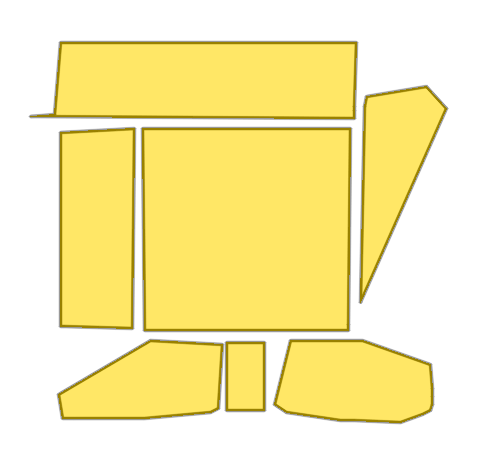}
}{}\\
\rotatebox{90}{\whitetxt{ssssss}Ours}
\jsubfig{
\includegraphics[height=2.4cm]{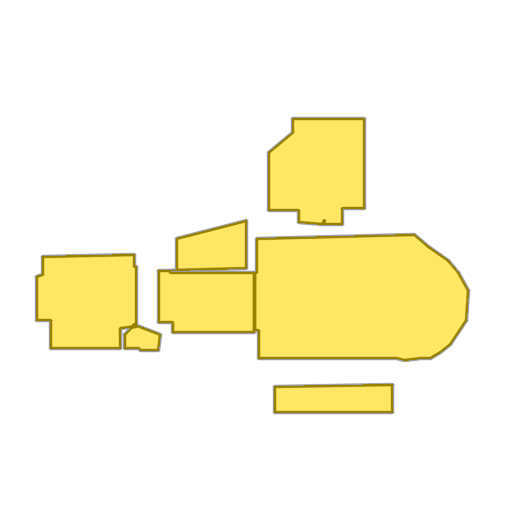}
\includegraphics[height=2.4cm]{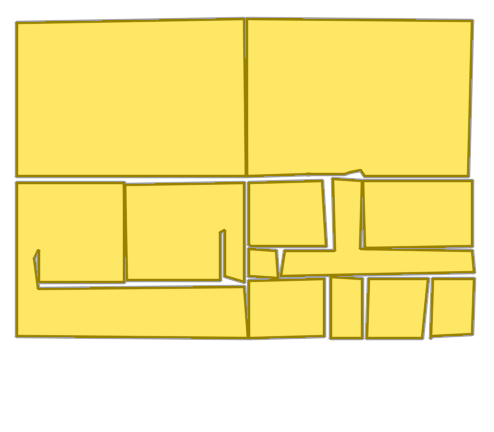}
\includegraphics[height=2.4cm]{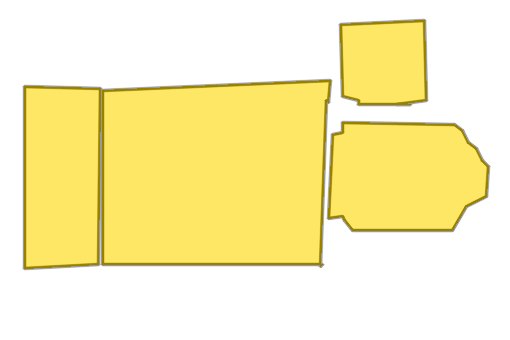}
\includegraphics[height=2.4cm]{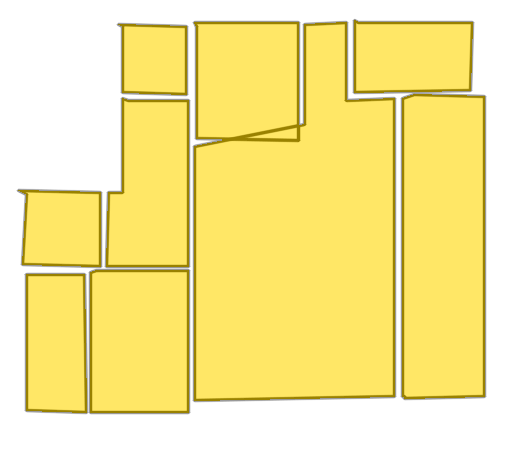}
\includegraphics[height=2.4cm]{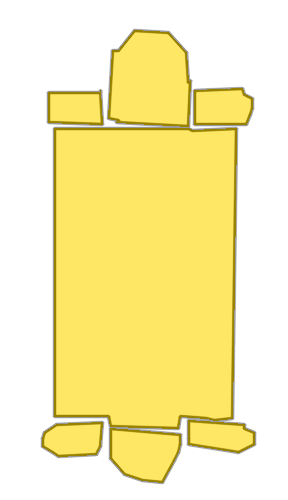}
\includegraphics[height=2.4cm]{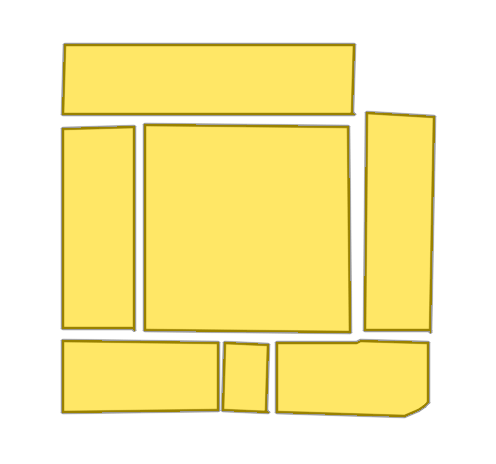}
}{}\\

\vspace{-4pt}

\caption{Qualitative comparison with RoomFormer, over WAFFLE floorplan images (both models are trained on CubiCasa5K). As illustrated above, our model exhibits stronger generalization capabilities over the structures of real-world Internet data. Building names from left-to-right: Church of Saint James, the Greater in Rovny, Teltow Canal Power Station, Church of Saint Nicholas, Imkerhaus, Palais du Louvre, Palmer Mansion.}
\label{fig:gallery_waffle}
\end{figure*}

We provide qualitative examples of our method over Structured3D-B dataset in \cref{fig:gallery} and CubbiCasa5K and on unseen WAFFLE samples in \cref{fig:teaser}. In a meantime, we provide visual comparisons with the RoomFormer model over CubiCasa5K test samples in \cref{fig:vis_comp_cubi} and WAFFLE images in \cref{fig:gallery_waffle}. In both cases, the models are trained over the CubiCasa5K train set. These figures illustrate superior visual quality compared to the RoomFormer baseline. In particular, we observe that the RoomFormer model often yields "short-cut" triangular polygons (\emph{e.g.}, leftmost example in \cref{fig:vis_comp_cubi}), while our model allows for more accurately reconstructing the floorplan's structure.

\begin{figure*}[!ht]
\centering %
\rotatebox{90}{\whitetxt{sss}Input Raster}
\jsubfig{
\includegraphics[height=2.35cm]{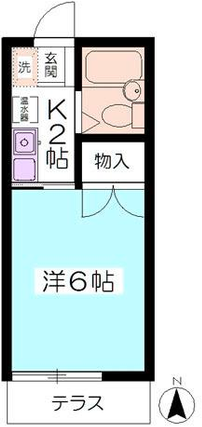}
\includegraphics[height=2.35cm]{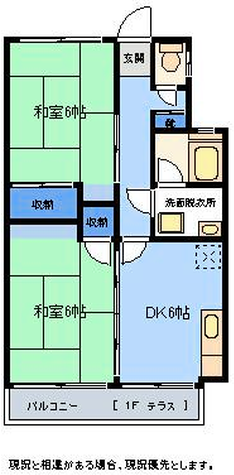}
\includegraphics[height=2.35cm]{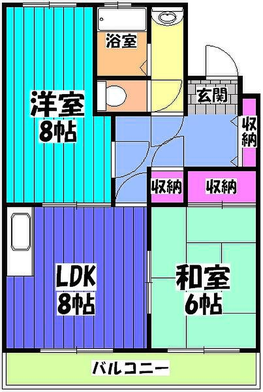}
\includegraphics[height=2.35cm]{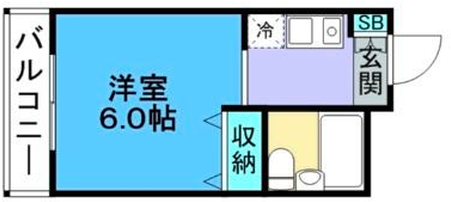}
\includegraphics[height=2.35cm]{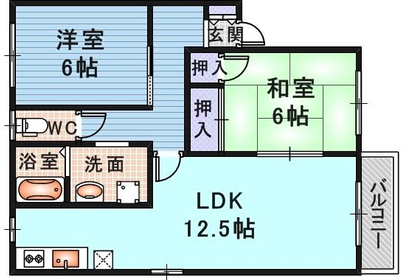}
\includegraphics[height=2.35cm]{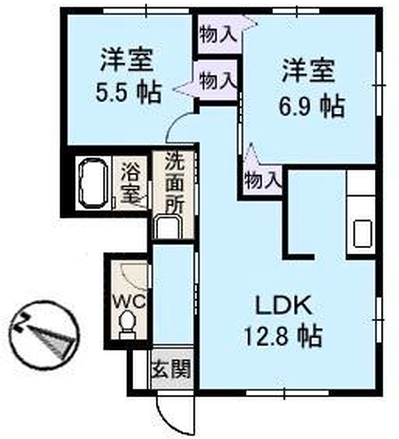}
}{}\\

\rotatebox{90}{\whitetxt{sssssss}GT}
\jsubfig{
\includegraphics[height=2.35cm]{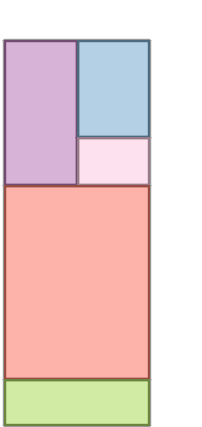}
\includegraphics[height=2.35cm]{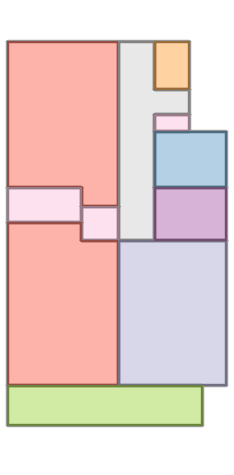}
\includegraphics[height=2.35cm]{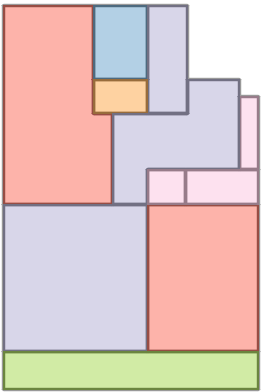}
\includegraphics[height=2.35cm]{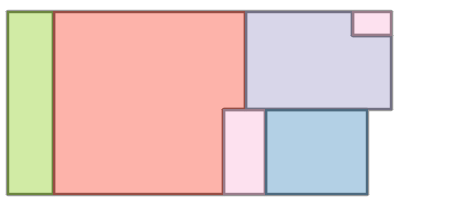}
\includegraphics[height=2.35cm]{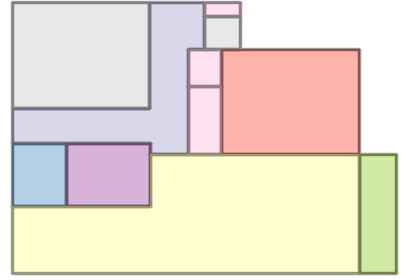}
\includegraphics[height=2.35cm]{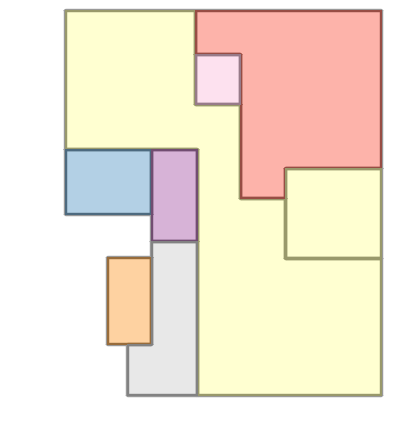}
}{}\\

\rotatebox{90}{\whitetxt{ss}Raster2Graph}
\jsubfig{
\includegraphics[height=2.35cm]{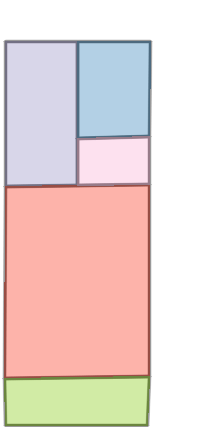}
\includegraphics[height=2.35cm]{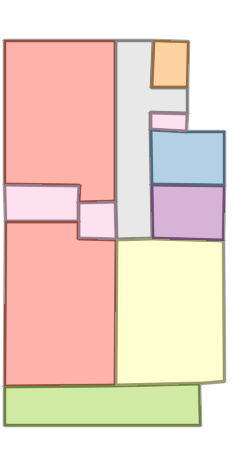}
\includegraphics[height=2.35cm]{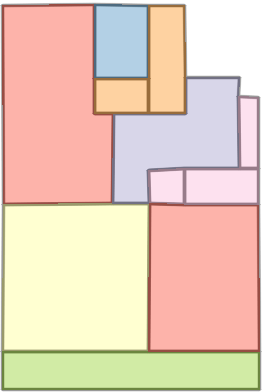}
\includegraphics[height=2.35cm]{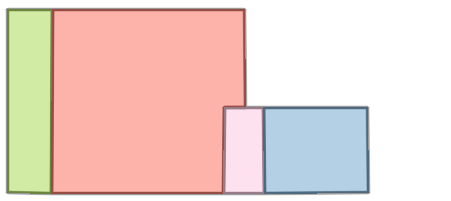}
\includegraphics[height=2.35cm]{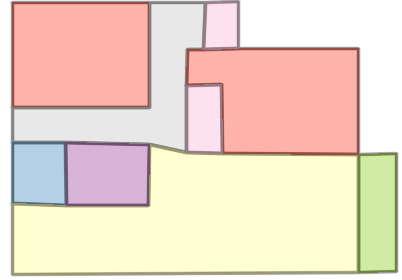}
\includegraphics[height=2.35cm]{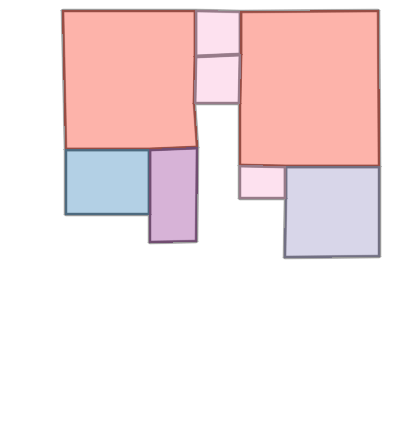}
}{}\\

\rotatebox{90}{\whitetxt{sssssss}Ours}
\jsubfig{
\includegraphics[height=2.35cm]{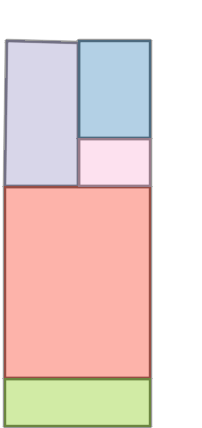}
\includegraphics[height=2.35cm]{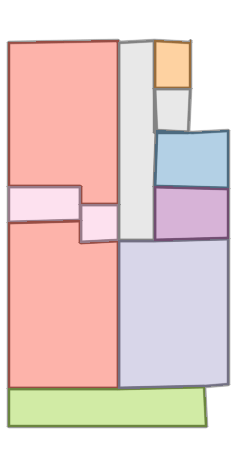}
\includegraphics[height=2.35cm]{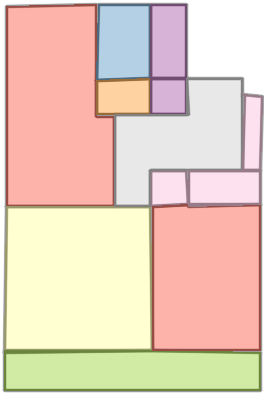}
\includegraphics[height=2.35cm]{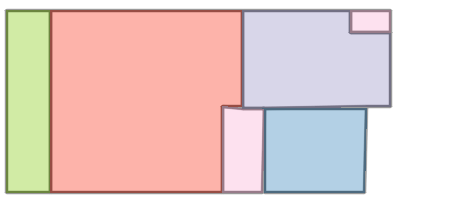}
\includegraphics[height=2.35cm]{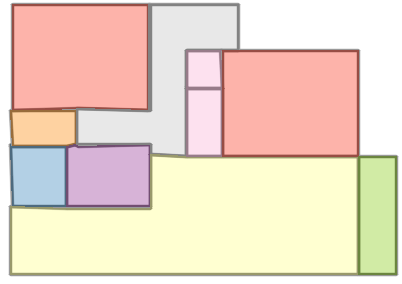}
\includegraphics[height=2.35cm]{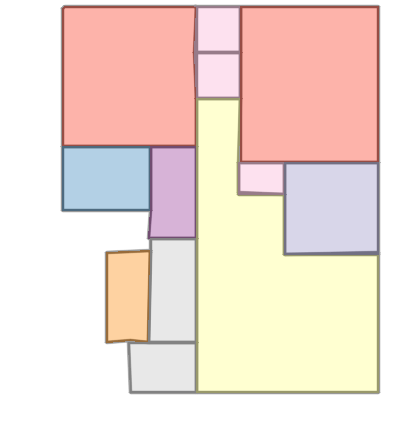}
}{}\\
\vspace{4pt}
\begin{tikzpicture}
\fill[r2g_unknown, opacity=0.59] (0,1.5) rectangle (1.8,1);
\node[black] at (0.9,1.25) {\small\textbf{Unknown}};

\fill[r2g_living_room, opacity=0.59] (2.1,1.5) rectangle (3.9,1);
\node[black] at (3,1.25) {\small\textbf{Living room}};

\fill[r2g_kitchen, opacity=0.59] (4.2,1.5) rectangle (6.0,1);
\node[black] at (5.1,1.25) {\small\textbf{Kitchen}};

\fill[r2g_bedroom, opacity=0.59] (6.3,1.5) rectangle (8.1,1);
\node[black] at (7.2,1.25) {\small\textbf{Bedroom}};

\fill[r2g_bathroom, opacity=0.59] (8.4,1.5) rectangle (10.2,1);
\node[black] at (9.3,1.25) {\small\textbf{Bathroom}};

\fill[r2g_restroom, opacity=0.59] (10.5,1.5) rectangle (12.3,1);
\node[black] at (11.4,1.25) {\small\textbf{Restroom}};

\fill[r2g_balcony, opacity=0.59] (12.6,1.5) rectangle (14.4,1);
\node[black] at (13.5,1.25) {\small\textbf{Balcony}};

\fill[r2g_closet, opacity=0.59] (14.7,1.5) rectangle (16.5,1);
\node[black] at (15.6,1.25) {\small\textbf{Closet}};

\fill[r2g_corridor, opacity=0.59] (4.2,0.75) rectangle (6.0,0.25);
\node[black] at (5.1,0.5) {\small\textbf{Corridor}};

\fill[r2g_washing_room, opacity=0.59] (6.3,0.75) rectangle (8.1,0.25);
\node[black] at (7.2,0.5) {\small\textbf{Washing room}};

\fill[r2g_PS, opacity=0.59] (8.4,0.75) rectangle (10.2,0.25);
\node[black] at (9.3,0.5) {\small\textbf{PS}};

\fill[r2g_outside, opacity=0.59] (10.5,0.75) rectangle (12.3,0.25);
\node[black] at (11.4,0.5) {\small\textbf{Outside}};

\end{tikzpicture}

\vspace{2pt}

\caption{Qualitative comparison with Raster2Graph on their dataset. Our method achieves more accurate floorplan reconstructions in comparison to their model, which often produces incomplete results.}

\label{fig:vis_comp_r2g}
\end{figure*}

In \cref{fig:vis_comp_r2g}, we compare our method with Raster2Graph on their dataset. As clearly seen, our method achieves superior reconstruction quality compared to other counterparts. Notably, Raster2Graph often fails to recover complete floorplan structures, while our approach remains robust across diverse layouts. 

For more qualitative results, please refer to \cref{sec:more_results}.

\subsection{Ablations}
\label{sec:ablations}

\begin{table}[t]
\centering
\small
\caption{Ablation studies, evaluating the effect of our \emph{FeatFusion} mechanism, the learnable tokens, and performing a left-to-right ordering of the  polygons during training, over the \emph{Structure3D-B} dataset.}
\label{tab:network_abl}
\resizebox{\linewidth}{!}{
\begin{tabular}{cccccc}
\toprule
FeatFusion & Anchor & Ordering  & Room F1 & Corner F1 & Angle F1 \\
\midrule
& & & 94.1 & 91.1 & 82.0 \\
\checkmark & & &  96.3 & 93.7 & 82.6 \\
\checkmark & \checkmark & &  97.4 & 95.3 & 86.0 \\
\rowcolor{pink!40} \checkmark & \checkmark & \checkmark & 99.6 & 98.3 & 92.7 \\

\bottomrule
\end{tabular}
}
\vspace{-3pt}
\end{table}

We conduct extensive ablations, evaluating the effect of various components in our framework, on the Structure3D-B dataset.   
For simplicity of the ablations, all models are trained for 1,350 epochs.
As a result, the reported performance in this section does not reflect the best results our model is capable of achieving. 

\Cref{tab:network_abl} 
highlights the impact of three key components—FeatFusion, which merges polygon and image features, the learnable anchors, and the left-to-right ordering of polygons in the sequence—on floorplan reconstruction performance; qualitative results over a single sample are provided in Figure \ref{fig:abl_vis}. Overall, each proposed component contributes meaningfully to the model performance. In particular, learnable anchors provide a significant boost in model performance, while integrating polygon generation ordering yields the best performance. Together, these findings confirm that every component plays a distinct role in achieving precise floorplan reconstruction. %

Additional ablations, quantifying the effect of the sequence length, quantization resolution, and the coefficient of the coordinate loss, are reported in \cref{sec:more_abl}.

\subsection{Limitations}
\begin{figure}[t]
  \centering

\jsubfig{
\includegraphics[height=1.7cm]{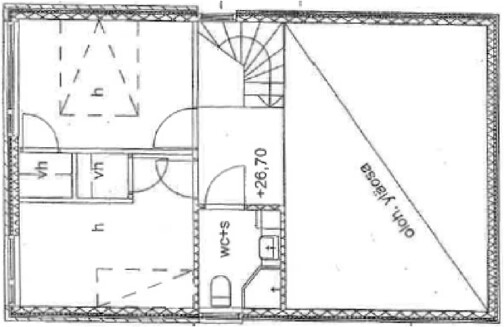}}{Input}
\jsubfig{
\includegraphics[height=1.7cm]{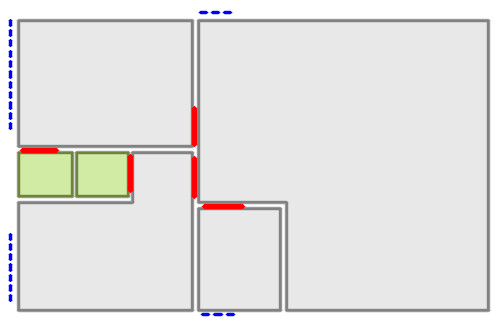}}{GT}
\jsubfig{
\includegraphics[height=1.7cm]{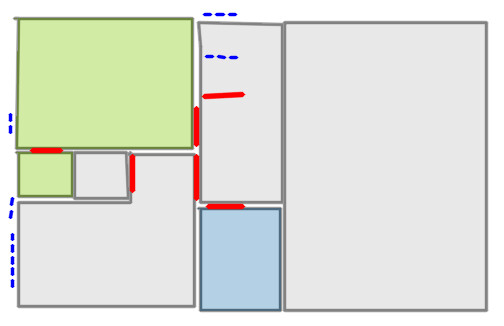}
}{Prediction}
\vspace{-8pt}
  \caption{Limitation example, illustrating that our method may generate windows and doors inside rooms. \textcolor{red}{Red line} denotes a door and a \textcolor{blue} {dashed line} denotes a window. }%
  \label{fig:limitations}
\end{figure}

\label{sec:limitations}
While our approach achieves strong performance in both geometric reconstruction and generalization, we find that performance over less prevalent semantic structures such as doors and windows can be further refined.
As shown in \cref{fig:limitations}, the model occasionally fails to accurately localize windows and doors, resulting in artifacts such as cross-over windows. 
Future work can investigate tailored architectural changes to better accommodate other element types, potentially modeling these elements separately from room entities.

\section{Conclusion}
\label{sec:conc}

In this work, we proposed to frame raster-to-vector floorplan conversion as a sequence-to-sequence task. We introduced a framework that predicts vectorized representation as labeled polygon sequence. The driving mechanism of our framework is an anchor-based autorgressive decoder, that learns to predict the next corner token conditioned on previously generated corners. Technically, our decoder introduces several architectural components, such as the integration of learnable anchors and the \emph{FeatFusion} concatenation operation, enabling for effectively learning the generation of complex polygon sequences. Our experiments demonstrate that our approach outperforms prior work targeting similar tasks across various geometric and semantic metrics. 

\methodname{} demonstrates promising generalization performance to \emph{in-the-wild} Internet data, representing a step towards the goal of modeling historical buildings, defined by hand-drawn floorplans. Future work can incorporate mechanisms that further improve results on out-of-distribution data, such as appearance-based augmentations. In particular, combining our system with open-vocabulary  predictions could potentially allow for reconstructing the rich semantics reflected in diverse real-world floorplans. 
\change{Another promising extension is to explicitly incorporate semantic conditions during inference, for example through a lightweight condition adapter. This would enable a variety of controls, such as using input room semantic labels to guide the decoding toward generating desired room coordinates.}
More broadly, the ability to recover accurate vectorized floorplan representations will likely become increasingly important as generative models grow more powerful, enabling controllable downstream applications, such as floorplan-guided 3D generation of large architectural scenes, that extend beyond traditional analysis and editing settings.

\newcommand{\figheight}{2.6cm}

\begin{figure*}[!ht]
\centering %
\rotatebox{90}{\whitetxt{sssssssss}Input}
\jsubfig{
\includegraphics[height=\figheight]{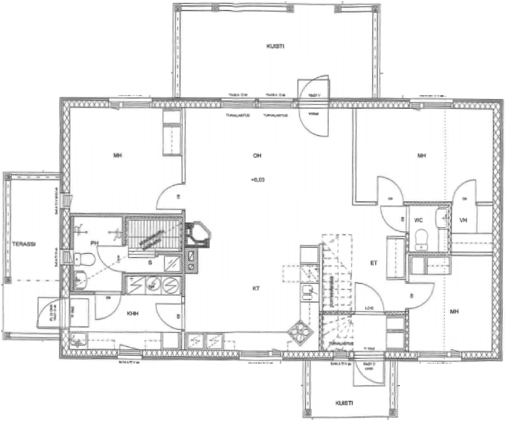}
\includegraphics[height=\figheight]{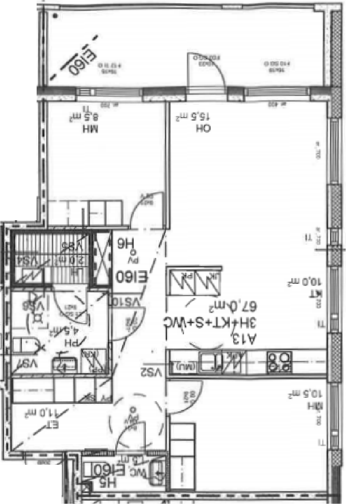}
\includegraphics[height=\figheight]{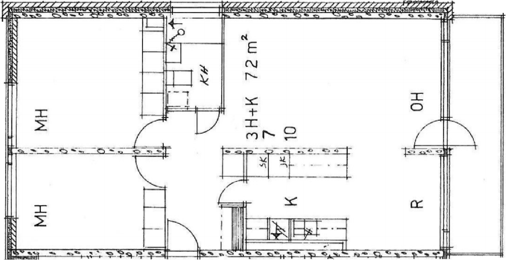}

\includegraphics[height=\figheight]{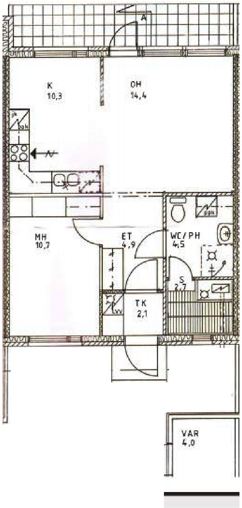}
\includegraphics[height=\figheight]{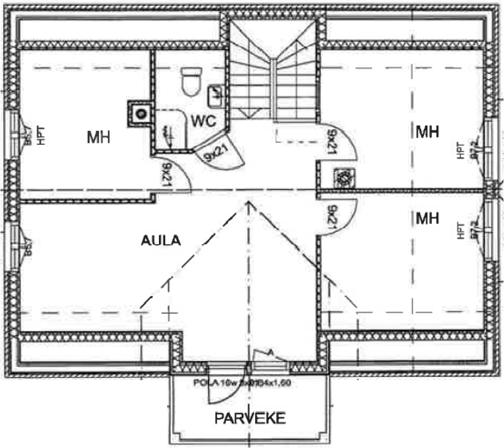}
\includegraphics[height=\figheight]{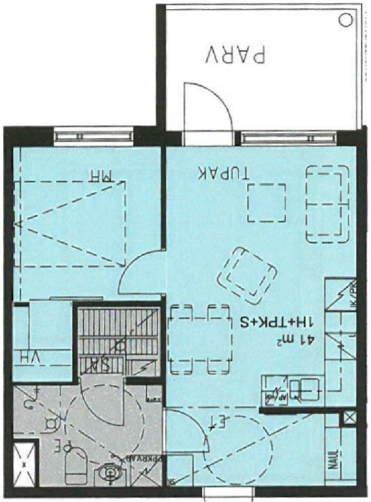}

}{}\\

\rotatebox{90}{\whitetxt{ssssssssss}GT}
\jsubfig{
\includegraphics[height=2.6cm]{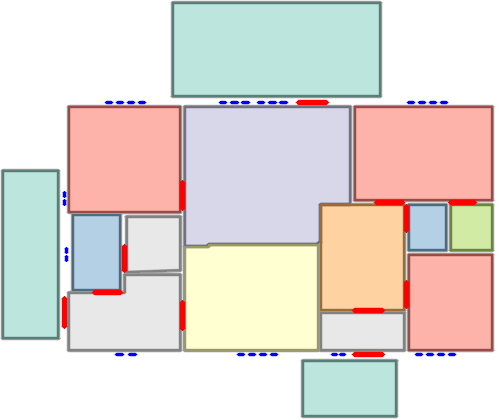}
\includegraphics[height=2.6cm]{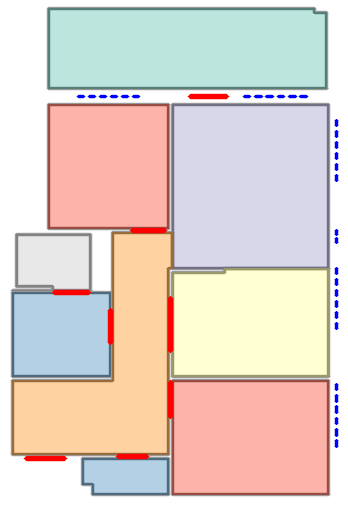}
\includegraphics[height=2.6cm]{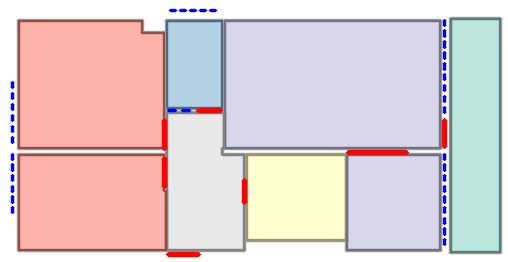}

\includegraphics[height=2.6cm]{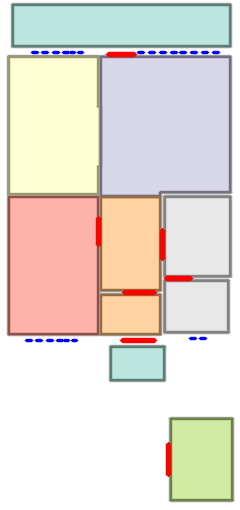}
\includegraphics[height=2.6cm]{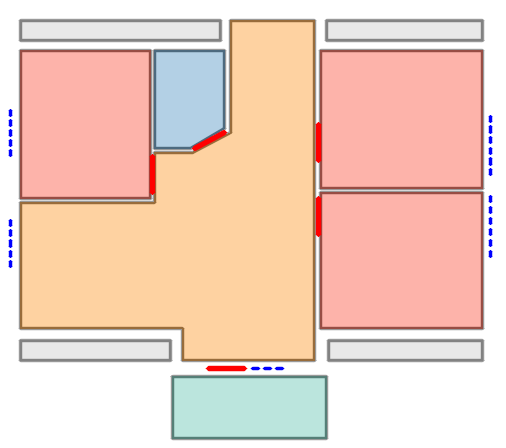}
\includegraphics[height=2.6cm]{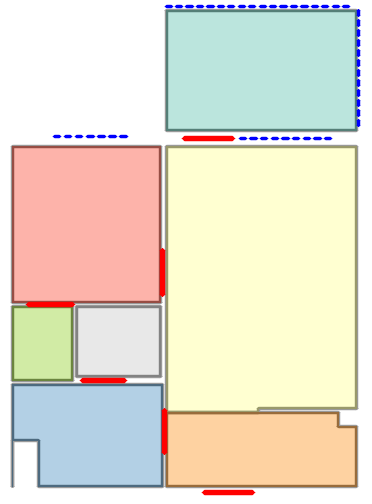}

}{}\\

\rotatebox{90}{\whitetxt{sssss}RoomFormer}
\jsubfig{
\includegraphics[height=2.6cm]{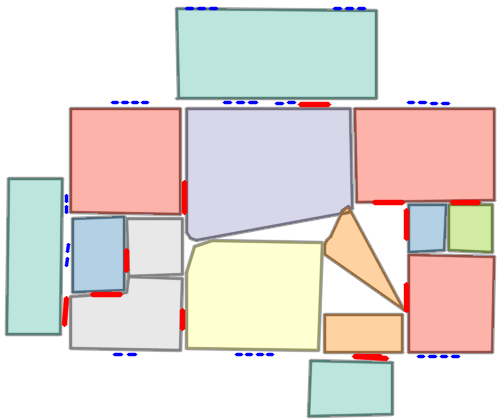}
\includegraphics[height=2.6cm]{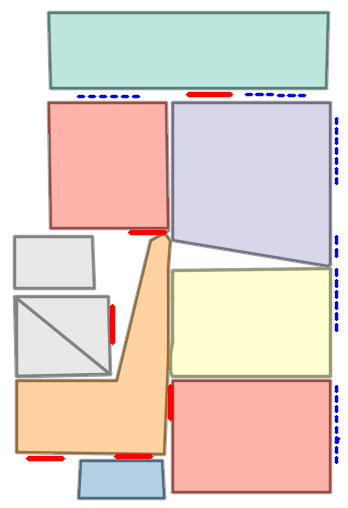}
\includegraphics[height=2.6cm]{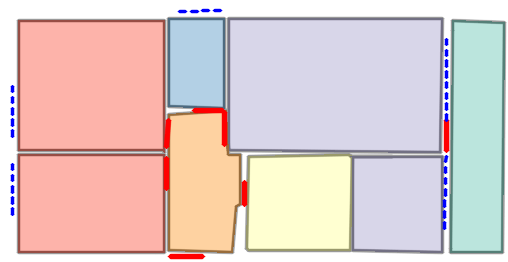}

\includegraphics[height=2.6cm]{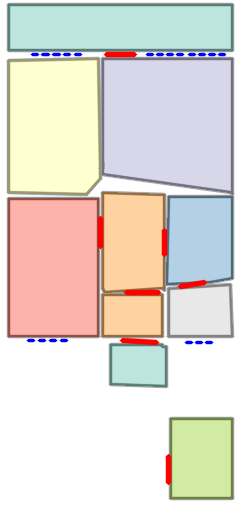}
\includegraphics[height=2.6cm]{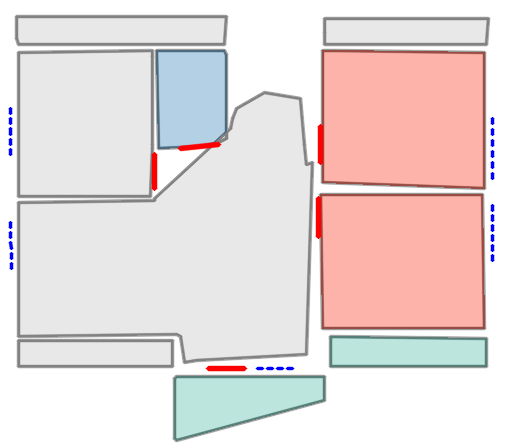}
\includegraphics[height=2.6cm]{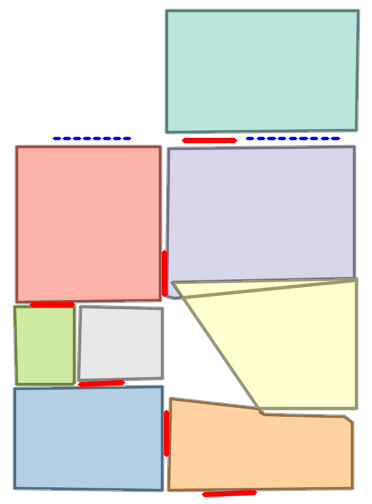}

}{}\\

\rotatebox{90}{\whitetxt{sssssssss}Ours}
\jsubfig{
\includegraphics[height=2.6cm]{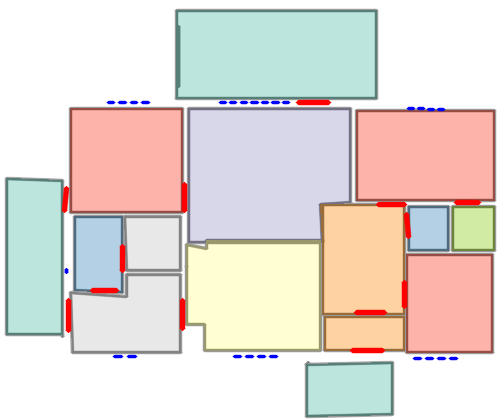}
\includegraphics[height=2.6cm]{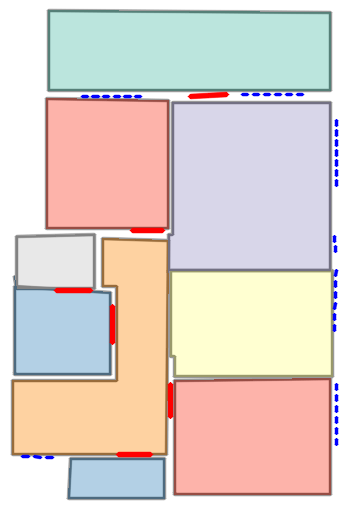}
\includegraphics[height=2.6cm]{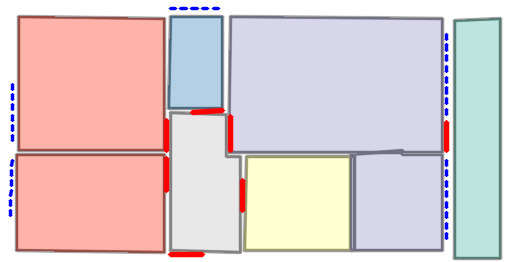}

\includegraphics[height=2.6cm]{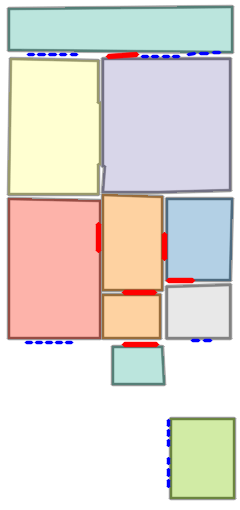}
\includegraphics[height=2.6cm]{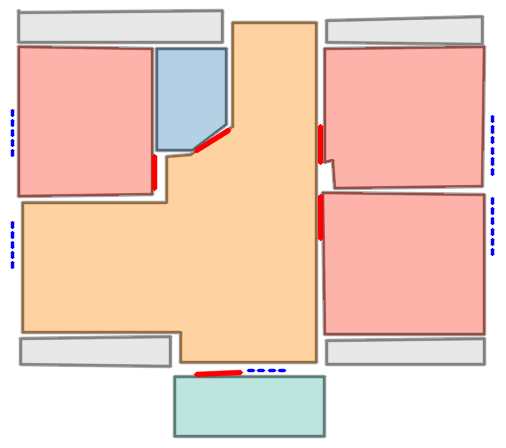}
\includegraphics[height=2.6cm]{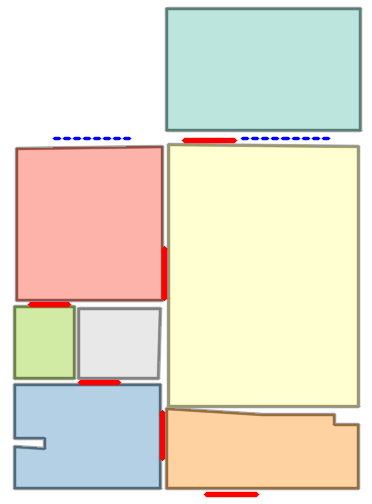}

}{}\\
\vspace{4pt}
\begin{tikzpicture}

\fill[outdoor, opacity=0.59] (0,1) rectangle (1.4,0.5);
\node[black] at (0.7,0.75) {\small\textbf{Outdoor}};

\fill[kitchen, opacity=0.59] (1.8,1) rectangle (3.2,0.5);
\node[black] at (2.5,0.75) {\small\textbf{Kitchen}};

\fill[livingroom, opacity=0.59] (3.6,1) rectangle (5.0,0.5);
\node[black] at (4.3,0.75) {\small\textbf{Living room}};

\fill[bedroom, opacity=0.59] (5.4,1) rectangle (6.8,0.5);
\node[black] at (6.1,0.75) {\small\textbf{Bed room}};

\fill[bath, opacity=0.59] (7.2,1) rectangle (8.6,0.5);
\node[black] at (7.9,0.75) {\small\textbf{Bath}};

\fill[entry, opacity=0.59] (9.0,1) rectangle (10.4,0.5);
\node[black] at (9.7,0.75) {\small\textbf{Entry}};

\fill[storage, opacity=0.59] (10.8,1) rectangle (12.2,0.5);
\node[black] at (11.5,0.75) {\small\textbf{Storage}};

\fill[garage, opacity=0.59] (12.6,1) rectangle (14.0,0.5);
\node[black] at (13.3,0.75) {\small\textbf{Garage}};

\fill[undefined, opacity=0.59] (14.4,1) rectangle (15.8,0.5);
\node[black] at (15.1,0.75) {\small\textbf{Undefined}};

\draw[red, very thick, opacity=0.59] (6,0) -- (7,0);
\node[right] at (7.2,0) {\textcolor{red}{Door} };

\draw[blue, dashed, very thick, opacity=0.59] (9,0) -- (10,0);
\node[right] at (10.2,0) {\textcolor{blue}{Window}};

\end{tikzpicture}

\vspace{4pt}

\caption{Qualitative results on the \cubi dataset, comparing Raster2Seq to the RoomFormer model.
}
\label{fig:vis_comp_cubi}
\end{figure*}

\clearpage

\bibliographystyle{ACM-Reference-Format}
\bibliography{main}      

\clearpage
\appendix

\section{Data preparation}
\label{sec:data_prep}
\subsection{Structured3D}
\begin{figure}[t]
  \centering
  \begin{subfigure}[b]{0.15\textwidth}
    \includegraphics[width=\textwidth]{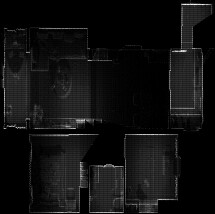}
    \caption{Density map}
  \end{subfigure}
  \hfill
  \begin{subfigure}[b]{0.15\textwidth}
    \includegraphics[width=\textwidth]{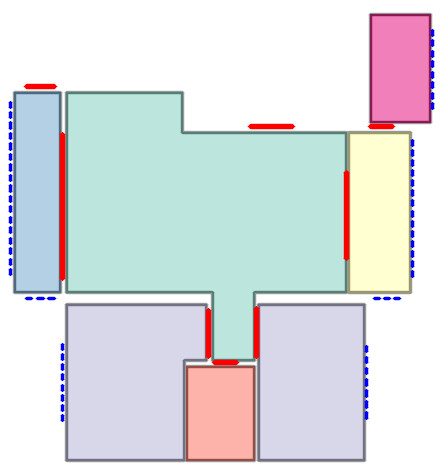}
    \caption{Floorplan map}
  \end{subfigure}
  \hfill
  \begin{subfigure}[b]{0.15\textwidth}
    \includegraphics[width=\textwidth]{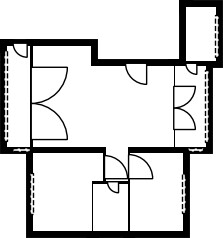}
    \caption{Output binary image}
  \end{subfigure}
  \vspace{-12pt}
  \caption{Binary image conversion on Structured3D data. Using the annotated floorplan map, we generate a binary image shown in the last column. Note that the density map on the left is only shown as reference (it is not utilized in the conversion).}
  \label{fig:s3d_processing}
\end{figure}

Since Structured3D \cite{zheng2020structured3d,stekovic2021montefloor} is provided as density maps projected from 3D point clouds, we convert these maps into RGB-format floorplan images using the accompanying data annotations (see \cref{fig:s3d_processing}) to better mimic the appearance of standard RGB floorplans, typically in black-and-white format. Since the raster images in the Structured3D dataset are synthetically generated by a rendering engine, they differ substantially from real-world images in appearance. Meanwhile, the data statistics and annotations after being converted to binary format are preserved without modification. 

\subsection{CubiCasa5K}
\begin{figure}[t]
\centering
\includegraphics[width=0.95\linewidth]{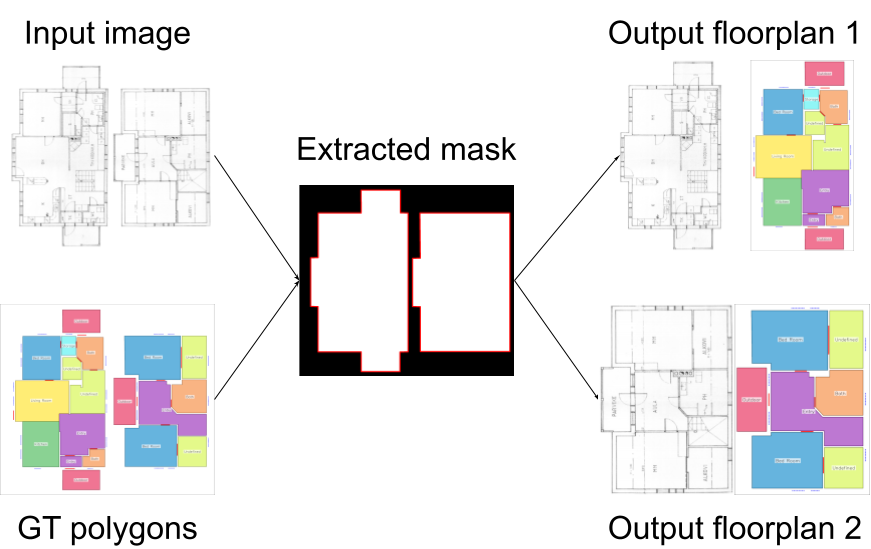}
\vspace{-8pt}
\caption{The process of extracting floorplan instances from an image.}
\label{fig:cubi_prep}
\end{figure}

CubiCasa5K \cite{kalervo2019cubicasa5k} is originally proposed for segmentation task where the given annotations are pixel-wise segmentation maps. We first select 10 semantic room classes, namely Outdoor, Wall, Kitchen, Living Room, Bed Room, Bath, Entry, Railing, Storage, Garage, Undefined, along with two additional classes (Window and Door). We convert the segmentation maps of these corresponding classes to polygons which are used as the real-value corners for each room. Since each image may contain more than one floorplan instances, we further process to separate out individual instance which is saved into a separate file (see \cref{fig:cubi_prep}). Specifically, we take the closed-contours on the binary mask which indicate the foreground/regions of corresponding floorplan instances. After separating out individual instances, we also shift the coordinate of corners based on the bounding box covering floorplan regions. 

\subsection{Raster2Graph}

We preprocess the data following Raster2Graph's codebase \cite{hu2024r2g}.

\section{Additional Implementation Details}
\label{sec:method_details}
\minihead{Image Feature Extractor}. We 
instantiate the image feature extractor with a ResNet-50 backbone \cite{he2016deep} followed by a transformer encoder \cite{zhu2021deformable}, following the feature extraction module in \cite{yue2023roomformer}. This block produces the feature vector $f_{img}$ which serves as input to the autoregressive decoder for polygon sequence generation. Followed \cite{yue2023roomformer,liu2024polyroom,hu2024r2g}, we initialize the ResNet backbone with ImageNet pretrained weights, then fully finetune the entire network (both backbone and autoregressive decoder) end-to-end.

\minihead{Bilinear Quantizer.} In order to train our model on the proposed input sequence, we need to find a suitable way to convert these continuous coordinates to corresponding discrete embeddings. Followed \cite{he2017mask,liu2023polyformer}, we discritize the 2D contiguous coordinates into 1D embedding space, with an introduction of a learnable codebook of size $\mathbb{R}^{H_b \times W_b\times D}$ where $H_b \times W_b$ is number of quantization bins and $D$ is embedding dimension. In detail, given a 2D coordinate $(x,y)$, it is applied floor ($\lfloor . \rfloor$) and celing operations ($\lceil . \rceil$) to produce precise embeddings for its 4 neighbor points in the 2D grid. Formally, the final embedding $e_{x, y}$ is obtained by a bilinear interpolation to get the exact values of input coordinates: 

\begin{equation}
\begin{aligned}
e_{x, y} =\; & (\lceil x \rceil - x)(\lceil y \rceil - y) \cdot e_{\lfloor x \rfloor, \lfloor y \rfloor} 
+ (x - \lfloor x \rfloor)(\lceil y \rceil - y) \cdot e_{\lceil x \rceil, \lfloor y \rfloor} \\
&+ (\lceil x \rceil - x)(y - \lfloor y \rfloor) \cdot e_{\lfloor x \rfloor, \lceil y \rceil} 
+ (x - \lfloor x \rfloor)(y - \lfloor y \rfloor) \cdot e_{\lceil x \rceil, \lceil y \rceil}
\end{aligned}
\label{eq:quant}
\end{equation}

These quantized values are used as input to the decoder layer, alongside the encoded image features to regress the contiguous coordinate values.

\begin{algorithm}[t]
\caption{Sequential Corner Generation}
\label{alg:generation}
\KwIn{Input image $I$, model $\text{Model}(\cdot)$, maximum steps $L_{\max}$}
Initialize sequence $S \leftarrow [\,]$ \\
Initialize temporary sequence $s \leftarrow [\,]$ \\
Initialize input state $c_0 \leftarrow [0,0,p=\emptyset]$; $q_0 \leftarrow \texttt{<BOS>}$ \\
\For{$l = 1$ \KwTo $L_{\max}$}{
    \tcp{Predict next token}
    $\hat{c}_l, \hat{q}_l \leftarrow \text{Model}(I, c_{0:l-1}, q_{0:l-1})$ \\
    $\tilde{q}_l \leftarrow \arg\max(\hat{q}_l)$ \tcp*{Predicted token type}
    $\tilde{p}_l \leftarrow \arg\max(\hat{c}_l.p)$ \tcp*{Predicted semantic class}
    \If{$\tilde{q}_l = \texttt{<EOS>}$}{
        \textbf{break}
    }
    \eIf{$\tilde{q}_l = \texttt{<SEP>}$}{
        Append $s$ to $S$ \tcp*{Save current room seq}
        $s \leftarrow []$ \\
    }{
        \tcp{Save corner and label}
        Append [$\hat{c}_l.x$, $\hat{c}_l.y$, $\tilde{p}_l$] to $s$ 
    }
    $c_l \leftarrow \hat{c}_l$; $q_l \leftarrow \hat{q}_l$
}
\Return $S$
\end{algorithm}

\minihead{Model configs.} The model consists of 12 layers in total, evenly divided between the encoder and decoder. Since the input sequence length varies across images, we follow standard language modeling practice \cite{mikolov2010rnn,vaswani2017attention} by padding each input to a fixed length of $L$ during training.

\begin{table}[t]
\centering
\small
\caption{Ablation on coordinate quantization resolution.}
\label{tab:quant_res}
\begin{tabular}{lccccccccc}
\toprule
Num bins &  Room F1 & Corner F1 & Angle F1 \\
\midrule
16x16 & 95.8 & 93.4 & 83.4 \\
\rowcolor{pink!40} 32x32 & 96.3 & 93.7 & 82.6 \\
64x64 & 94.4 & 91.6 & 82.8 \\
\bottomrule
\end{tabular}
\end{table}

\begin{table}[t]
\centering
\small
\caption{Effect of input sequence length on floorplan generation performance.}
\label{tab:seq_length}
\begin{tabular}{lccc}
\toprule
Seq length & Room F1 & Corner F1 & Angle F1 \\
\midrule
256 & 92.1 & 88.0 & 73.1 \\
\rowcolor{pink!40} 512 & 96.3 & 93.7 & 82.6 \\
1024 & 95.2 & 92.4 & 80.0 \\
\bottomrule
\end{tabular}

\end{table}

\begin{table}[!ht]
\centering
\small
\caption{Ablation on the order of window and doors in the labeled polygon sequence, comparing to a version that treats them like rooms within the standard left-to-right ordering (top) vs. our method which appends them after the room sequence (bottom). This experiment is conducted on Structure3D-B and evaluated at 1749 epoch.}
\label{tab:wd_order}
\resizebox{\linewidth}{!}{
\begin{tabular}{lccc}
\toprule
 &  Room F1 & Corner F1 & Angle F1 \\
\midrule
In-between (Standard left-to-right) & 97.7 & 95.4 & 85.1 \\

\rowcolor{pink!40} Post-room (Appending at the end) & 98.4 &	96.4 &	88.7 \\
\bottomrule
\end{tabular}
}
\end{table}

In the decoder, we introduce a causal attention layer with our proposed post-fusion mechanism, enabling the model to generate outputs autoregressively. This layer is placed before the deformable attention module. The model configuration generally follow the former where we keep model dimension at 256 channels. The number of learnable anchors is also set to 512 to match the input length. More details of model configs are shown in \cref{tab:model_config}.

Additionally, to accelerate the generation, we introduce KV caches \cite{pope2023efficiently} for the decoder where the keys and values of previous tokens are stored in the cache, eliminating the cost of recomputing them in every loop.

\minihead{Learnable Anchors.} We use learnable anchors for all tokens—both corners and special tokens like <SEP> and <EOS>—with anchors being fully aware of token types in the sequence. Note that we ignore the coordinate loss for special token positions, which are only supervised by token-type classification, allowing the model to dynamically adjust anchor values based on gradients. At inference, the coordinate values of special tokens are ignored—only their token types are used to identify the end of the current room sequence or the end of generation (see Algorithm \ref{alg:generation}).

\section{Training details}
\label{sec:training_details}

\minihead{Ours.} For training, we also adopt the same hyper-parameters as RoomFormer, detailed in \cref{tab:hyperparams}. For pretraining, we train our model for 1400 epochs on Structured3D and 500 epochs on CubiCasa5K. During fine-tuning with semantic loss, we continue training for an additional 450 epochs on Structured3D and 450 epochs on CubiCasa5K. We set $\lambda_{coord} = 20$ and $\lambda_{sem} = 1$ by default while $\lambda_{token}$ is varied for each dataset. All experiments are conducted on a single NVIDIA A6000 GPU and take approximately 1-2 days to complete.

As detailed in the paper (and our ablations), we find that during fine-tuning, the order of coordinates in the sequence matter. Rather than naively inserting window and door coordinates, appending them to the end of the sequence leads to a substantial improvement of model performance. We hypothesize that during pretraining, the model was exposed only to room coordinates. Based on this observation, we append window and door coordinates at the end of the sequence—after the room polygons—to ensure a smooth transition in labeled polygon representation from the pretraining to the finetuning stage (see \cref{tab:wd_order}).

\minihead{FRI-Net.} The procedure for generating GT occupancy maps for training follows the FRI-Net implementation and no point cloud is needed. These binary maps are generated from GT room polygons (1 for inside, 0 for outside) and used as supervision.

\section{Labeled Polygon Sequence Generation}
\label{sec:inference}
Algorithm~\ref{alg:generation} illustrates the corner generation process of our method (inference), which iteratively predicts the next point in the sequence.

\begin{table}[t]
\centering
\caption{Cross-evaluation of floorplan interior segmentation performance on the \emph{WAFFLE} test set. \textdagger{Reported in \cite{ganon2025waffle}}.}
\label{tab:waffle}
\begin{tabular}{lccc}
\toprule
\textbf{Method} & \textbf{IoU} & \textbf{Prec} & \textbf{Rec} \\
\midrule
CubiCasa5K pretrained\textdagger & 46.1 & 79.9 & 52.1 \\
FRI-Net & 56.7 & 63.4 & 84.2 \\
RoomFormer & 60.5 & 65.7 & 88.3 \\
Ours & \textbf{73.9} & \textbf{81.6} & \textbf{88.6} \\
\bottomrule
\end{tabular}
\end{table}

\begin{table}[t]
    \centering
    \small
    \caption{Training hyper-parameters and training time.}
    \label{tab:hyperparams}
    \resizebox{\linewidth}{!}{
        \begin{tabular}{lcccc}
        \toprule
                      & Structured3D-3DScans & Structured3D-BW & CubiCasa5K & Raster2Graph \\ \midrule
        Learning rate & $2\text{e-4}$ & $2\text{e-4}$ & $2\text{e-4}$ & $2\text{e-4}$ \\
        AdamW optimizer ($\beta_1,\beta_2$)  & 0.9, 0.999    & 0.9, 0.999 & 0.9, 0.999 & 0.9, 0.999     \\
        Input image channels & 1 & 3 & 3 & 3 \\
        Dropout      & 0.1           & 0.1      & 0.1 & 0.1             \\
        Max-grad-norm & 0.1             & 0.1              & 0.1 & 0.1             \\
        $\lambda_{coord}$ & 20 & 20 & 20 & 20 \\
        $\lambda_{token}$ & 1 & 1 & 5 & 5 \\
       
        GPUs (A6000)         & 1            &   1             & 1       & 2      \\
        \midrule
        \multicolumn{5}{c}{\textbf{Pretraining}} \\
        \midrule
        Epochs        & 500          & 1400            &    500      &  850   \\
        Batch size per GPU    & 32            &  32 &   64 & 64            \\
        Train hours    &   5.0h      &  14.3h              &     19.4h  & 21.1h      \\ 
        \midrule
        \multicolumn{5}{c}{\textbf{Semantic Finetuning}} \\
        \midrule
        Epochs        & 700          & 450            &   450           & 550 \\
        Batch size per GPU    & 32            &  32 & 56    &  56        \\
        Train hours    & 6.9h        & 4.6h               &   16h       & 13.6h  \\ 
         $\lambda_{sem}$ & 1 & 1 & 1 & 1 \\
        
        \bottomrule
        \end{tabular}
        }
\end{table}

\begin{table}[t]
      \centering
      \small
        \caption{Model config}
        \label{tab:model_config}
        \begin{tabular}{lc}
        \toprule
        Config                  & \multicolumn{1}{l}{Value} \\ \midrule
        No. encoder layers                   & 6                         \\
        No. decoder layers                   & 6                         \\
        Hidden size             & 256                      \\
        FFN hidden size & 1024 \\
        No. attention heads & 8 \\
        No. sampling points per deformable attention head & 4 \\

        \bottomrule
        \end{tabular}
        
    \end{table}

\section{Additional Results}
\label{sec:more_results}

\begin{figure*}[t]
\centering %
\rotatebox{90}{\whitetxt{sssssssss}Input}
\jsubfig{
\includegraphics[height=2.8cm]{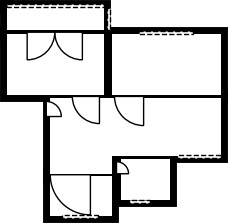}
\includegraphics[height=2.8cm]{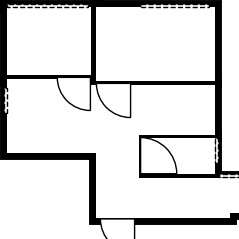}
\includegraphics[height=2.8cm]{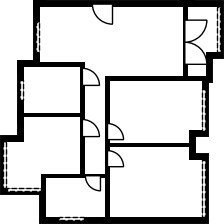}
\includegraphics[height=2.8cm]{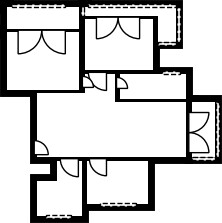}
\includegraphics[height=2.8cm]{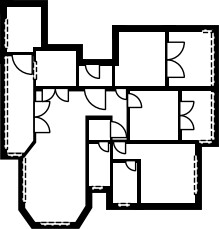}
\includegraphics[height=2.8cm]{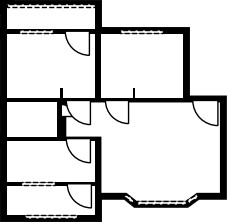}
}{}\\

\rotatebox{90}{\whitetxt{ssssssss}Output}
\jsubfig{
\includegraphics[height=2.8cm]{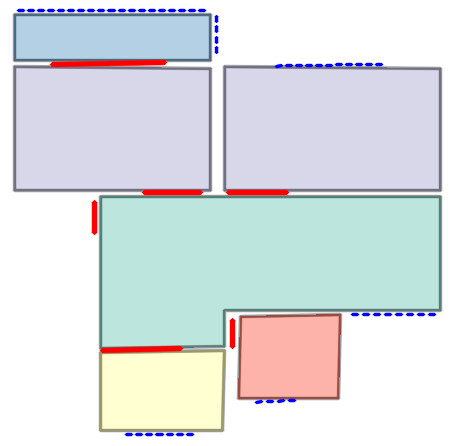}
\includegraphics[height=2.8cm]{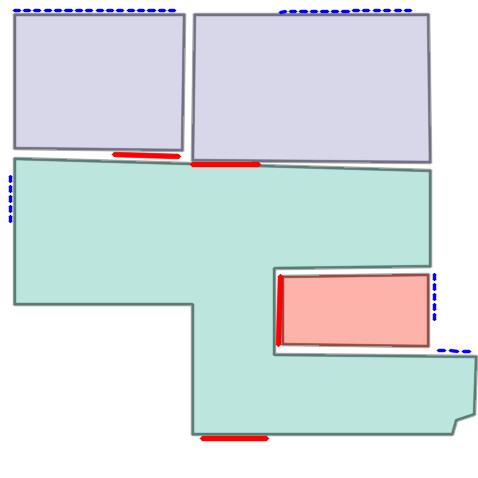}
\includegraphics[height=2.8cm]{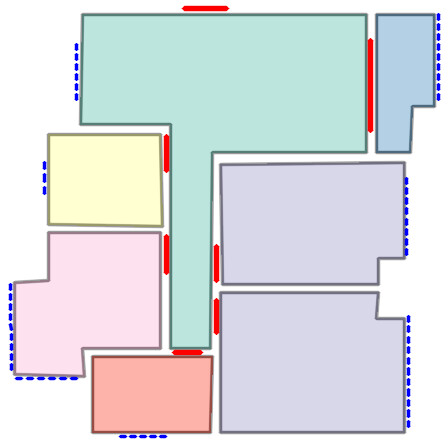}
\includegraphics[height=2.8cm]{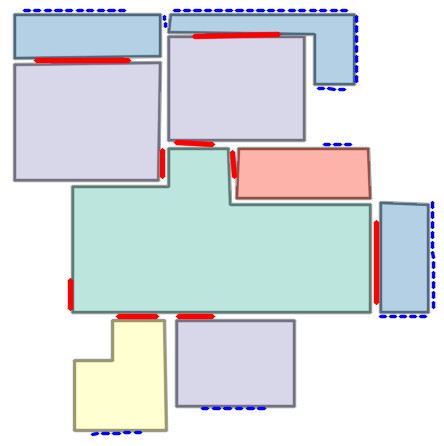}
\includegraphics[height=2.8cm]{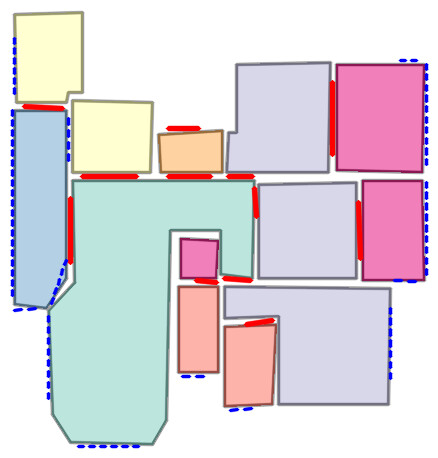}
\includegraphics[height=2.8cm]{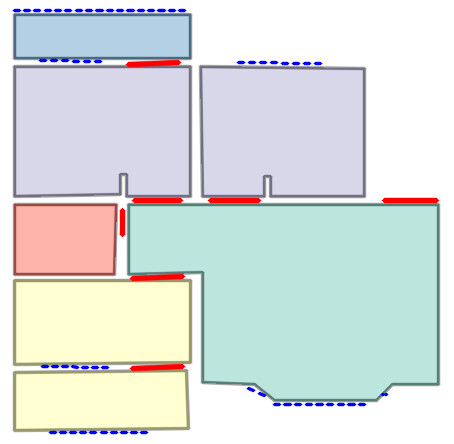}
}{}\\

\rotatebox{90}{\whitetxt{sssssssss}Input}
\jsubfig{
\includegraphics[height=2.8cm]{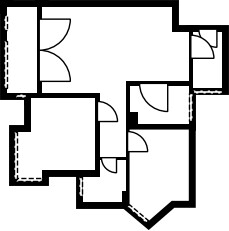}
\includegraphics[height=2.8cm]{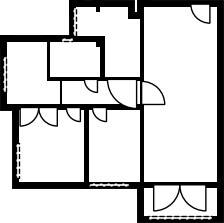}
\includegraphics[height=2.8cm]{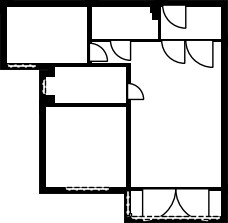}
\includegraphics[height=2.8cm]{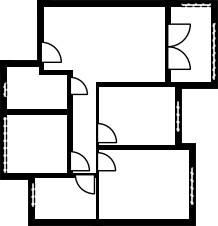}
\includegraphics[height=2.8cm]{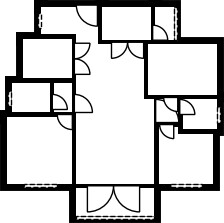}
\includegraphics[height=2.8cm]{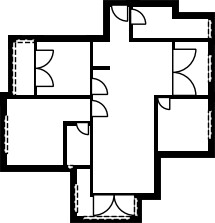}
}{}\\

\rotatebox{90}{\whitetxt{ssssssss}Output}
\jsubfig{
\includegraphics[height=2.8cm]{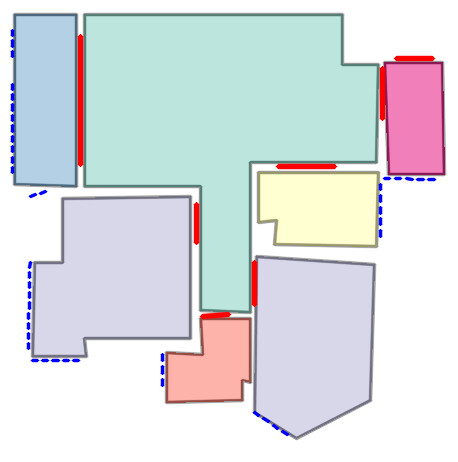}
\includegraphics[height=2.8cm]{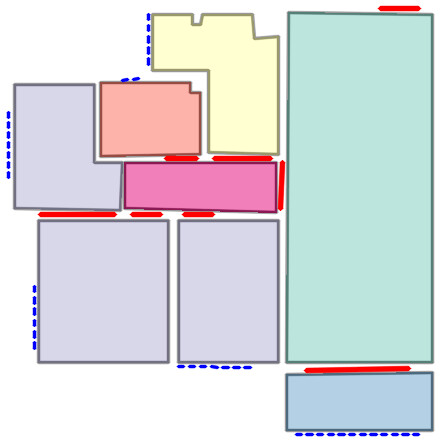}
\includegraphics[height=2.8cm]{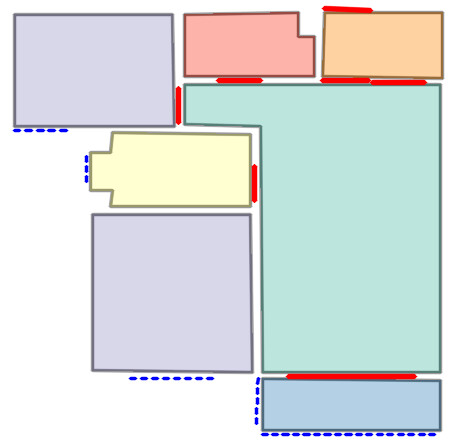}
\includegraphics[height=2.8cm]{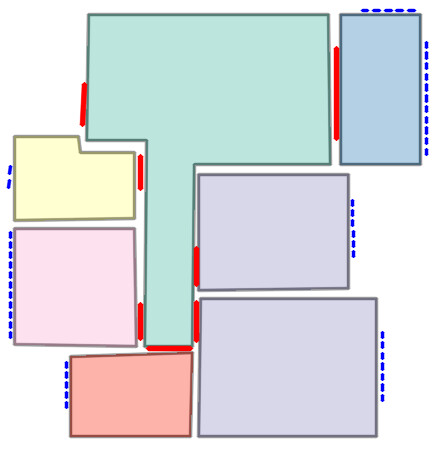}
\includegraphics[height=2.8cm]{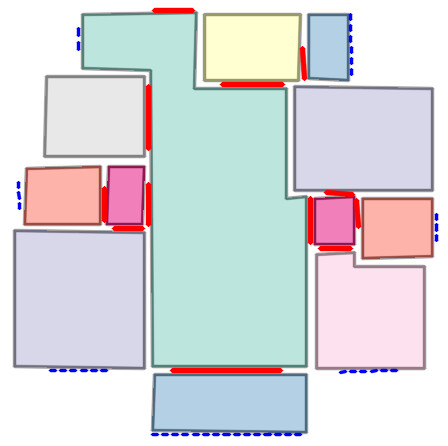}
\includegraphics[height=2.8cm]{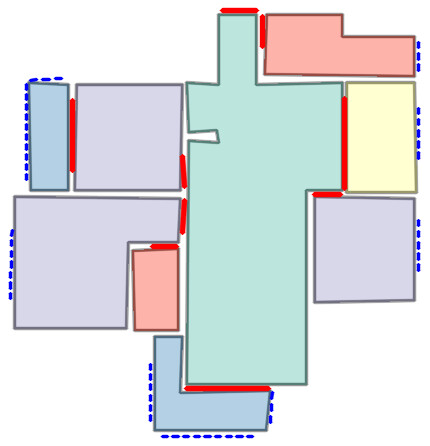}
}{}\\

\rotatebox{90}{\whitetxt{sssssssss}Input}
\jsubfig{
\includegraphics[height=2.8cm]{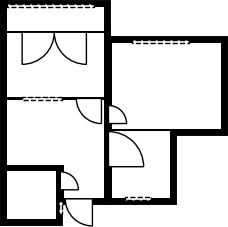}
\includegraphics[height=2.8cm]{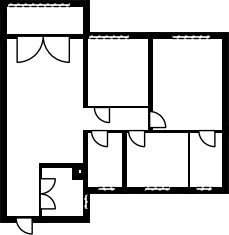}
\includegraphics[height=2.8cm]{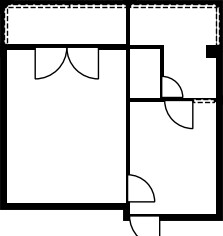}
\includegraphics[height=2.8cm]{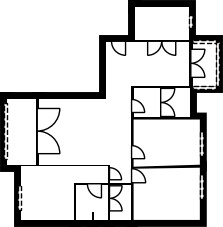}
\includegraphics[height=2.8cm]{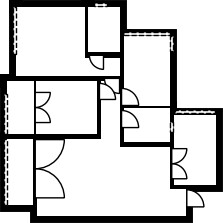}
\includegraphics[height=2.8cm]{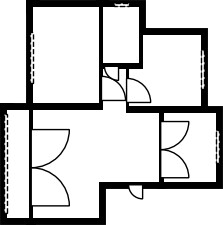}
}{}\\

\rotatebox{90}{\whitetxt{ssssssss}Output}
\jsubfig{
\includegraphics[height=2.8cm]{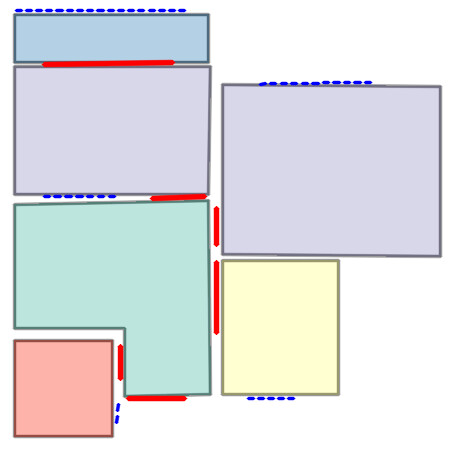}
\includegraphics[height=2.8cm]{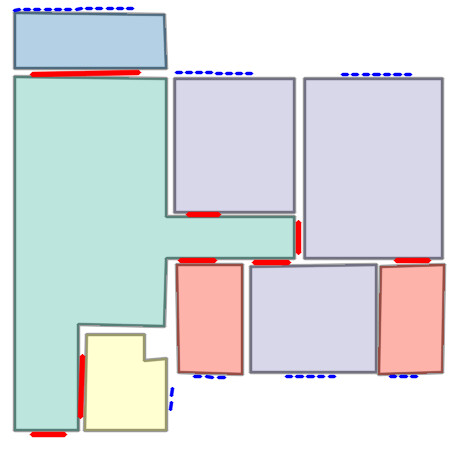}
\includegraphics[height=2.8cm]{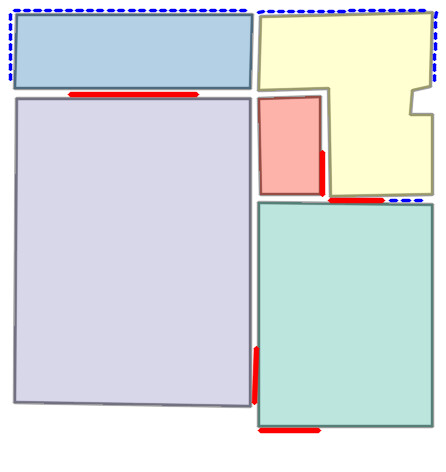}
\includegraphics[height=2.8cm]{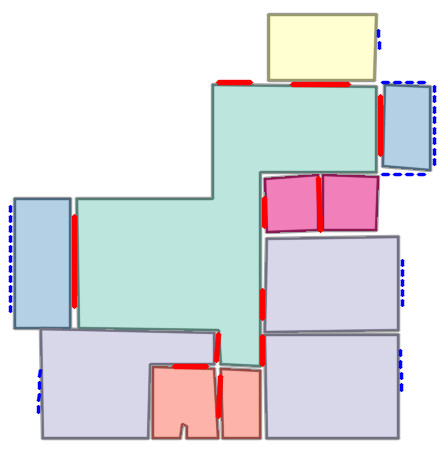}
\includegraphics[height=2.8cm]{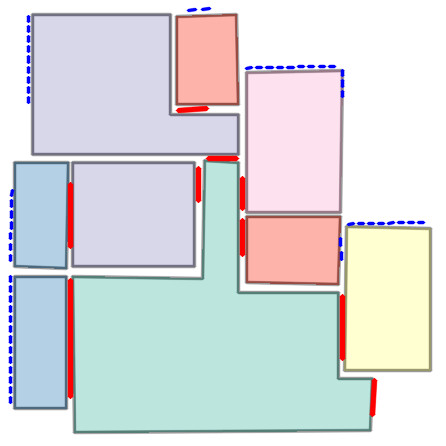}
\includegraphics[height=2.8cm]{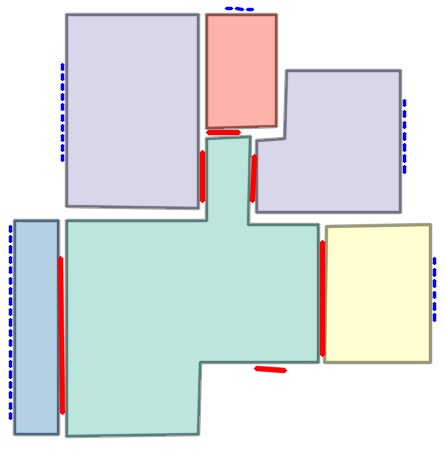}
}{}\\

\vspace{4pt}
\begin{tikzpicture}

\fill[s3d_livingroom, opacity=0.59] (0,1.5) rectangle (1.8,1);
\node[black] at (0.9,1.25) {\small\textbf{Living Room}};

\fill[s3d_kitchen, opacity=0.59] (2.1,1.5) rectangle (3.9,1);
\node[black] at (3,1.25) {\small\textbf{Kitchen}};

\fill[s3d_bedroom, opacity=0.59] (4.2,1.5) rectangle (6.0,1);
\node[black] at (5.1,1.25) {\small\textbf{Bedroom}};

\fill[s3d_bathroom, opacity=0.59] (6.3,1.5) rectangle (8.1,1);
\node[black] at (7.2,1.25) {\small\textbf{Bathroom}};

\fill[s3d_balcony, opacity=0.59] (8.4,1.5) rectangle (10.2,1);
\node[black] at (9.3,1.25) {\small\textbf{Balcony}};

\fill[s3d_corridor, opacity=0.59] (10.5,1.5) rectangle (12.3,1);
\node[black] at (11.4,1.25) {\small\textbf{Corridor}};

\fill[s3d_diningroom, opacity=0.59] (12.6,1.5) rectangle (14.4,1);
\node[black] at (13.5,1.25) {\small\textbf{Dining room}};

\fill[s3d_study, opacity=0.59] (14.7,1.5) rectangle (16.5,1);
\node[black] at (15.6,1.25) {\small\textbf{Study}};

\fill[s3d_studio, opacity=0.59] (0,0.9) rectangle (1.8,0.4);
\node[black] at (0.9,0.65) {\small\textbf{Studio}};

\fill[s3d_storeroom, opacity=0.59] (2.1,0.9) rectangle (3.9,0.4);
\node[black] at (3,0.65) {\small\textbf{Store room}};

\fill[s3d_garden, opacity=0.59] (4.2,0.9) rectangle (6.0,0.4);
\node[black] at (5.1,0.65) {\small\textbf{Garden}};

\fill[s3d_laundryroom, opacity=0.59] (6.3,0.9) rectangle (8.1,0.4);
\node[black] at (7.2,0.65) {\small\textbf{Laundry room}};

\fill[s3d_office, opacity=0.59] (8.4,0.9) rectangle (10.2,0.4);
\node[black] at (9.3,0.65) {\small\textbf{Office}};

\fill[s3d_basement, opacity=0.59] (10.5,0.9) rectangle (12.3,0.4);
\node[black] at (11.4,0.65) {\small\textbf{Basement}};

\fill[s3d_garage, opacity=0.59] (12.6,0.9) rectangle (14.4,0.4);
\node[black] at (13.5,0.65) {\small\textbf{Garage}};

\fill[s3d_misc, opacity=0.59] (14.7,0.9) rectangle (16.5,0.4);
\node[black] at (15.6,0.65) {\small\textbf{Misc}};

\draw[red, very thick, opacity=0.59] (6,0) -- (7,0);
\node[right] at (7.2,0) {\textcolor{red}{Door} };

\draw[blue, dashed, very thick, opacity=0.59] (9,0) -- (10,0);
\node[right] at (10.2,0) {\textcolor{blue}{Window}};

\end{tikzpicture}

\vspace{-4pt}

\caption{Additional qualitative results on Structured3D.}
\label{fig:s3d_more}
\end{figure*}

\begin{figure*}[t]
\centering %
\rotatebox{90}{\whitetxt{sssssssss}Input}
\jsubfig{
\includegraphics[height=2.505cm]{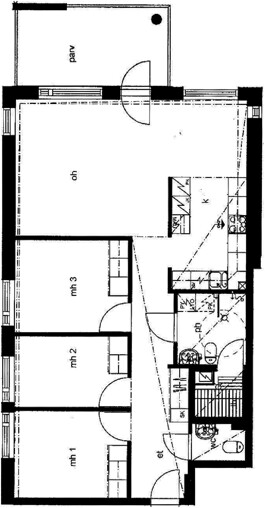}
\includegraphics[height=2.505cm]{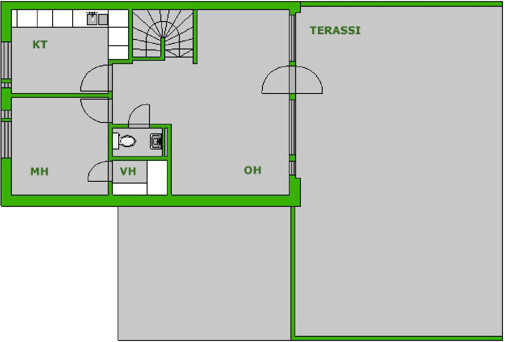}
\includegraphics[height=2.505cm]{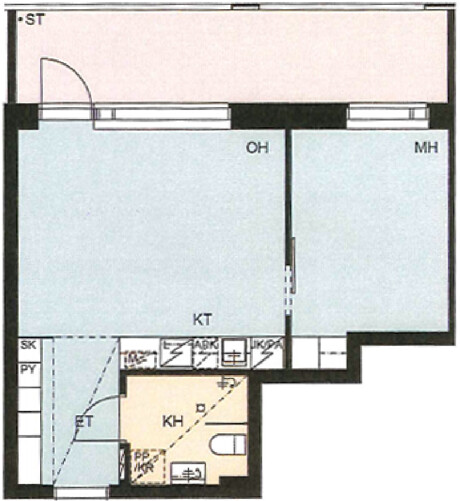}
\includegraphics[height=2.505cm]{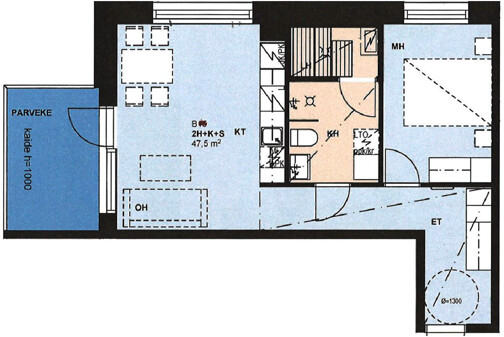}
\includegraphics[height=2.505cm]{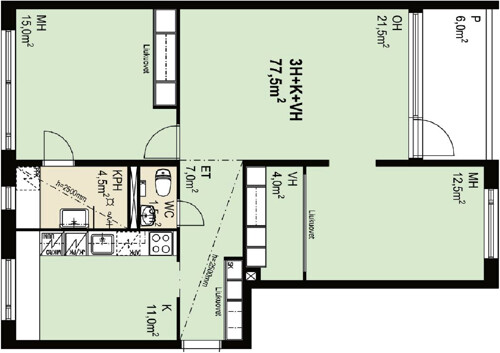}
\includegraphics[height=2.505cm]{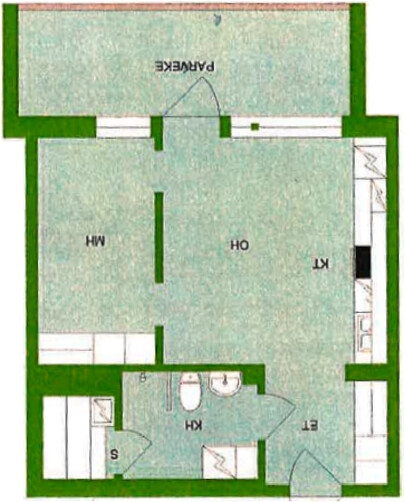}
}{}\\

\rotatebox{90}{\whitetxt{sssssss}Output}
\jsubfig{
\includegraphics[height=2.505cm]{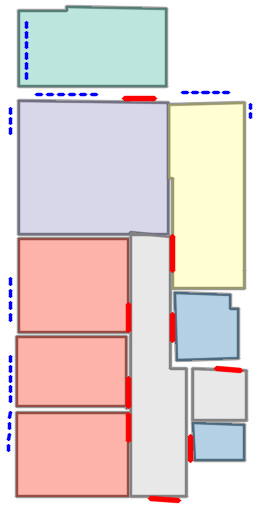}
\includegraphics[height=2.505cm]{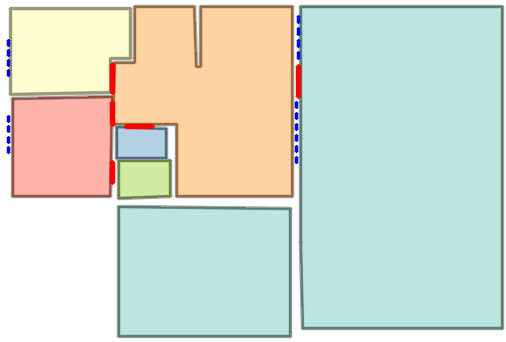}
\includegraphics[height=2.505cm]{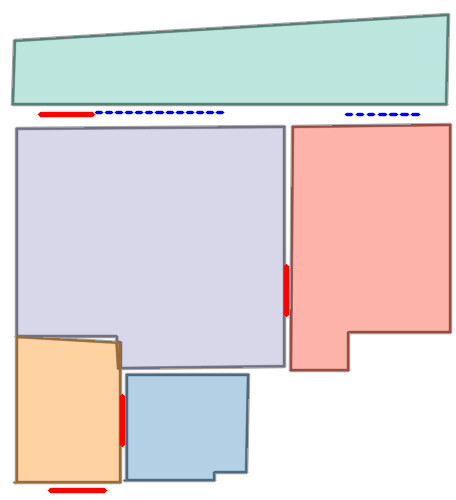}
\includegraphics[height=2.505cm]{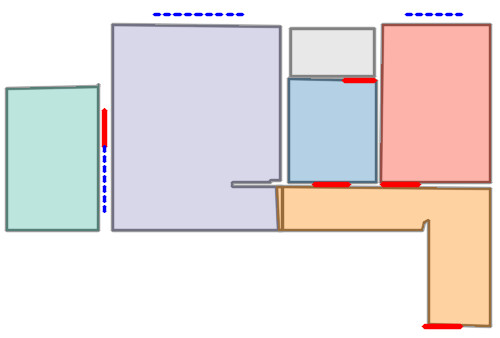}
\includegraphics[height=2.505cm]{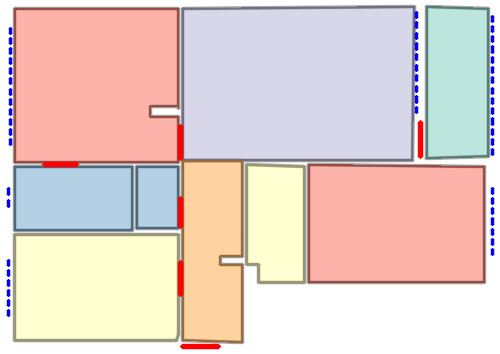}
\includegraphics[height=2.505cm]{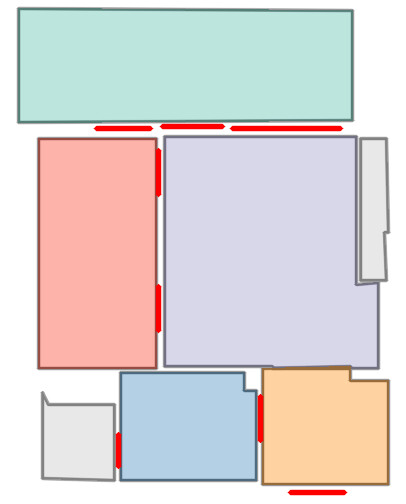}
}{}\\

\rotatebox{90}{\whitetxt{sssssssss}Input}
\jsubfig{
\includegraphics[height=2.505cm]{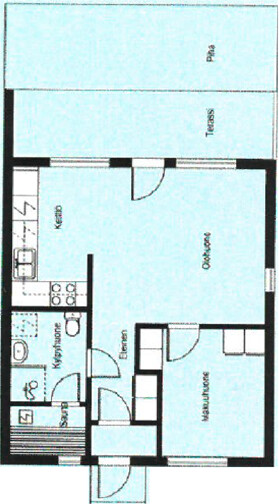}
\includegraphics[height=2.505cm]{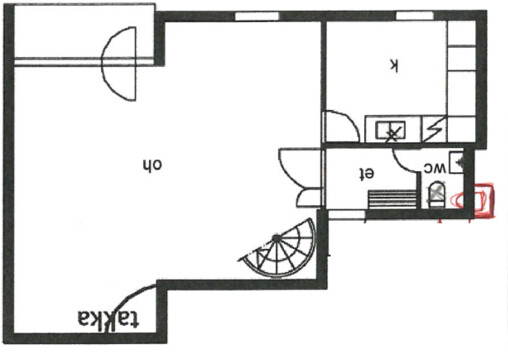}
\includegraphics[height=2.505cm]{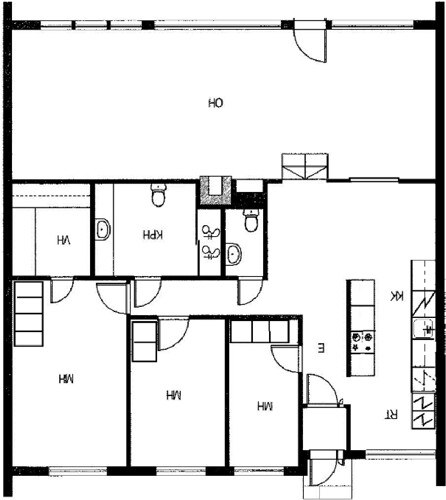}
\includegraphics[height=2.505cm]{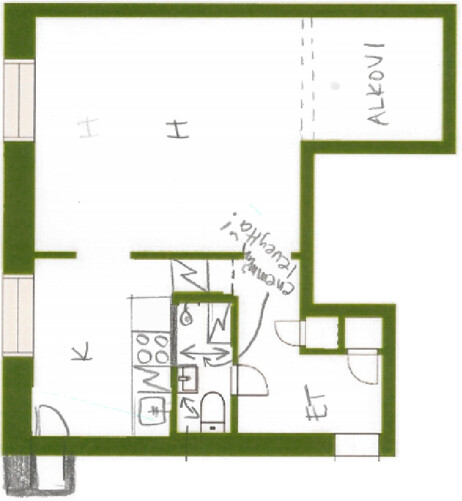}
\includegraphics[height=2.505cm]{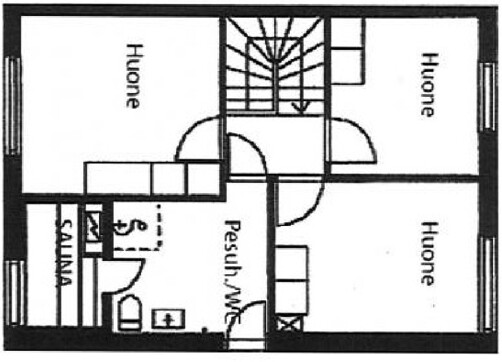}
\includegraphics[height=2.505cm]{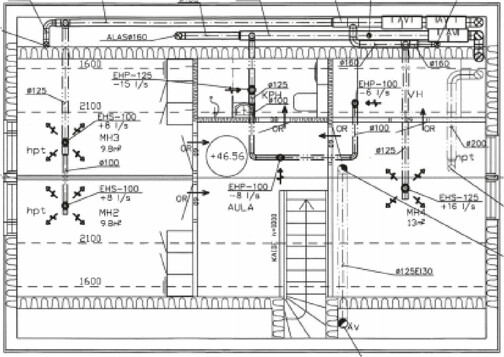}
}{}\\

\rotatebox{90}{\whitetxt{sssssss}Output}
\jsubfig{
\includegraphics[height=2.505cm]{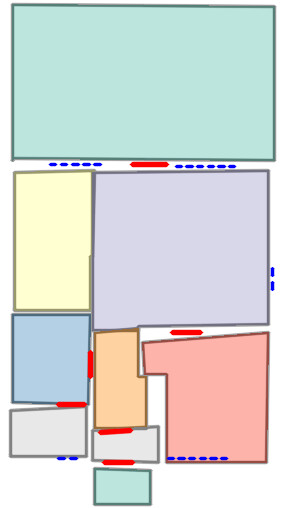}
\includegraphics[height=2.505cm]{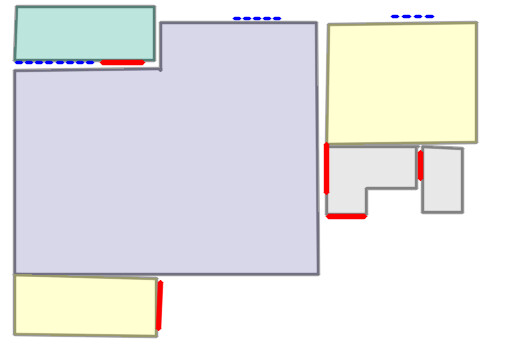}
\includegraphics[height=2.505cm]{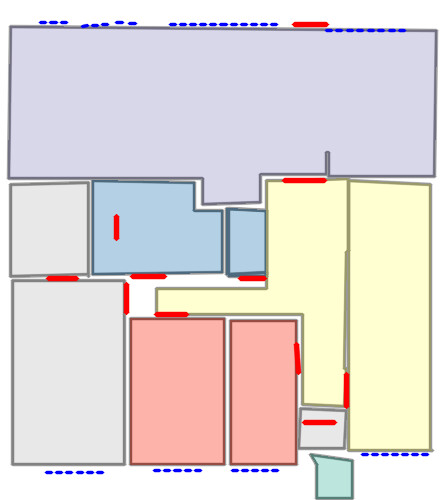}
\includegraphics[height=2.505cm]{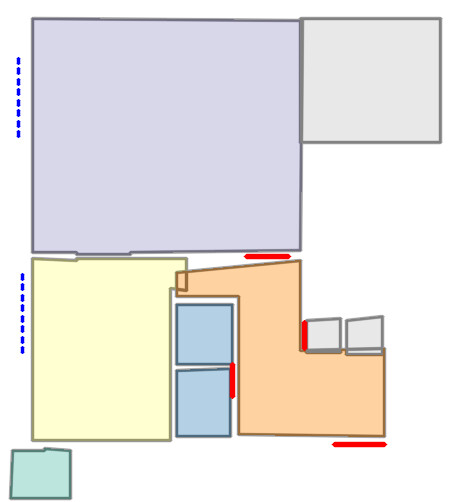}
\includegraphics[height=2.505cm]{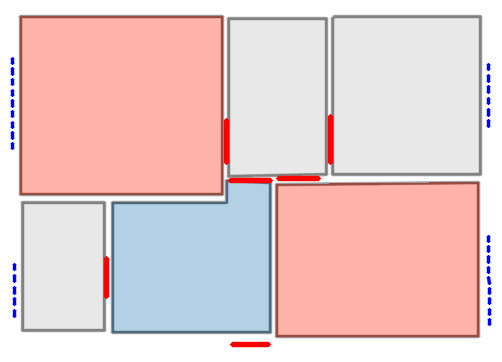}
\includegraphics[height=2.505cm]{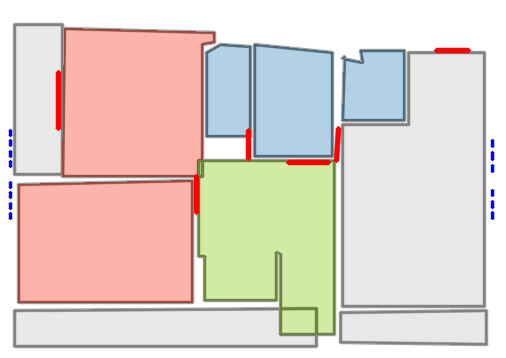}
}{}\\

\rotatebox{90}{\whitetxt{sssssss}Input}
\jsubfig{
\includegraphics[height=2.45cm]{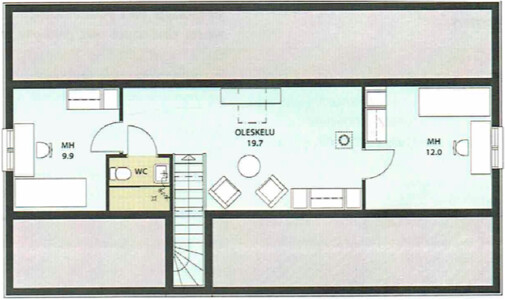}
\includegraphics[height=2.45cm]{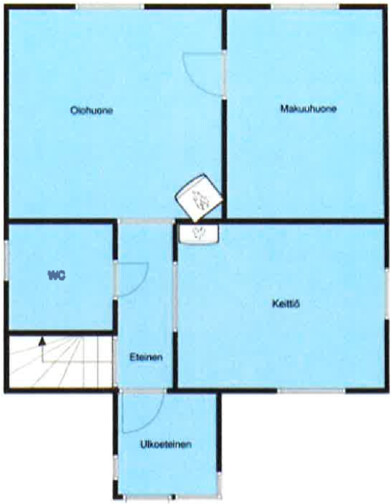}
\includegraphics[height=2.45cm]{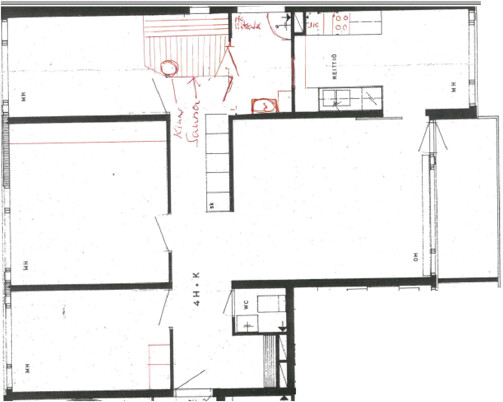}
\includegraphics[height=2.45cm]{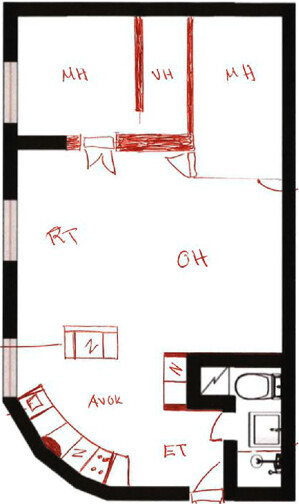}
\includegraphics[height=2.45cm]{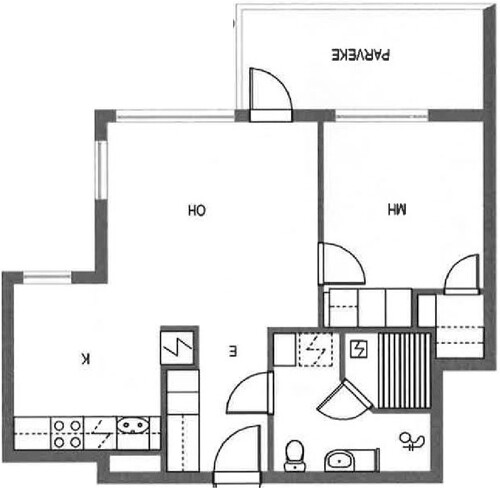}
\includegraphics[height=2.45cm]{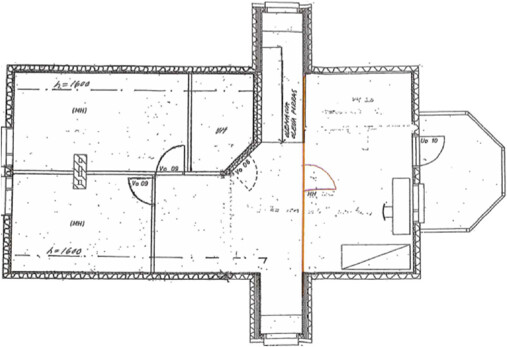}

}{}\\

\rotatebox{90}{\whitetxt{sssssss}Output}
\jsubfig{
\includegraphics[height=2.45cm]{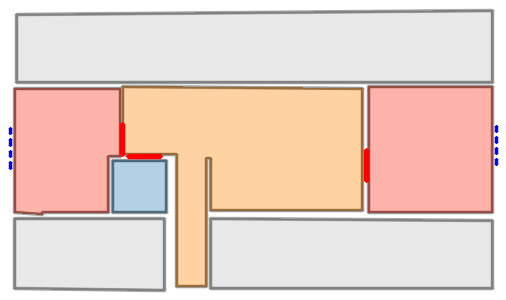}
\includegraphics[height=2.45cm]{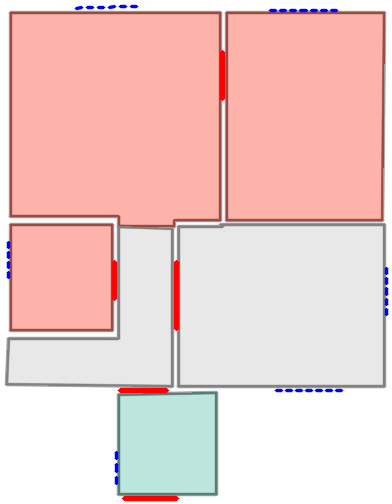}
\includegraphics[height=2.45cm]{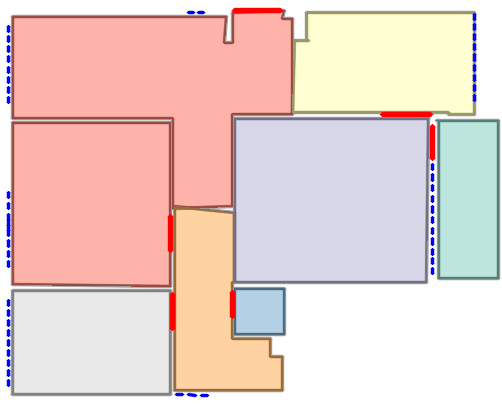}
\includegraphics[height=2.45cm]{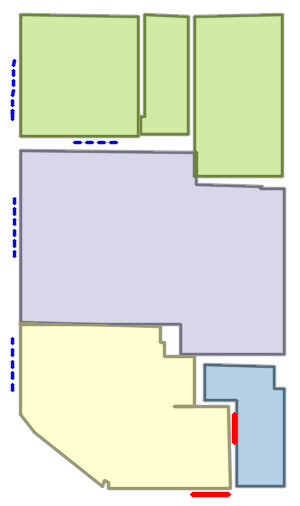}
\includegraphics[height=2.45cm]{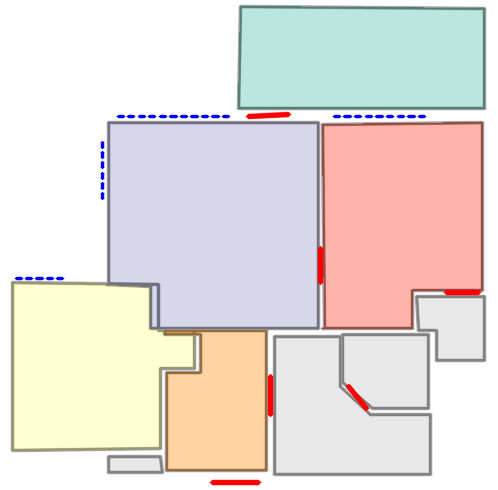}
\includegraphics[height=2.45cm]{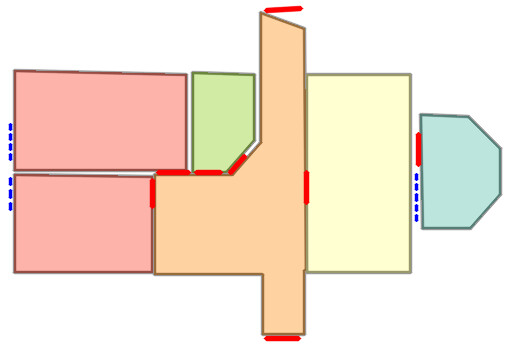}

}{}\\

\rotatebox{90}{\whitetxt{sssssssss}Input}
\jsubfig{
\includegraphics[height=2.45cm]{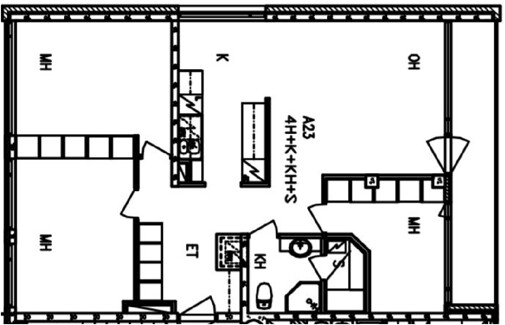}
\includegraphics[height=2.45cm]{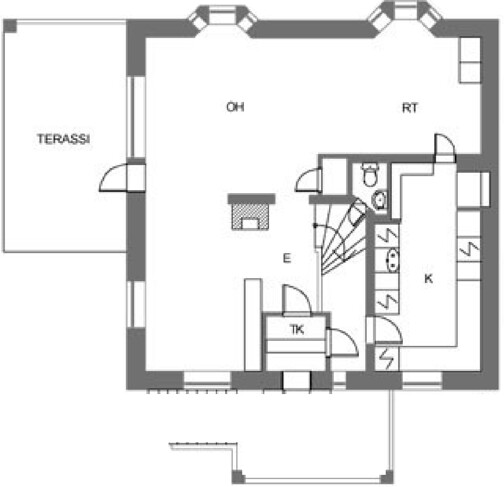}
\includegraphics[height=2.45cm]{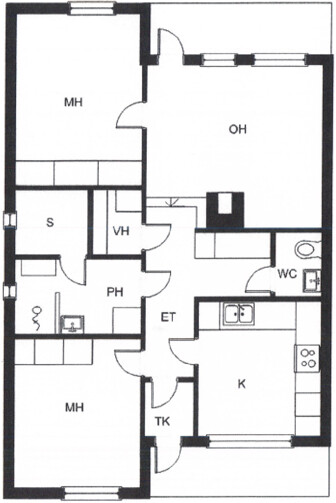}
\includegraphics[height=2.45cm]{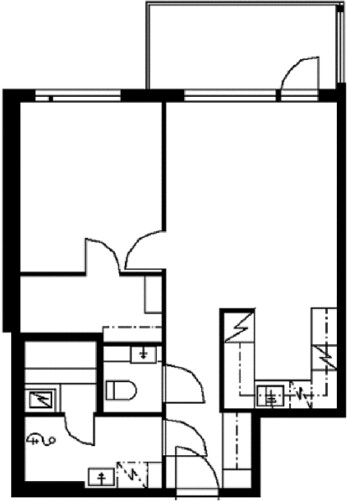}
\includegraphics[height=2.45cm]{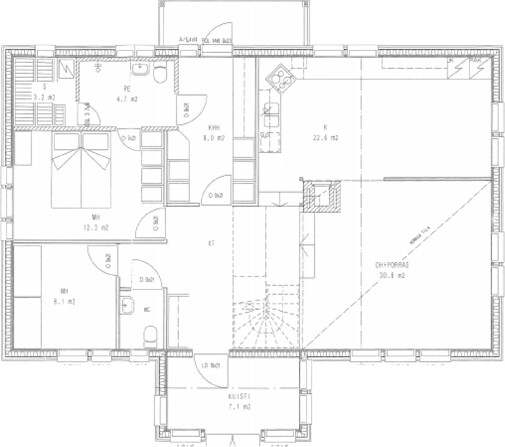}
\includegraphics[height=2.45cm]{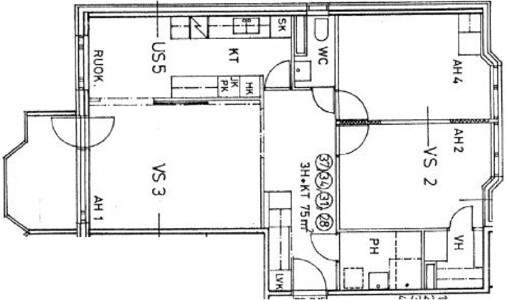}
}{}\\
\rotatebox{90}{\whitetxt{sssssss}Output}
\jsubfig{
\includegraphics[height=2.45cm]{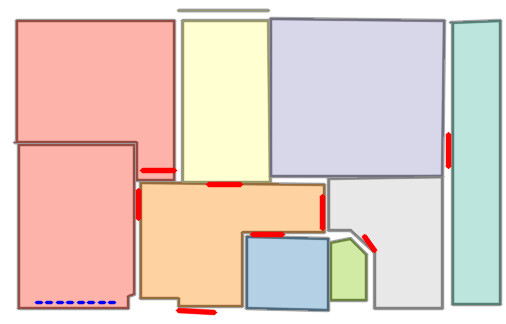}
\includegraphics[height=2.45cm]{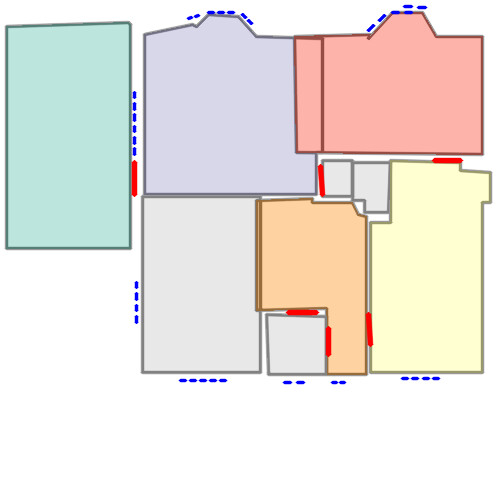}
\includegraphics[height=2.45cm]{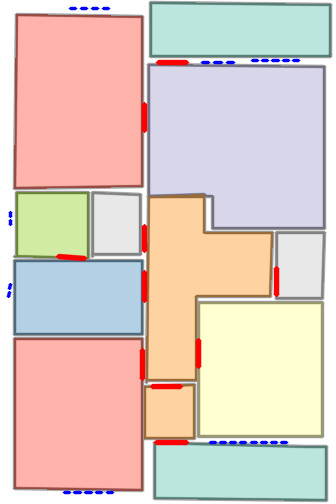}
\includegraphics[height=2.45cm]{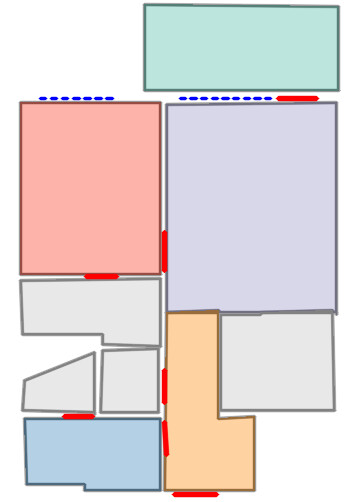}
\includegraphics[height=2.45cm]{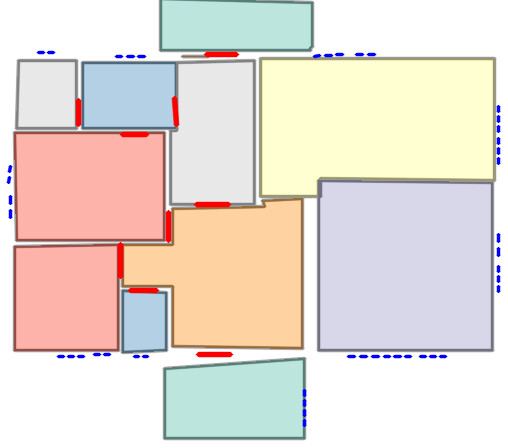}
\includegraphics[height=2.45cm]{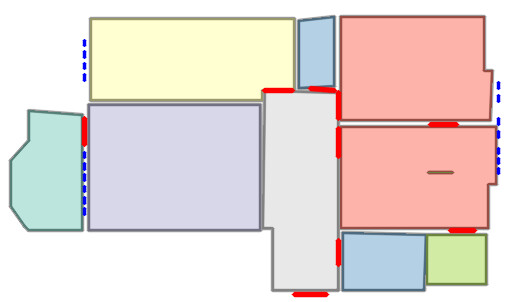}
}{}\\

\vspace{4pt}
\begin{tikzpicture}

\fill[outdoor, opacity=0.59] (0,1) rectangle (1.4,0.5);
\node[black] at (0.7,0.75) {\small\textbf{Outdoor}};

\fill[kitchen, opacity=0.59] (1.8,1) rectangle (3.2,0.5);
\node[black] at (2.5,0.75) {\small\textbf{Kitchen}};

\fill[livingroom, opacity=0.59] (3.6,1) rectangle (5.0,0.5);
\node[black] at (4.3,0.75) {\small\textbf{Living room}};

\fill[bedroom, opacity=0.59] (5.4,1) rectangle (6.8,0.5);
\node[black] at (6.1,0.75) {\small\textbf{Bed room}};

\fill[bath, opacity=0.59] (7.2,1) rectangle (8.6,0.5);
\node[black] at (7.9,0.75) {\small\textbf{Bath}};

\fill[entry, opacity=0.59] (9.0,1) rectangle (10.4,0.5);
\node[black] at (9.7,0.75) {\small\textbf{Entry}};

\fill[storage, opacity=0.59] (10.8,1) rectangle (12.2,0.5);
\node[black] at (11.5,0.75) {\small\textbf{Storage}};

\fill[garage, opacity=0.59] (12.6,1) rectangle (14.0,0.5);
\node[black] at (13.3,0.75) {\small\textbf{Garage}};

\fill[undefined, opacity=0.59] (14.4,1) rectangle (15.8,0.5);
\node[black] at (15.1,0.75) {\small\textbf{Undefined}};

\draw[red, very thick, opacity=0.59] (6,0) -- (7,0);
\node[right] at (7.2,0) {\textcolor{red}{Door} };

\draw[blue, dashed, very thick, opacity=0.59] (9,0) -- (10,0);
\node[right] at (10.2,0) {\textcolor{blue}{Window}};

\end{tikzpicture}

\vspace{-4pt}

\caption{Additional qualitative results on CubiCasa5K.}
\label{fig:cubi_more}
\end{figure*}

\begin{figure*}[!ht]
\centering %
\rotatebox{90}{\whitetxt{sssssss}Input}
\jsubfig{
\includegraphics[height=2.35cm]{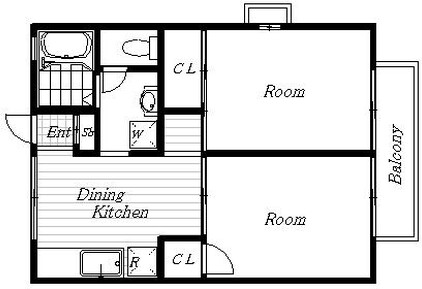}
\includegraphics[width=2.35cm,angle=90]{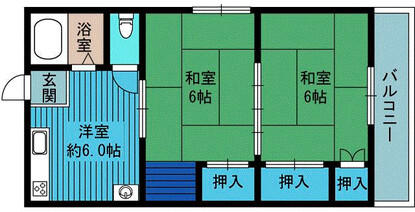}

\includegraphics[width=2.35cm,angle=90]{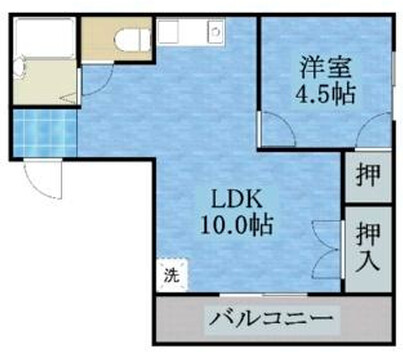}
\includegraphics[height=2.35cm]{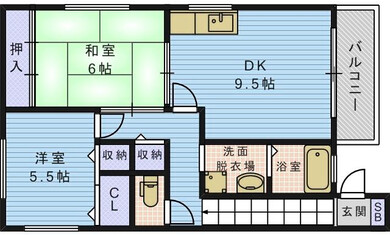}
\includegraphics[height=2.35cm]{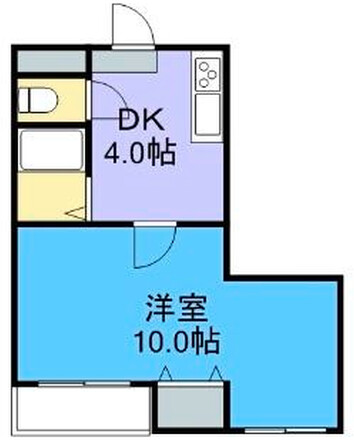}
\includegraphics[height=2.35cm]{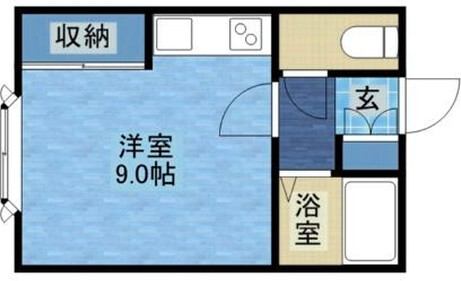}
}{}\\

\rotatebox{90}{\whitetxt{sssssss}Output}
\jsubfig{
\includegraphics[height=2.35cm]{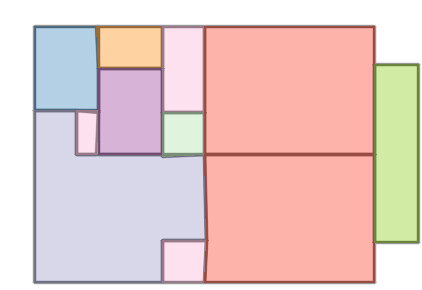}
\includegraphics[width=2.35cm,angle=90]{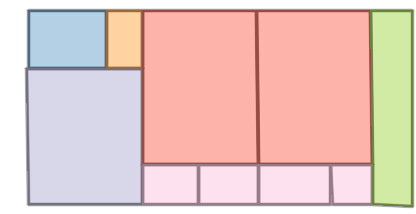}
\includegraphics[width=2.35cm,angle=90]{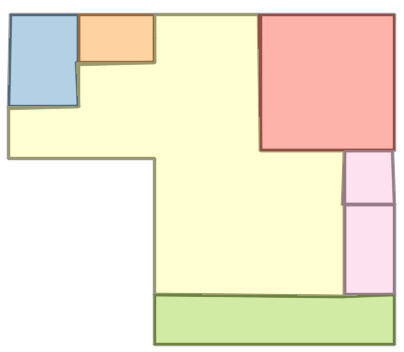}
\includegraphics[height=2.35cm]{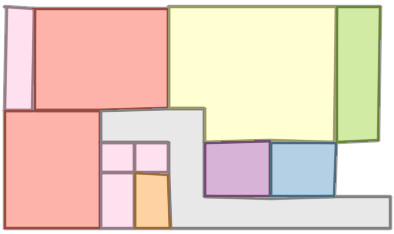}
\includegraphics[height=2.35cm]{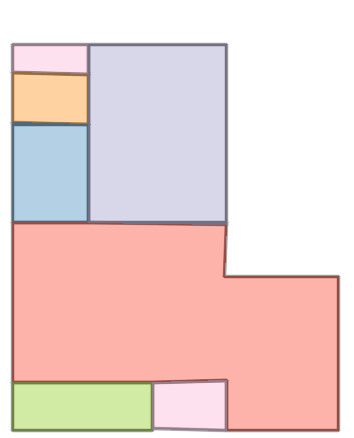}
\includegraphics[height=2.35cm]{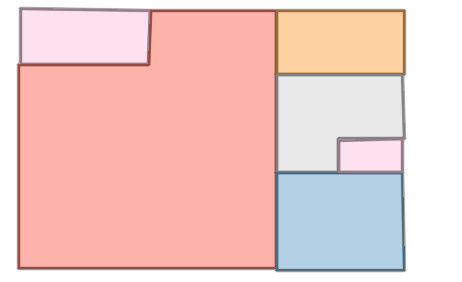}

}{}\\

\rotatebox{90}{\whitetxt{sssssss}Input}
\jsubfig{
\includegraphics[height=2.35cm]{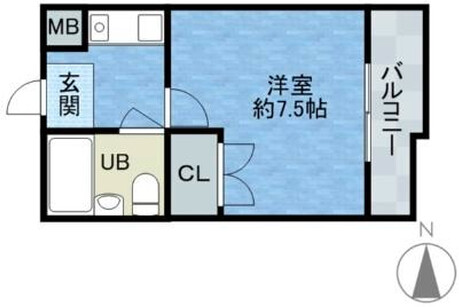}
\includegraphics[width=2.35cm,angle=90]{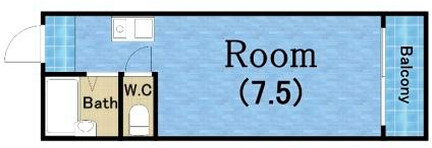}
\includegraphics[height=2.35cm]{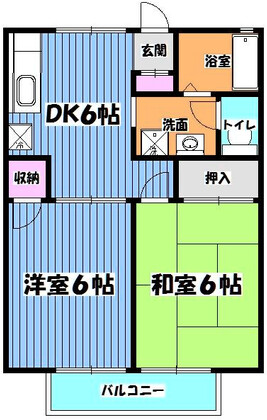}
\includegraphics[height=2.35cm]{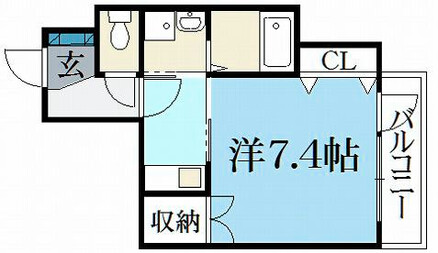}
\includegraphics[height=2.35cm]{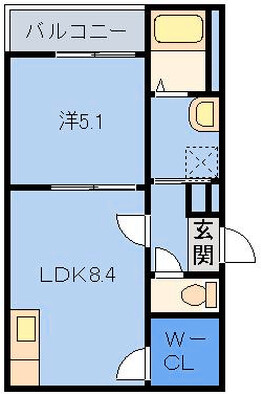}
\includegraphics[height=2.35cm]{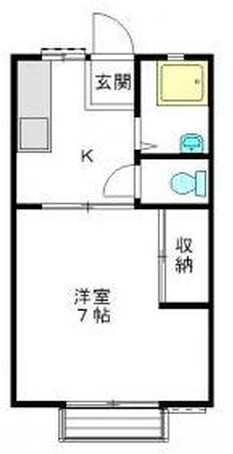}
\includegraphics[width=2.35cm,angle=90]{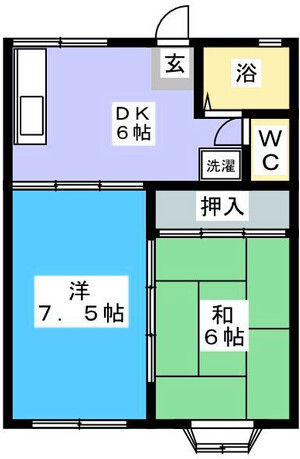}
}{}\\

\rotatebox{90}{\whitetxt{sssssss}Output}
\jsubfig{
\includegraphics[height=2.35cm]{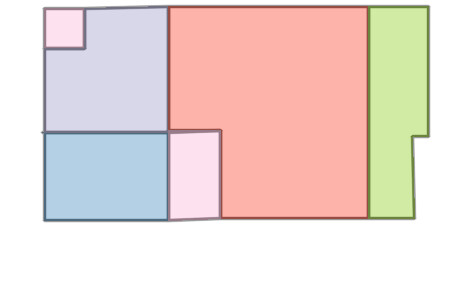}
\includegraphics[width=2.35cm,angle=90]{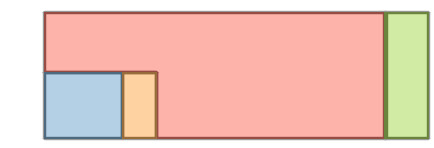}
\includegraphics[height=2.35cm]{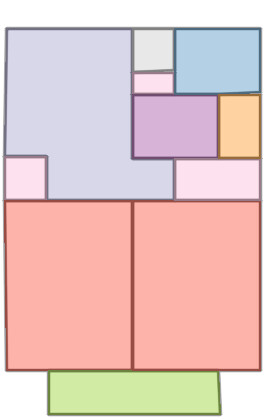}
\includegraphics[height=2.35cm]{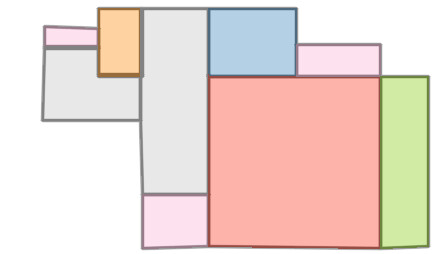}
\includegraphics[height=2.35cm]{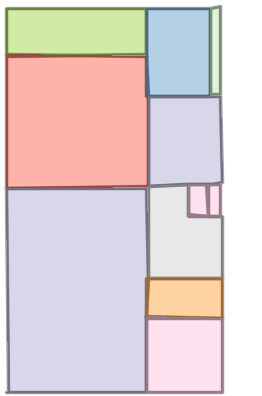}
\includegraphics[height=2.35cm]{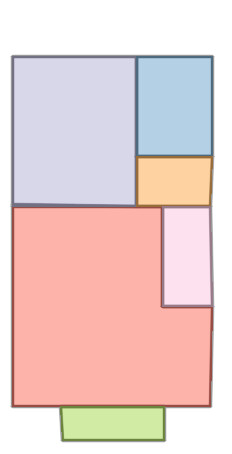}
\includegraphics[width=2.35cm,angle=90]{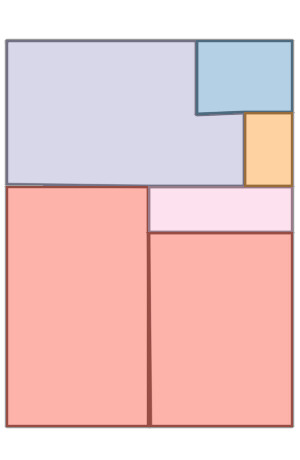}
}{}\\

\rotatebox{90}{\whitetxt{sssssss}Input}
\jsubfig{
\includegraphics[height=2.35cm]{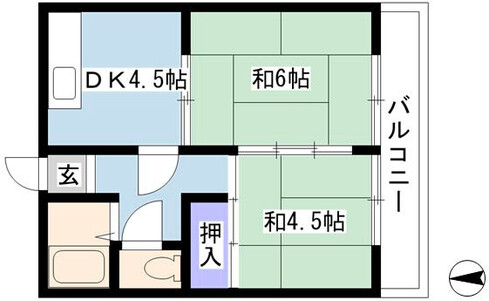}
\includegraphics[height=2.35cm]{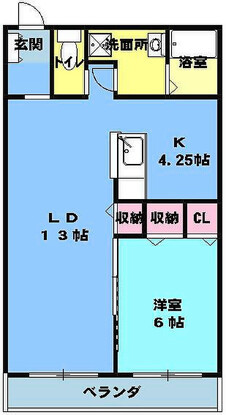}
\includegraphics[height=2.35cm]{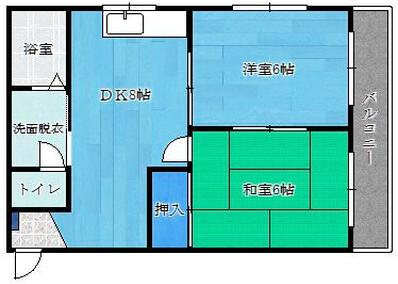}
\includegraphics[height=2.35cm]{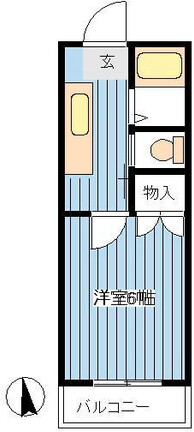}
\includegraphics[width=2.35cm,angle=90]{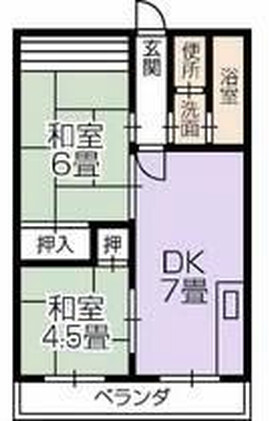}
\includegraphics[height=2.35cm]{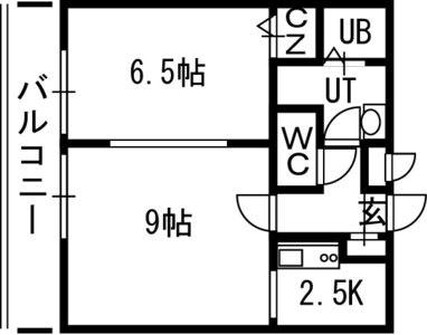}
}{}\\

\rotatebox{90}{\whitetxt{sssssss}Output}
\jsubfig{
\includegraphics[height=2.35cm]{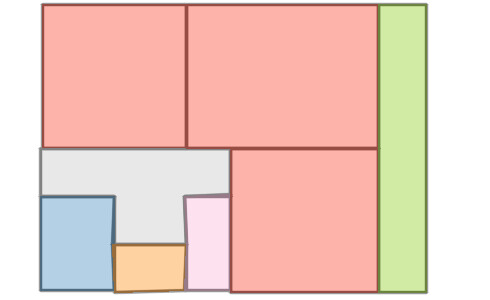}
\includegraphics[height=2.35cm]{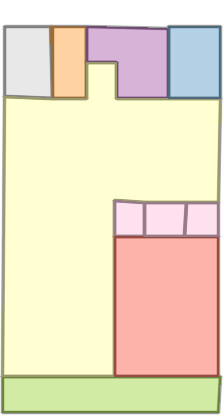}
\includegraphics[height=2.35cm]{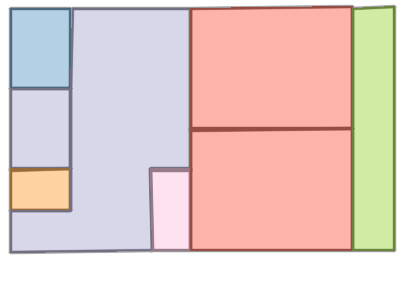}
\includegraphics[height=2.35cm]{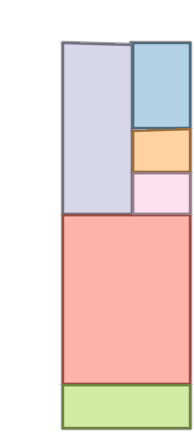}
\includegraphics[width=2.35cm,angle=90]{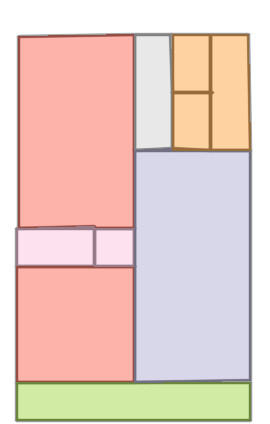}
\includegraphics[height=2.35cm]{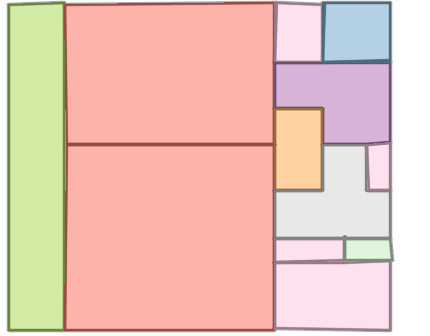}
}{}\\

\rotatebox{90}{\whitetxt{sssssss}Input}
\jsubfig{
\includegraphics[height=2.35cm]{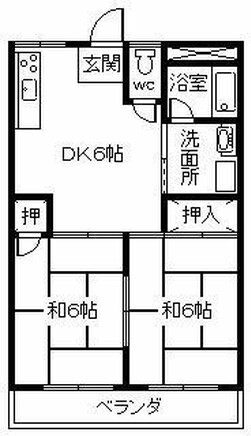}
\includegraphics[height=2.35cm]{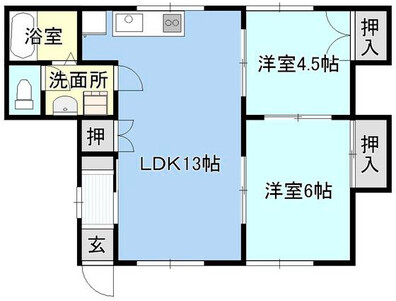}
\includegraphics[height=2.35cm]{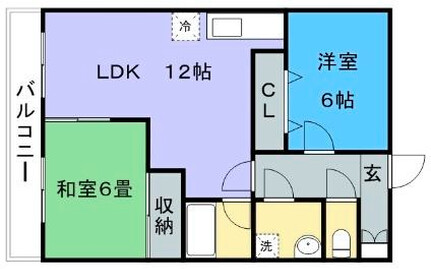}
\includegraphics[width=2.35cm,angle=90]{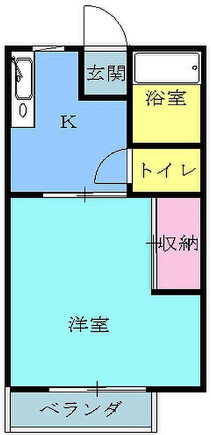}
\includegraphics[height=2.35cm]{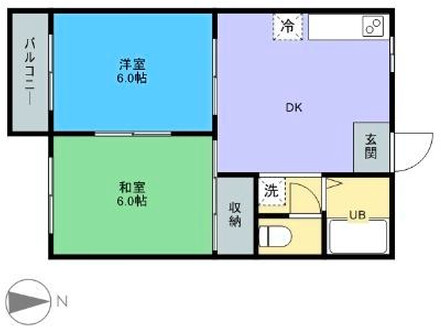}
}{}\\

\rotatebox{90}{\whitetxt{sssssss}Output}
\jsubfig{
\includegraphics[height=2.35cm]{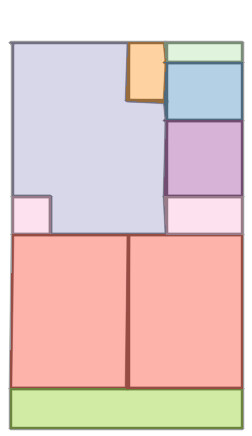}
\includegraphics[height=2.35cm]{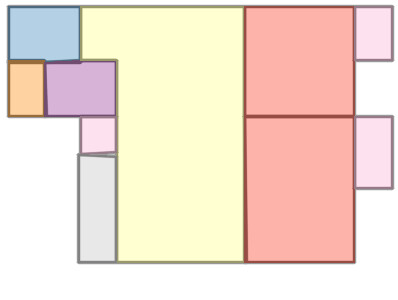}
\includegraphics[height=2.35cm]{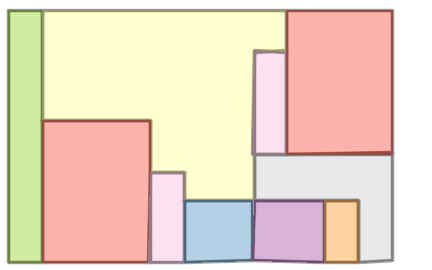}
\includegraphics[width=2.35cm,angle=90]{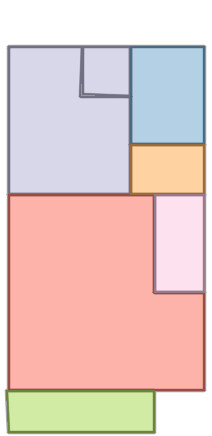}
\includegraphics[height=2.35cm]{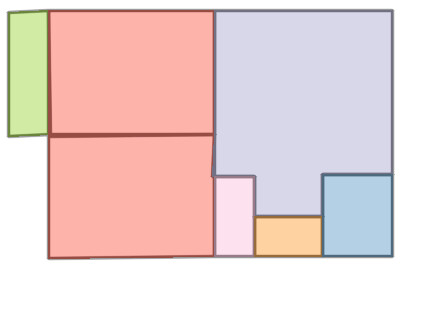}
}{}\\

\vspace{4pt}
\begin{tikzpicture}
\fill[r2g_unknown, opacity=0.59] (0,1.5) rectangle (1.8,1);
\node[black] at (0.9,1.25) {\small\textbf{Unknown}};

\fill[r2g_living_room, opacity=0.59] (2.1,1.5) rectangle (3.9,1);
\node[black] at (3,1.25) {\small\textbf{Living room}};

\fill[r2g_kitchen, opacity=0.59] (4.2,1.5) rectangle (6.0,1);
\node[black] at (5.1,1.25) {\small\textbf{Kitchen}};

\fill[r2g_bedroom, opacity=0.59] (6.3,1.5) rectangle (8.1,1);
\node[black] at (7.2,1.25) {\small\textbf{Bedroom}};

\fill[r2g_bathroom, opacity=0.59] (8.4,1.5) rectangle (10.2,1);
\node[black] at (9.3,1.25) {\small\textbf{Bathroom}};

\fill[r2g_restroom, opacity=0.59] (10.5,1.5) rectangle (12.3,1);
\node[black] at (11.4,1.25) {\small\textbf{Restroom}};

\fill[r2g_balcony, opacity=0.59] (12.6,1.5) rectangle (14.4,1);
\node[black] at (13.5,1.25) {\small\textbf{Balcony}};

\fill[r2g_closet, opacity=0.59] (14.7,1.5) rectangle (16.5,1);
\node[black] at (15.6,1.25) {\small\textbf{Closet}};

\fill[r2g_corridor, opacity=0.59] (4.2,0.75) rectangle (6.0,0.25);
\node[black] at (5.1,0.5) {\small\textbf{Corridor}};

\fill[r2g_washing_room, opacity=0.59] (6.3,0.75) rectangle (8.1,0.25);
\node[black] at (7.2,0.5) {\small\textbf{Washing room}};

\fill[r2g_PS, opacity=0.59] (8.4,0.75) rectangle (10.2,0.25);
\node[black] at (9.3,0.5) {\small\textbf{PS}};

\fill[r2g_outside, opacity=0.59] (10.5,0.75) rectangle (12.3,0.25);
\node[black] at (11.4,0.5) {\small\textbf{Outside}};

\end{tikzpicture}

\vspace{2pt}

\caption{\change{Additional qualitative results on Raster2Graph.}}

\label{fig:raster2graph_more}
\end{figure*}

\begin{figure*}[t]
\centering %

\rotatebox{90}{\whitetxt{sssssssss}Input}
\jsubfig{
\includegraphics[height=2.4cm]{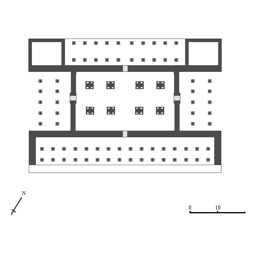}
\includegraphics[height=2.4cm]{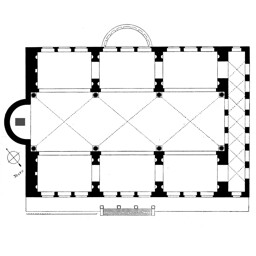}
\includegraphics[height=2.4cm]{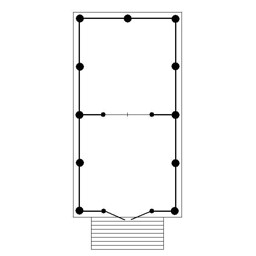}
\includegraphics[height=2.4cm]{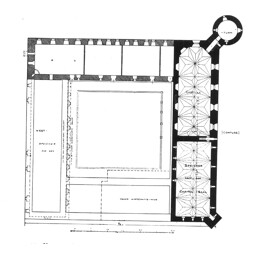}
\includegraphics[height=2.4cm]{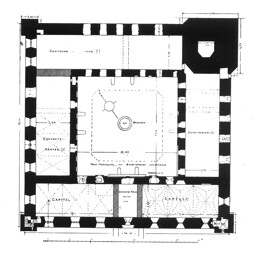}
\includegraphics[height=2.4cm]{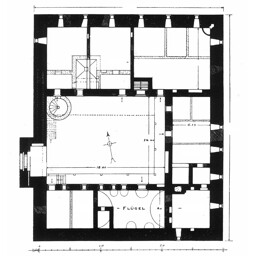}
}{}\\

\rotatebox{90}{\whitetxt{ssssss}Ours}
\jsubfig{
\includegraphics[height=2.4cm]{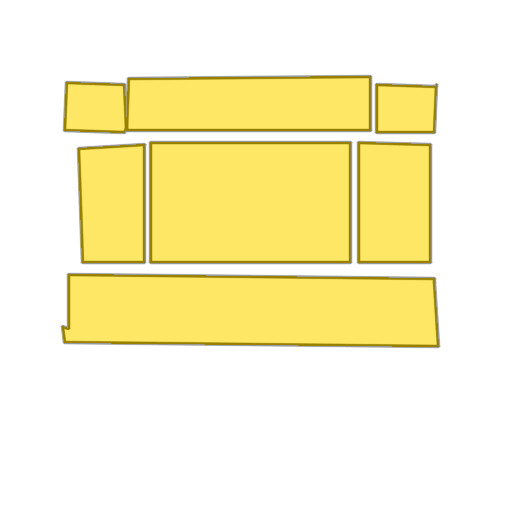}
\includegraphics[height=2.4cm]{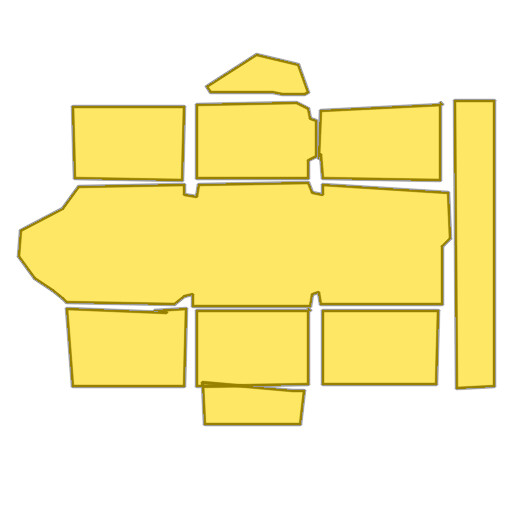}
\includegraphics[height=2.4cm]{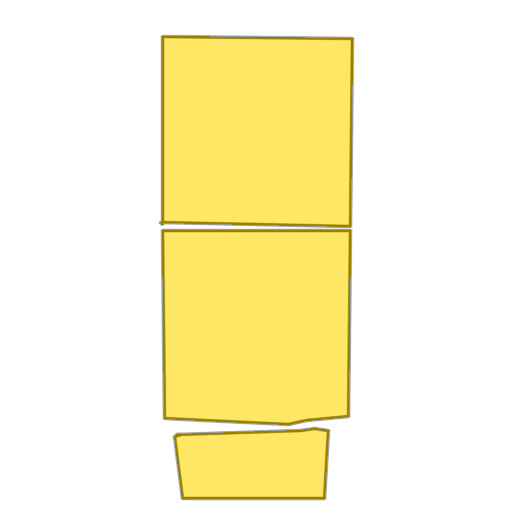}
\includegraphics[height=2.4cm]{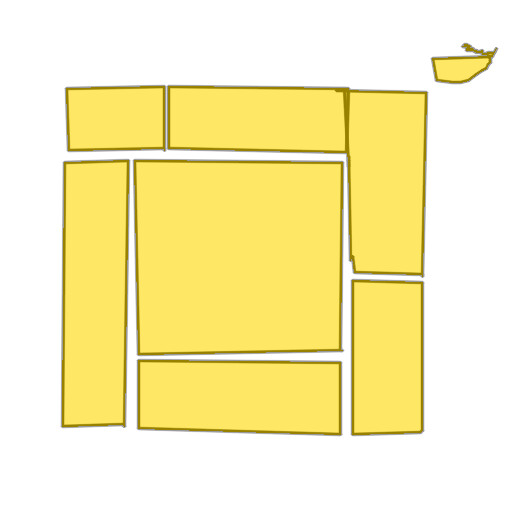}
\includegraphics[height=2.4cm]{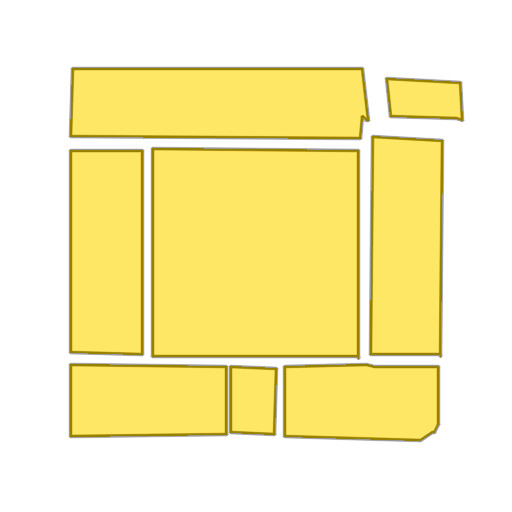}
\includegraphics[height=2.4cm]{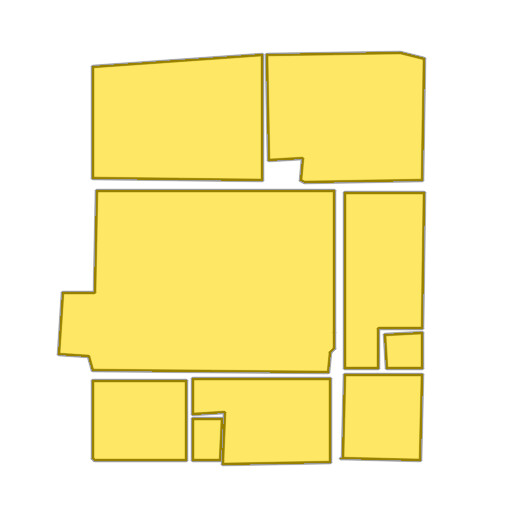}
}{}\\

\rotatebox{90}{\whitetxt{sssssssss}Input}
\jsubfig{
\includegraphics[height=2.4cm]{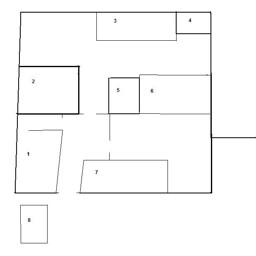}
\includegraphics[height=2.4cm]{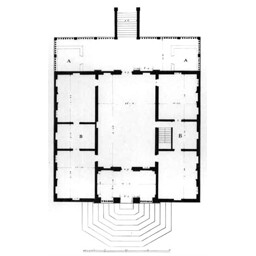}
\includegraphics[height=2.4cm]{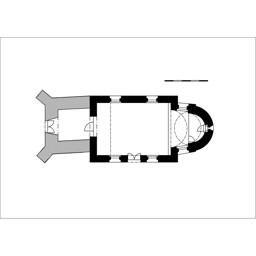}
\includegraphics[height=2.4cm]{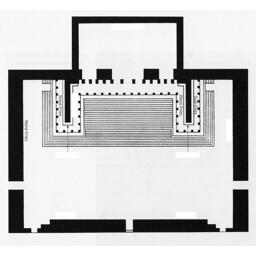}
\includegraphics[height=2.4cm]{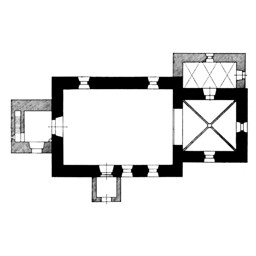}
\includegraphics[height=2.4cm]{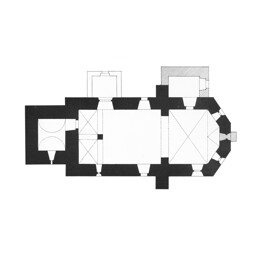}
}{}\\
\rotatebox{90}{\whitetxt{ssssss}Ours}
\jsubfig{
\includegraphics[height=2.4cm]{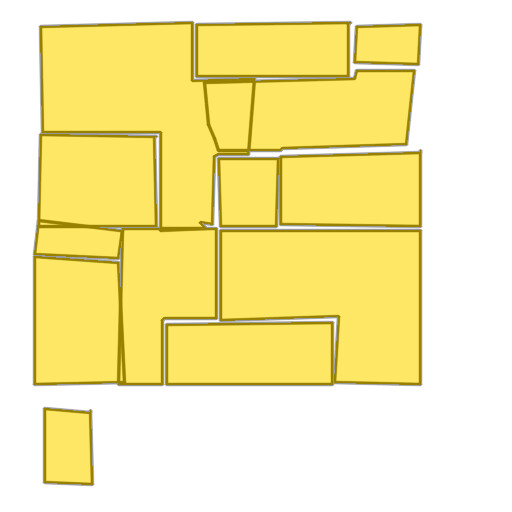}
\includegraphics[height=2.4cm]{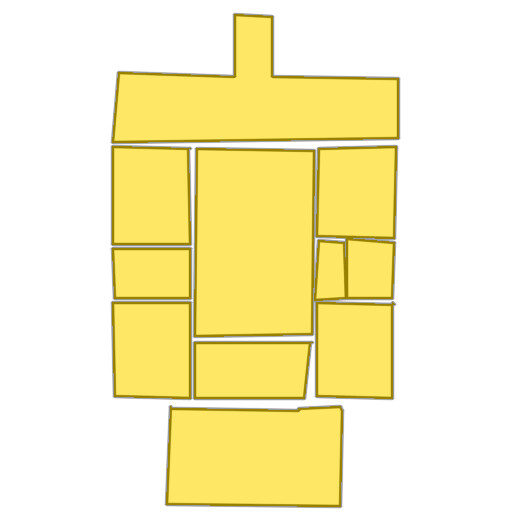}
\includegraphics[height=2.4cm]{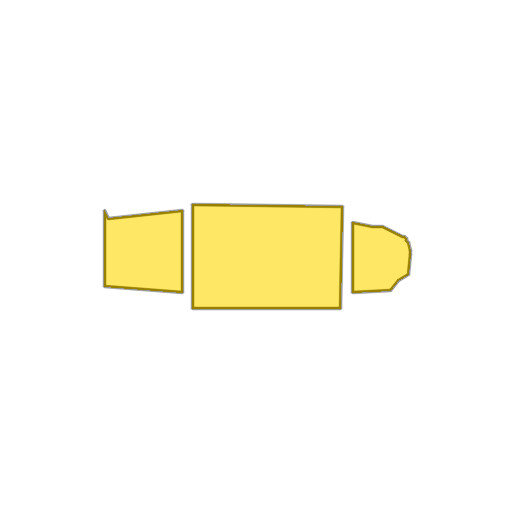}
\includegraphics[height=2.4cm]{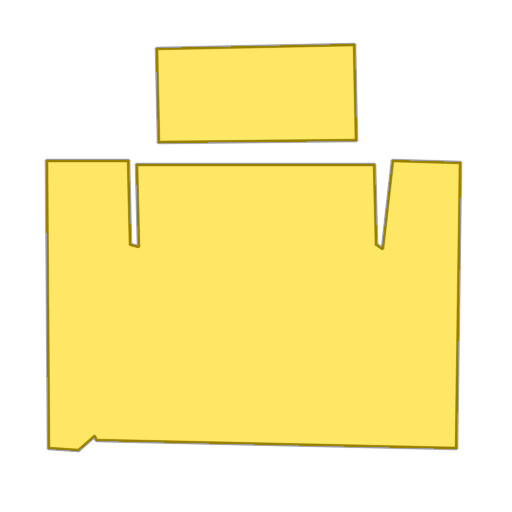}
\includegraphics[height=2.4cm]{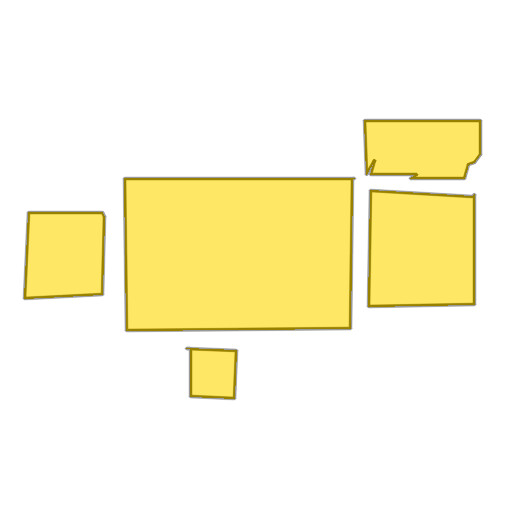}
\includegraphics[height=2.4cm]{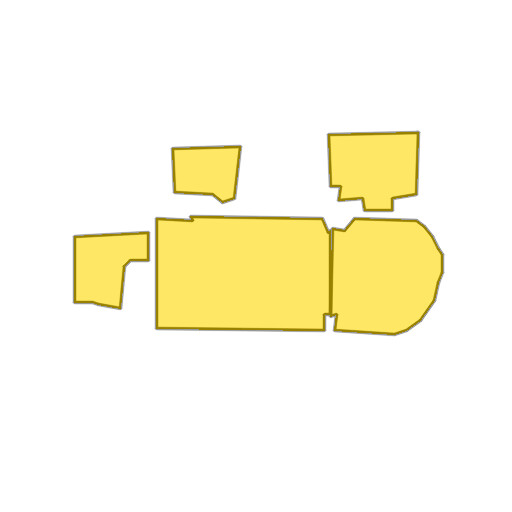}

}{}\\

\vspace{-4pt}

\caption{\change{Additional qualitative results on WAFFLE floorplan images.} Note that these predictions are obtained from our model which is trained on the CubiCasa5K dataset.}
\label{fig:waffle_more}
\end{figure*}

\change{
In this section, we provide comprehensive comparisons on the Structured3D-B, CubiCasa5K, and Raster2Graph datasets, as well as zero-shot evaluation on the WAFFLE segmentation benchmark. We also report performance on the standard Structured3D density maps, followed by an analysis of  performance trade-off between our method and RoomFormer arising from the introduction of semantic labels. Then, we report performance of noisy density maps and speed comparison against baselines. Finally, we present experiments on VLM-based refinement and a downstream application to 3D floorplan reconstruction. Additional qualitative results on Structured3D-B, CubiCasa5K, and \change{Raster2Graph} are provided in \cref{fig:s3d_more}, \cref{fig:cubi_more}, and \cref{fig:raster2graph_more}, along with qualitative results of zero-shot WAFFLE results in \cref{fig:waffle_more} and VLM-based refinement in \cref{fig:refinement}. For additional visualizations, please refer to our accompanying interactive tool.}

\subsection{Full performance comparison.}
\label{subsec:full_comp}
\begin{table*}[t]
\centering
\small
\caption{Quantitative evaluation  on the \emph{Structured3D-B} test set~\cite{zheng2020structured3d},  where the input image is a binary floorplan image. Best results are in bold. }
\label{tab:s3d_bw}
\resizebox{\linewidth}{!}{
\begin{tabular}{lccccccccc|cccccc}
\toprule
\multirow{2}{*}{Method} & \multicolumn{3}{c}{Room} & \multicolumn{3}{c}{Corner} & \multicolumn{3}{c|}{Angle} & \multicolumn{3}{c}{Room Semantic} & \multicolumn{3}{c}{Window \& Door}\\
\cmidrule(lr){2-4} \cmidrule(lr){5-7} \cmidrule(lr){8-10} \cmidrule(lr){11-13} \cmidrule(lr){14-16}
& Prec. & Rec. & F1 & Prec. & Rec. & F1 & Prec. & Rec. & F1 & Prec. & Rec. & F1 & Prec. & Rec. & F1 \\
\midrule
HEAT & 95.3 & 94.1 & 94.7 &	81.8 &	87.4 & 84.5 & 77.0 & 82.3 & 79.6 &-&-&-&-&-&-  \\
PolyRoom & 99.4 &	98.5 & 98.9 & \textbf{99.0} & 93.1 & 96.0 & \textbf{94.7} & 89.3 & 91.9 &-&-&-&-&-&- \\
FRI-Net & 97.5 & 95.4 & 96.5 & 88.5 & 82.6 & 85.4 & 86.2 & 80.5 & 83.3 & - & - & - & - & - & - \\

RoomFormer & 95.8 & 94.4 & 95.1 & 93.0 & 90.5 & 91.7 & 84.4 & 82.1 & 83.2 & 74.7 & 73.8 & 74.2 & 95.0 & 93.1 & 94.1 \\
\textbf{Ours} & \textbf{99.6} & \textbf{99.7} & \textbf{99.6} & \underline{98.9} & \textbf{97.7} & \textbf{98.3} & \underline{93.3} & \textbf{92.2} & \textbf{92.7} & \textbf{76.9} & \textbf{76.9} & \textbf{76.9} & \textbf{98.5} & \textbf{98.5} & \textbf{98.5} \\

\bottomrule
\end{tabular}
}

\end{table*}

\begin{table*}[t]
\centering
\small
\caption{Quantitative evaluation on the \emph{CubiCasa5K} test set~\cite{kalervo2019cubicasa5k}.}
\label{tab:cc5k}
\resizebox{\linewidth}{!}{
\begin{tabular}{lccccccccc|cccccc}
\toprule
\multirow{2}{*}{Method} & \multicolumn{3}{c}{Room} & \multicolumn{3}{c}{Corner} & \multicolumn{3}{c|}{Angle} & \multicolumn{3}{c}{Room Semantic} & \multicolumn{3}{c}{Window \& Door} \\
\cmidrule(lr){2-4} \cmidrule(lr){5-7} \cmidrule(lr){8-10}  \cmidrule(lr){11-13} \cmidrule(lr){14-16} 
& Prec. & Rec. & F1 & Prec. & Rec. & F1 & Prec. & Rec. & F1 & Prec. & Rec. & F1 & Prec. & Rec. & F1 \\
\midrule
HEAT & 79.9 & 76.6 & 78.2 & 56.2 & 51.4 & 53.7 & 33.8 & 31.0 & 32.3 & - & - & - & - & - & - \\
FRI-Net & 82.1 & 72.7 &	77.1 & \textbf{69.2} & 40.1 & 50.8 & \textbf{51.8} & 30.0 & \textbf{38.0} & - & - & - & - & - & - \\
RoomFormer & 84.7 & 82.3 & 83.5 & 58.1 & 53.1 & 55.5 & 35.7 & 32.6 & 34.1 & 63.8 & 62.3 & 63.0 & \textbf{80.8} & 76.3 & \textbf{78.5} \\
\textbf{Ours} & \textbf{89.3} & \textbf{88.0} & \textbf{88.7} & \underline{61.0} & \textbf{57.8} & \textbf{59.4} & \underline{38.4} & \textbf{36.4} & \underline{37.4} & \textbf{64.4} & \textbf{63.2} & \textbf{63.8} & 78.9 & \textbf{76.7} & 77.8 \\

\bottomrule
\end{tabular}
}

\end{table*}

\begin{table*}[t]
\centering
\small
\caption{Quantitative evaluation on the \emph{Raster2Graph} test set~\cite{hu2024r2g}.}
\label{tab:r2g}
\resizebox{\linewidth}{!}{
\begin{tabular}{lccccccccc|ccc}
\toprule
\multirow{2}{*}{Method} & \multicolumn{3}{c}{Room} & \multicolumn{3}{c}{Corner} & \multicolumn{3}{c|}{Angle} & \multicolumn{3}{c}{Room Semantic} \\
\cmidrule(lr){2-4} \cmidrule(lr){5-7} \cmidrule(lr){8-10}  \cmidrule(lr){11-13}
& Prec. & Rec. & F1 & Prec. & Rec. & F1 & Prec. & Rec. & F1 & Prec. & Rec. & F1 \\
\midrule
HEAT & \textbf{98.0} & 93.9 & 95.9	& 81.2	& 78.2	& 79.7 & 51.9 & 49.9 & 50.9 & - & - & - \\
FRI-Net & 94.9 & 88.4 & 91.5 & \textbf{86.6} & 62.1 & 72.3 & 63.2 & 45.3 & 52.8 & - & - & - \\
RoomFormer & 92.0 & 91.8 & 91.9 & 74.8 & 74.3 & 74.5 & 51.2 & 50.9 & 51.1 & 79.6 & 79.5 & 79.5 \\
Raster2Graph & 97.1 & 93.0 & 95.0 & 79.9 & 76.8 & 78.3 & \textbf{68.6} & 66.0 & \textbf{67.3} & 85.2 & 81.7 & 83.4 \\
\textbf{Ours} & \underline{97.2} & \textbf{96.8} & \textbf{97.0} & 80.4 & \textbf{80.1} & \textbf{80.3} & \underline{66.7} & \textbf{66.5} & \underline{66.6} & \textbf{85.3} & \textbf{84.9} & \textbf{85.1} \\

\bottomrule
\end{tabular}
}

\end{table*}

\begin{table*}[t]
\small
\centering
\caption{Quantitative evaluation  on the \emph{Structured3D} test set~\cite{zheng2020structured3d}, where the input is a density map generated from top-view projection of the 3D point cloud. In the bottom rows, we report performance using PD~\cite{chen2023polydiffuse}, a recent refinement method. As illustrated above, our method demonstrates competitive performance on this benchmark, and is compatible with existing refinement methods, which enable further performance gains.} 
\label{tab:s3d_point}
\begin{tabular}{llllllllll}

\toprule

\multirow{2}{*}{Method} & \multicolumn{3}{c}{Room} & \multicolumn{3}{c}{Corner} & \multicolumn{3}{c}{Angle} \\

\cmidrule(lr){2-4} \cmidrule(lr){5-7} \cmidrule(lr){8-10}

& Prec. & Rec. & F1 & Prec. & Rec. & F1 & Prec. & Rec. & F1 \\

\midrule

MonteFloor~\cite{stekovic2021montefloor} & 95.6 & 94.4 & 95.0 & 88.5 & 77.2 & 82.5 & 86.3 & 75.4 & 80.5 \\

HEAT~\cite{chen2022heat} & 96.9 & 94.0 & 95.4 & 81.7 & 83.2 & 82.5 & 77.6 & 79.0 & 78.3 \\

PolyRoom~\cite{liu2024polyroom} & 98.9 & 97.7 & 98.3 & \textbf{94.6} & 86.1 & \textbf{90.2} & 89.3 & 81.4 & 85.2 \\

FRI-Net~\cite{xu2024fri}  & \textbf{99.5} & \textbf{98.7} & \textbf{99.1} & 90.8 & 84.9 & 87.8 & \textbf{89.6} & \textbf{84.3} & \textbf{86.9} \\

RoomFormer~\cite{yue2023roomformer} & 97.9 & 96.9 & 97.5 & 89.4 & 85.5 & 87.4 & 83.2 & 79.7 & 81.4 \\

RoomFormer (w/ semantic) & 95.3{\scriptsize(-2.6)} & 93.5{\scriptsize(-3.4)} & 94.4{\scriptsize(-3.1)} & 85.7{\scriptsize(-3.7)} & 81.8{\scriptsize(-3.7)} & 83.7{\scriptsize(-3.7)} & 78.0{\scriptsize(-5.2)} & 74.5{\scriptsize(-5.2)} & 76.2{\scriptsize(-5.2)} \\

\textbf{Ours} & 99.0 & 98.4 & 98.7 & 92.0 & 87.1 & 89.4 & 84.7 & 80.3 & 82.5 \\

 \textbf{Ours} (w/ semantic) & \underline{99.1} & \underline{98.6} & \underline{98.8} & \underline{92.1} & \textbf{88.1} & \underline{90.0} & 86.1 & \underline{82.5} & 84.2 \\

\midrule

FRI-Net + PD \cite{xu2024fri} & \textbf{99.6} & 98.6 & 99.1 & \textbf{94.2} & 88.2 & 91.1 & \textbf{91.9} & 86.7 & \textbf{89.2} \\

RoomFormer + PD \cite{chen2023polydiffuse} & 98.7 & 98.1 & 98.4 & 92.8 & \textbf{89.3} & 91.0 & 90.8 & \textbf{87.4} & 89.1 \\

Ours + PD & \underline{99.4} & \textbf{98.9} & \textbf{99.2} & \underline{93.2} & \underline{89.2} & \textbf{91.2} & \underline{91.0} & \underline{87.2} & 89.0 \\

\bottomrule

\end{tabular}

\end{table*}

Detailed results of Structured3D-B, CubiCasa5K, Raster2Graph are shown in \Cref{tab:s3d_bw}, \Cref{tab:cc5k}, \Cref{tab:r2g}, respectively. Overall, we achieve superior geometric performance in two key metrics, Room and Corner, while demonstrating strong semantic floorplan reconstruction results. This is attributed to our labeled polygon representation and our token-wise classification loss.

\subsection{Zero-shot performance on unseen WAFFLE.} Table~\ref{tab:waffle} presents the cross-evaluation results for interior segmentation on the WAFFLE test set, using a model trained on CubiCasa5k, without exposure to any WAFFLE samples. Our method achieves the best overall performance, with the highest IoU (73.9), precision (81.6), and recall (88.6). In comparison, RoomFormer falls behind in precision (65.7) and IoU (60.5), indicating less reliable predictions. The pretrained model, trained for segmentation on the CubiCasa5K dataset, shows the weakest performance—particularly in recall (52.1) and IoU (46.1)—highlighting its limited generalization capabilities. These results demonstrate the superior output quality and generalization ability of our method on complex and unseen floorplan samples.

\begin{table}[t]
\centering
\small
\caption{Semantic scores on Structured3D test set~\cite{zheng2020structured3d} where the input image is a point-cloud density map.}
\label{tab:s3d_org_sem}
\resizebox{\linewidth}{!}{
\begin{tabular}{lcccccc}
\toprule
\multirow{2}{*}{Method} & \multicolumn{3}{c}{Room Semantic} & \multicolumn{3}{c}{Window \& Door} \\
\cmidrule(lr){2-4} \cmidrule(lr){5-7}
& Prec. & Rec. & F1 & Prec. & Rec. & F1 \\
\midrule
RoomFormer & 71.5 & 70.0 & 70.7 & \textbf{83.4} & \textbf{79.0} & \textbf{81.1} \\
\textbf{Ours} & \textbf{76.8} & \textbf{76.5} & \textbf{76.7}  & 78.6 & 77.4 & 78.0 \\
\bottomrule
\end{tabular}
}

\end{table}

\subsection{\change{Performance on Structure3D-Density maps}}
\label{sec:density_maps}

We conduct a comparison on the standard Structured3D benchmark, providing our model with density map inputs for both training and testing.
As illustrated in \cref{tab:s3d_point}, our method generally outperforms existing baselines on key geometric metrics such as Room and Angle. Although FRI-Net achieves competitive performance with our method when using density maps, performance on image inputs is generally lower (see \cref{tab:s3d_point}).
We hypothesize that FRI-Net's reliance on disentangled representations of raw line primitives makes it less robust to the diverse structural and appearance variations present in RGB floorplans compared to the homogeneous nature of density maps.
We also report performance using PD~\cite{chen2023polydiffuse}--a polygon refinement approach. Our method achieves state-of-the-art performance, demonstrating its compatibility with advanced post-processing techniques.

\minihead{Semantic performance.} In \cref{tab:s3d_org_sem}, we provide the comparison on the semantic scores between our method and RoomFormer on 3D scan inputs. As seen, our method offers superior performance in terms of the room semantic criteria while obtaining slightly lower measures for window and door. Notably, when semantic room types are included, RoomFormer exhibits a significant performance drop of 2–5 points (see \cref{tab:s3d_point}). By contrast, our model effectively captures both spatial and semantic attributes without compromising performance. This further demonstrates the efficacy of our polygon representation.

\change{
\minihead{Robustness to noisy density maps.}
To illustrate robustness in this setting, we conduct an experiment that adds noise to the density maps via a masking scheme.
Specifically, we applied a 20\% dropout rate to randomly mask out projected density signals for both training and test samples.
Empirically, our method demonstrates strong robustness to these noisy inputs, with only a 1-point drop in RoomF1 (from 98.7), compared to a 2.6-point drop for RoomFormer (from 97.5). A similar trend holds for the Corner metric, where we observe a decline of approximately 1.7 points for our method versus 2.1 points for RoomFormer. For the Angle metric, both methods exhibit comparable degradation.}

\subsection{Runtime comparison}
\begin{table}[!ht]
    \centering
    \small
    \caption{Speed comparison. All are computed on a single A6000 GPU. Training time is reported on Raster2Graph dataset.}
    \label{tab:speed}
    \resizebox{\linewidth}{!}{
    \begin{tabular}{lccc}
    \toprule
    Method                  & Sampling time (bs=1) & Training Throughput (images/s) & Training (Epochs/Time) \\ 
    \midrule
    HEAT & 0.09s & 39 & 400/1.2d \\
    FRI-Net & 0.56s & 53 & 1800/3.8d \\
    RoomFormer & 0.04s & 24 & 800/7.6d \\
    Raster2Graph & 0.57s & 34 & 800/3.1d \\
    Ours & 0.52s & 63 & 1400/2.9d \\
    \bottomrule
    \end{tabular}
    }
    
\end{table}

We report sampling time, training throughput, and training time in \Cref{tab:speed}. As seen, our method achieves comparable inference speed to Raster2Graph (0.52s vs. 0.57s) though is slower than the single-pass RoomFormer (0.04s). Despite inference time trade-off, our approach delivers the highest training throughput (63 images/s vs. 24 for RoomFormer and 34 for Raster2Graph), enabling faster training in low-resource settings.

\subsection{\change{VLM-based refinement}}
\label{subsec:VLM_refinement}

\begin{figure*}[t]
\centering
\includegraphics[width=0.95\linewidth]{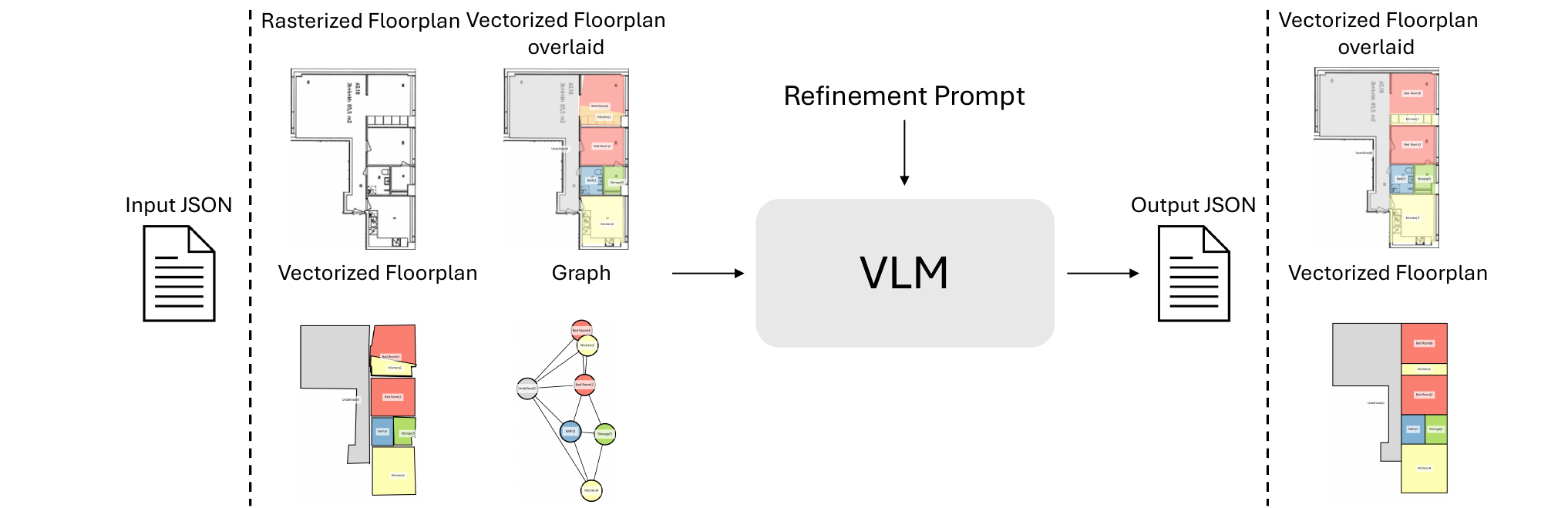}
\vspace{-8pt}
\caption{\change{VLM-based floorplan refinement. Given an input JSON that specifies the vectorized floorplan predicted by our method, we refine this reconstructed floorplan using a VLM that is additionally provided with the rasterized floorplan, the vectorized floorplan overlaid on the rasterized image, the vectorized floorplan alone, and the adjacency graph. Users can specify geometric constraints in the refinement prompt (detailed in Fig.~\ref{lst:prompt_refinement}); the VLM then outputs the refined JSON.}}
\label{fig:vlm_refinement}
\end{figure*}

\begin{figure}[t]
\centering
\includegraphics[width=0.95\linewidth]{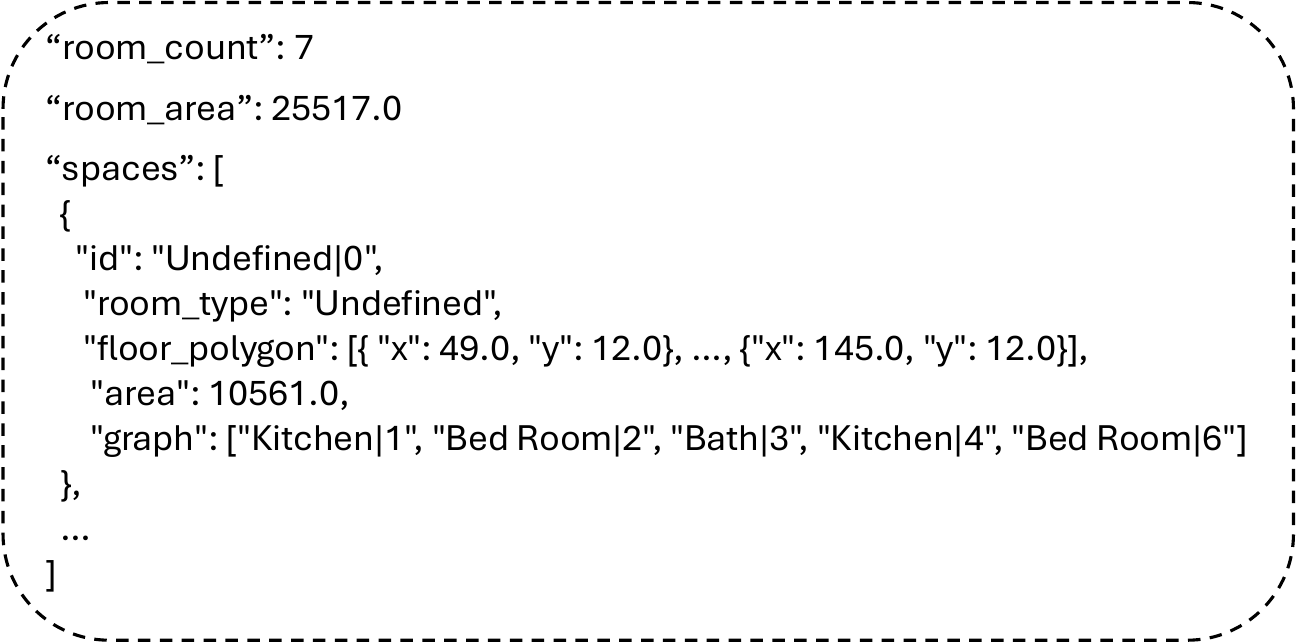}
\vspace{-8pt}
\caption{\change{Structured JSON format}, used for presenting the input JSON file in \cref{subsec:VLM_refinement}.}
\label{fig:json}
\end{figure}

\begin{figure*}[t]
\centering

\rotatebox{90}{\whitetxt{sssssssss}Input}
\jsubfig{
\includegraphics[height=2.505cm,angle=90,]{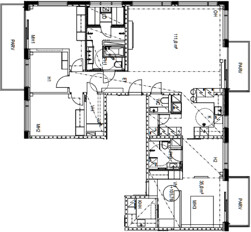}
\includegraphics[height=2.505cm,angle=0]{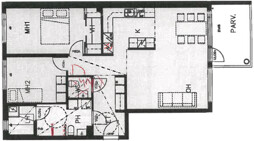}
\includegraphics[height=2.505cm,angle=0]{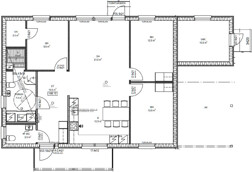}
\includegraphics[height=2.505cm,angle=0]{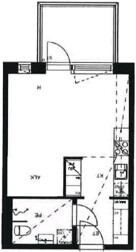}
\includegraphics[height=2.505cm,angle=0]{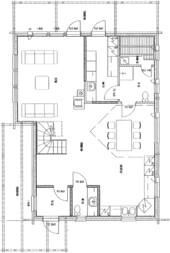}
\includegraphics[height=2.505cm,angle=0]{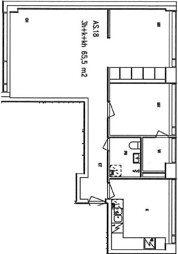}
}{}\\

\rotatebox{90}{\whitetxt{sssssssssss}GT}
\jsubfig{
\includegraphics[height=2.505cm,angle=90]{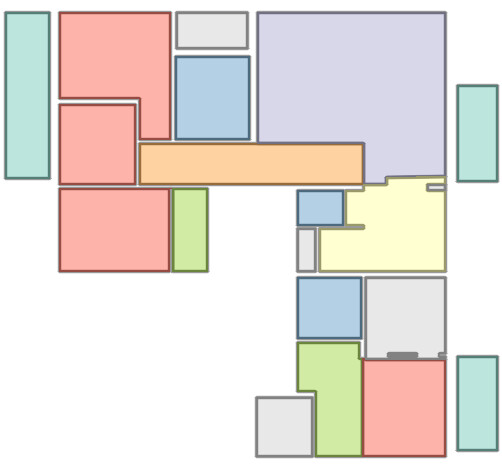}
\includegraphics[height=2.505cm,angle=0]{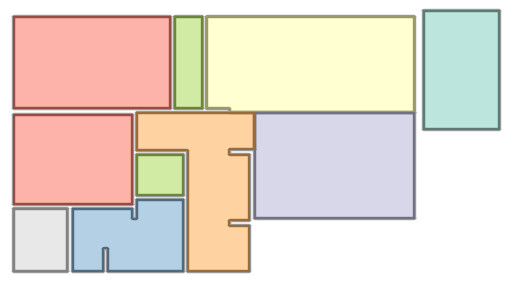}
\includegraphics[height=2.505cm,angle=0]{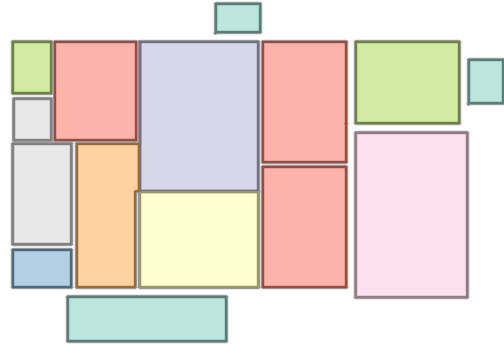}
\includegraphics[height=2.505cm,angle=0]{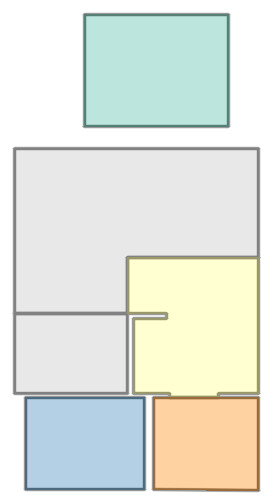}
\includegraphics[height=2.505cm,angle=0]{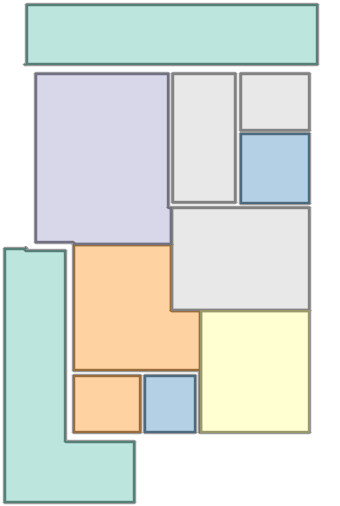}
\includegraphics[height=2.505cm,angle=0]{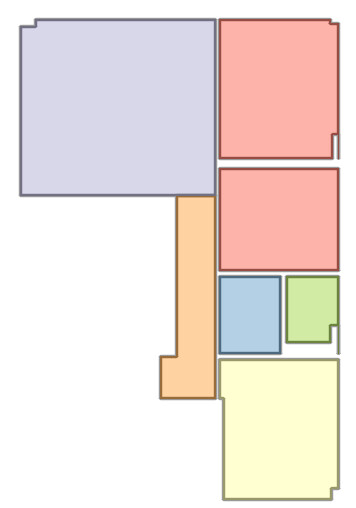}
}{}\\

\rotatebox{90}{\whitetxt{s}Before Refinement}
\jsubfig{
\includegraphics[height=2.505cm,angle=90]{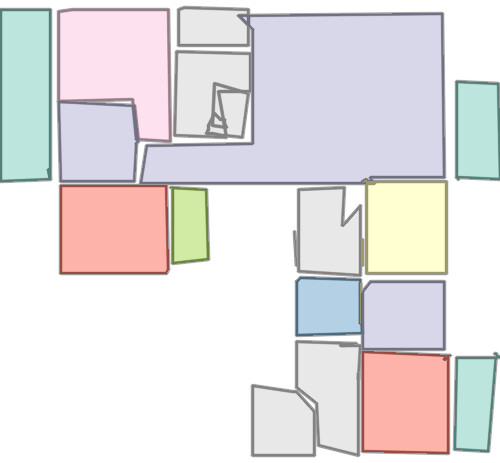}
\includegraphics[height=2.505cm,angle=0]{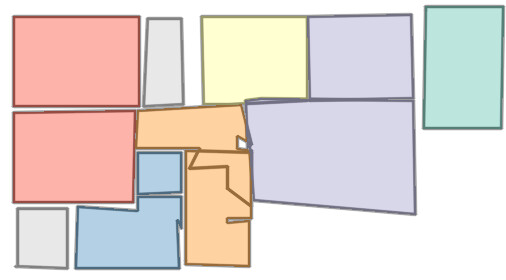}
\includegraphics[height=2.505cm,angle=0]{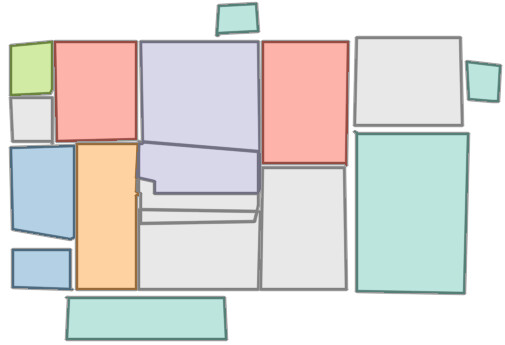}
\includegraphics[height=2.505cm,angle=0]{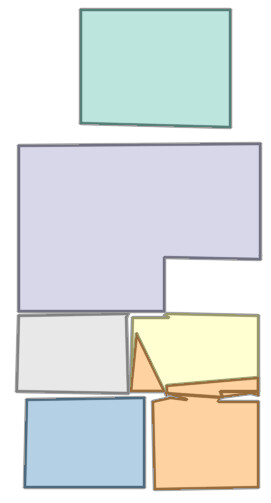}
\includegraphics[height=2.505cm,angle=0]{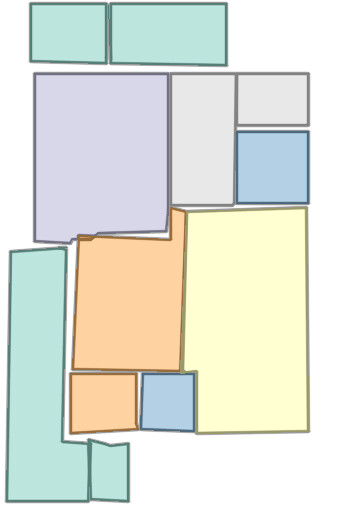}
\includegraphics[height=2.505cm,angle=0]{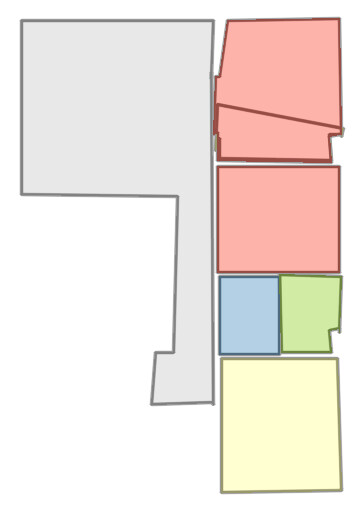}
}{}\\

\rotatebox{90}{\whitetxt{s}After refinement}
\jsubfig{
\includegraphics[height=2.505cm,angle=90]{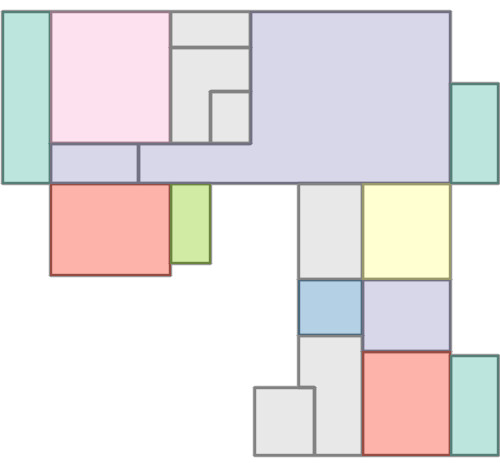}
\includegraphics[height=2.505cm,angle=0]{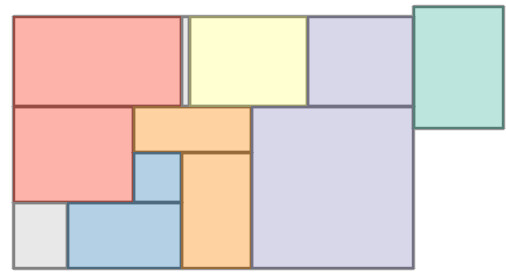}
\includegraphics[height=2.505cm,angle=0]{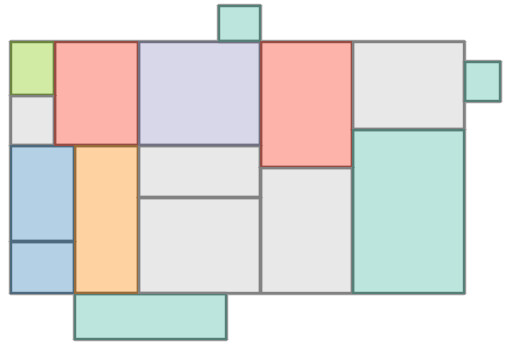}
\includegraphics[height=2.505cm,angle=0]{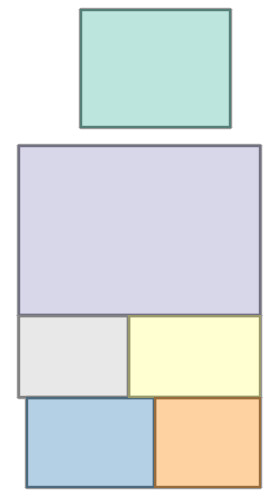}
\includegraphics[height=2.505cm,angle=0]{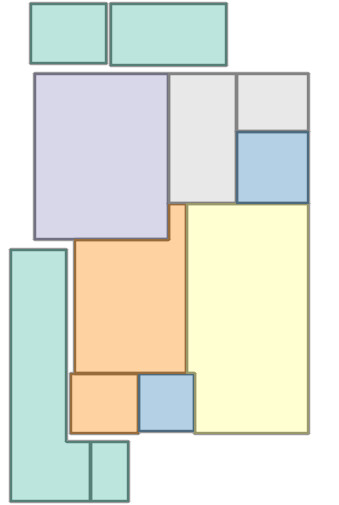}
\includegraphics[height=2.505cm,angle=0]{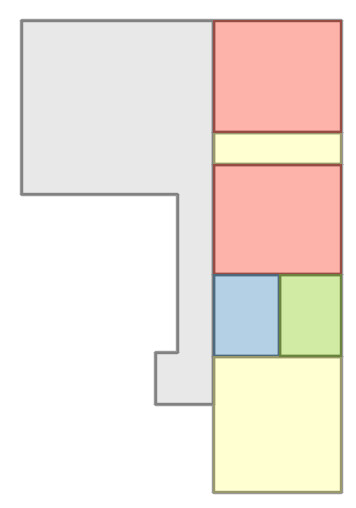}
}{}\\

\vspace{4pt}
\begin{tikzpicture}
\fill[outdoor, opacity=0.59] (0,1) rectangle (1.4,0.5);
\node[black] at (0.7,0.75) {\small\textbf{Outdoor}};
\fill[kitchen, opacity=0.59] (1.8,1) rectangle (3.2,0.5);
\node[black] at (2.5,0.75) {\small\textbf{Kitchen}};
\fill[livingroom, opacity=0.59] (3.6,1) rectangle (5.0,0.5);
\node[black] at (4.3,0.75) {\small\textbf{Living room}};
\fill[bedroom, opacity=0.59] (5.4,1) rectangle (6.8,0.5);
\node[black] at (6.1,0.75) {\small\textbf{Bed room}};
\fill[bath, opacity=0.59] (7.2,1) rectangle (8.6,0.5);
\node[black] at (7.9,0.75) {\small\textbf{Bath}};
\fill[entry, opacity=0.59] (9.0,1) rectangle (10.4,0.5);
\node[black] at (9.7,0.75) {\small\textbf{Entry}};
\fill[storage, opacity=0.59] (10.8,1) rectangle (12.2,0.5);
\node[black] at (11.5,0.75) {\small\textbf{Storage}};
\fill[garage, opacity=0.59] (12.6,1) rectangle (14.0,0.5);
\node[black] at (13.3,0.75) {\small\textbf{Garage}};
\fill[undefined, opacity=0.59] (14.4,1) rectangle (15.8,0.5);
\node[black] at (15.1,0.75) {\small\textbf{Undefined}};
\end{tikzpicture}

\vspace{-4pt}
\caption{\emph{Raster2Seq}'s refinement results on CubiCasa5K.}
\label{fig:refinement}
\end{figure*}

\change{While our method yields plausible floorplan reconstruction results that outperforms existing works over various metrics, our technique does not directly enforce geometric constraints within the framework. This is particularly evident over the CubiCasa5K dataset where the samples are often noisy and contains overlapping room annotations, causing some predicted outputs to exhibit similar artifacts (see \cref{fig:refinement}). Therefore, we conduct experiments on enforcing geometric constraints via a VLM-based vectorization refinement (see \cref{fig:vlm_refinement}), demonstrating that our semantic representation is also useful for post-process refinement schemes. 
Specifically, we provide a VLM (Gemini 2.5 Pro) with our labeled polygons, along with the rasterized input, a visualization of the vectorized floorplan both standalone and overlaid on the rasterized image, and an adjacency graph derived directly from our prediction, which provides relations between different room instances. The labeled polygons are represented in a structured JSON format following \cite{luo2024dstruct2design}, which consolidates each room's ID, type, polygon coordinates, area, and connectivity into a structured representation (\cref{fig:json}). This rich, multi-modal context enables the VLM to reason about geometric consistency and refine the vectorized output accordingly. To specify geometric constraints, we design a text prompt (Fig. \ref{lst:prompt_refinement}) that explicitly imposes two specific constraints: adjacent rooms must share edges without gaps or intersections, and all edges must remain orthogonal so that vectors snap precisely to walls.
}

\change{
For this additional experiment, we evaluate on a subset of 30 randomly selected samples from the CubiCasa5K test set (400 samples total).
As can be observed in \cref{tab:refinement}, we find that our method achieves significant geometric improvements (e.g., Corner and Angle scores increased from 54.0 to 59.0 and 33.0 to 45.1, respectively).
We also validate this improvement qualitatively in \cref{fig:refinement}. The vectorized floorplan exhibits remarkably tight gaps between adjacent rooms. Importantly, the overlapping rooms seen in some original examples have been fully separated, confirming that the specified geometric constraints are properly enforced.
}

\begin{table}[t]
\centering
\footnotesize

\setlength{\tabcolsep}{3pt}
\caption{\change{Refinement results on CubiCasa5K subset.}}
\label{tab:refinement}
\resizebox{0.48\textwidth}{!}{
\begin{tabular}{lllll}
\toprule
Method & Room & Corner & Angle & RoomSem \\
\midrule
Before Refinement & 83.7 & 54.0 & 33.0 & 61.1 \\
After Refinement & 81.7 & 59.0 & 45.1 & 60.3 \\

\bottomrule
\end{tabular}%
}

\vspace{-10pt}
\end{table}

\begin{figure*}[t]
\noindent\rule{\textwidth}{0.8pt}
\begin{lstlisting}[style=promptstyle]
You are a specialized Architectural Geometry AI. Your expertise lies in topological refinement: transforming JSON specifications and visual raster data-including bubble diagrams and vectorized drafts-into precise, non-overlapping floorplans by generating optimized $xy$ coordinates.

Goal: Produce an optimal arrangement of floorplan elements that maximizes area utilization. The algorithm must prioritize the spatial logic of the Floorplan Raster while using the Draft JSON only as a topological and proportional guide.

Inputs:
- JSON Specification: Contains preliminary room dimensions, labels, and connectivity requirements. Note: These numerical values (area, height, width) are derived from a rough draft and serve only as a proportional guide. They should be refined to match the visual scale and alignment of the Original Floorplan Raster.
- Original Floorplan Raster (Image A): The architectural blueprint for alignment and scale.
- Vectorized floorplan rendering (Image B): Shows the spatial arrangement and room IDs where each floorplan object is colored with type|id labels.
- Vectorized floorplan rendering overlaid (Image C): Shows the spatial arrangement and room IDs overlaid ontop of original floorplan raster.
- Adjacency Graph (Image D): Defines the topological connections.

Output: JSON file containing refined polygons.
The JSON object must contain 'output' key storing these attributes:
- room_count': the total number of room entries
- 'spaces': a list of refined rooms. Each room entry must include:
- 'id': formatted as `<room_type>|<unique_index>` (e.g. "bedroom|2" or "interior_door|0")
- 'room_type': the room type (e.g. "living_room", "kitchen", etc.)
- 'area' in square meters (all positive numbers)
- 'floor_polygon': an ordered list of '{x: , y:}'
vertices defining a polygon after refinement
- 'graph': store a list of adjacent space object 'id' (e.g. ["Bed Room|1", "Entry|2"])

Spatial Reference System:
- Coordinate Space: All vertex calculations must be performed within a fixed $[0, 256]$ coordinate system.
- Origin: $(0,0)$ represents the top-left corner of the Original Floorplan Raster.
- Polygons in JSON are ordered by couter-clockwise direction.

Refinement Constraints:
- Contextual Overlaps: While polygons should generally avoid unwarranted intersections, minor overlaps are permitted if they are supported by the Original Floorplan Raster (Image A) (e.g., representing shared wall thicknesses, nested structural layouts). Use the raster as the ultimate judge to determine whether an overlap is a draft error to be eliminated, or a valid structural representation.
- Watertight Adjacency: Rooms that share a boundary (as indicated by the Adjacency Graph) must utilize the exact same coordinate values for the shared edge to ensure no gaps at the joints.
- Identity Preservation: Every id (e.g., bedroom|0, entry|2) must be preserved and accurately repositioned.
- Scale Fidelity: The final area should be approximated as $\text{width} \times \text{height}$ to verify it matches the Original Raster's proportions. 
- Truth Hierarchy: In the event of a conflict between the Draft JSON and the Floorplan Raster, the Floorplan Raster is the primary source of truth for wall placement and room proportions.
- Proportional Scaling: Use the Draft JSON's height and width only as a reference for relative aspect ratios. The final coordinates must be recalculated to ensure the resulting area aligns with the visual footprint shown in the Raster.
- Coordinate Precision: Use floating-point coordinates with up to two decimal places for maximum precision.
- Attribute Preservation: for space object with attributes remained unchanged (i.e. 'graph'), directly copy them to the output object.
- Manhattan Style: All edges must be axis-aligned (horizontal or vertical) unless the Original Raster explicitly shows a non-90 angle.
- Order Compliance: Refinement strategy should follow the order of space objects as defined in 'id' attribute (e.g. 'bedroom|0', 'kitchen|1').

Here, these are detailed producure of floorplan refinement (Mandatory).

1. Problem analysis:
 
Examine the provided JSON and visual data (Image A-D). Identify specific geometric failures, such as:
- Unjustified overlapping polygons (e.g., where Living Room|0 bleeds into Entry|2 without structural justification in the Raster).
- Rooms that are disconnected despite being adjacent in the graph
- Edges that deviate from perfectly horizontal/vertical and need to be snapped to the Manhattan style.
- Scale Mismatches: Compare the draft polygons in the Image B & C to the corresponding spaces in the Floorplan Raster (Image A). Note where the draft's area or aspect ratio significantly deviates from the architectural raster.

2. Reasoning Plan
Outline the coordinate adjustments needed to produce the final answer.

3. Step-by-Step Execution
Process the refinement room-by-room. Provide a trace of how you are calculating the new vertex coordinates based on the $x,y$ grid to ensure alignment with visual footprint.
Show area check: explicitly calculate the room area (e.g. width \times height) to validate that the refined polygon proportions relatively align with the visual footprint of the Original Floorplan Raster.

4. Final Answer
Provide the refined JSON object, strictly adhering to the defined JSON format.

\boxed{<JSON content>}

Note:
- You MUST explicitly write out your reasoning, following the procedure above.
- The final JSON <JSON content> must be placed inside a \boxed{} block.
\end{lstlisting}
\noindent\rule{\textwidth}{0.8pt}
\caption{\change{Prompt used for VLM-based floorplan refinement (\cref{subsec:VLM_refinement}) to enforce geometric constraints.}}
\label{lst:prompt_refinement}
\end{figure*}

\change{Importantly, our semantic predictions — which most prior methods lack — serve as critical room identity cues, enabling the VLM to differentiate room instances and recognize adjacency relationships for more targeted refinement. Removing the semantic labels from our representation leads to approximately a 3-point drop in both Corner and Angle metrics, underscoring their importance for effective VLM-based refinement.}

\begin{figure}[t]
    \centering
    \rotatebox{90}{\tiny \hspace{15pt}Output \hspace{20pt}  Control \hspace{20pt} Floorplan \hspace{25pt}Input Image}
    \includegraphics[width=0.9\linewidth]{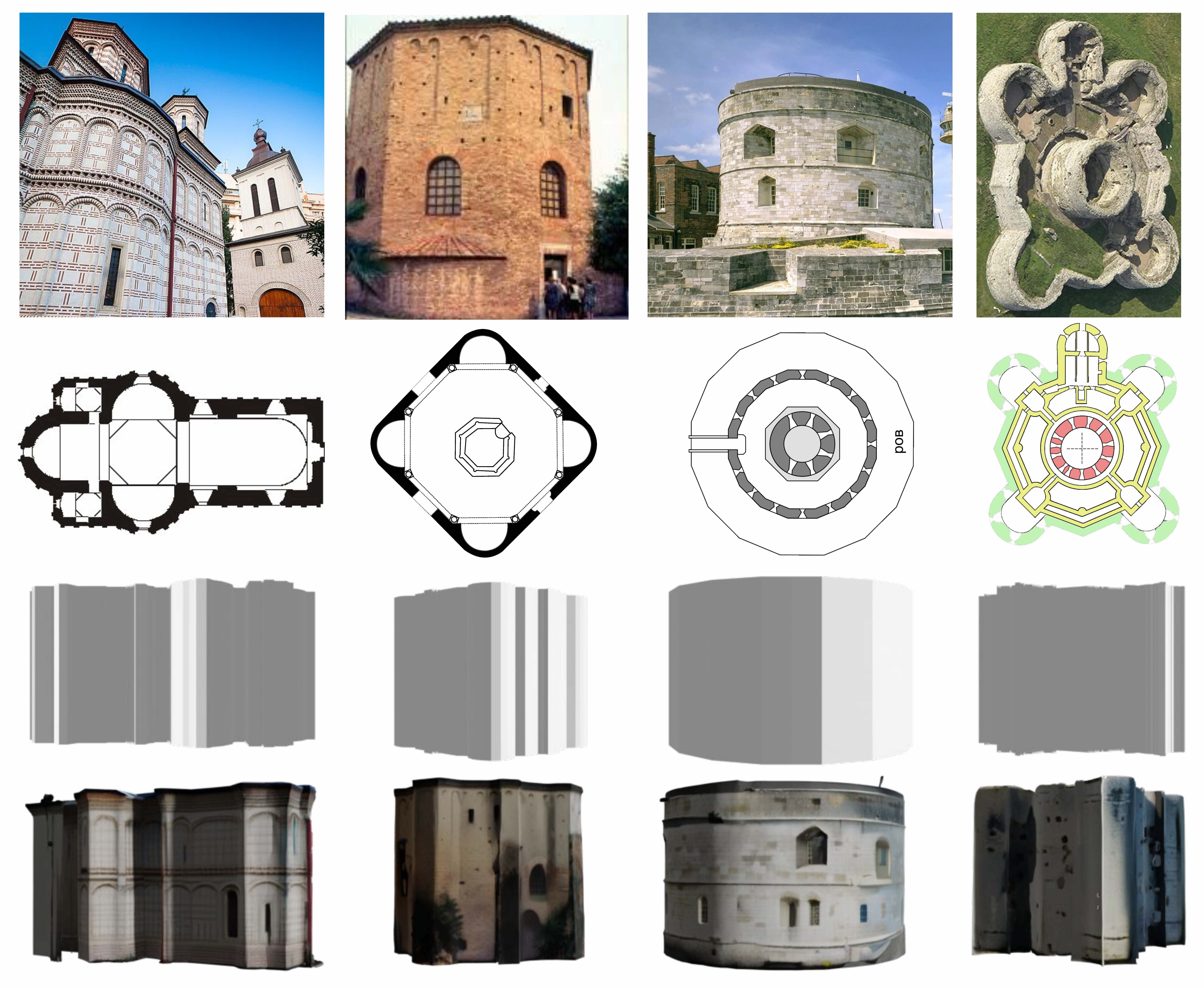}
    \caption{Downstream application: controllable 3D generation from vectorized floorplans. A 3D volume derived from the input floorplan (second row) serves as spatial control (third row) and, together with a conditioning image (top row), guides the generation of 3D scenes (bottom row).}
    \label{fig:generation}
\end{figure}

\subsection{Downstream application}
\label{subsec:trellis}
Vectorizing rasterized floorplans enables a range of downstream computational tasks that are difficult or impossible to perform directly in pixel space. As a concrete demonstration, we showcase its potential for controllable 3D scene generation. Given a vectorized floorplan, we construct a coarse 3D volume by extruding its boundary geometry along the vertical axis. This volume serves as explicit spatial guidance for a pretrained 3D generative model (\emph{i.e.}, TRELLIS~\cite{xiang2025structured}), using the test-time approach introduced in SpaceControl~\cite{fedele2025spacecontrol}. %
Figure \ref{fig:generation} shows several examples. These results illustrate that  floorplan vectorization provides a strong geometric prior, allowing the 3D generative model to faithfully reproduce complex architectural layouts from a single input RGB image, while maintaining global structural consistency.

\section{Additional Ablation Studies}
\label{sec:more_abl}

The experiments in this section are primarily conducted on Structured3D-B—the binary dataset—unless explicitly stated otherwise.

\minihead{Quantization resolution.} \Cref{tab:quant_res} presents an ablation study on the effect of coordinate quantization resolution—represented by the number of discretized bins (e.g., $16\times16$, $32\times32$)—on floorplan reconstruction performance. As shown, a 32×32 resolution yields the best overall performance, with the highest Room F1 (96.3) and Corner F1 (93.7), while maintaining competitive Angle F1 (82.6). Both coarser ($16\times16$) and finer ($64\times64$) quantizations result in reduced performance, suggesting that $32\times32$ offers the best trade-off between granularity and model performance. Given that the input coordinate values lie within the range [0, 256], using a fine-grained $64\times64$ quantization may introduce redundant precision and undesirable side effects.

\minihead{Random vs. Learnable anchors} As seen in \cref{tab:anchor}, using random initialized anchors barely bring any improvement compared to the baseline (which does not use anchors). By jointly training the anchors with model parameters, this give a significant boost to overall measurement, illustrating the importance of \emph{learnable} anchors in our framework. 

\begin{table}[t]
\centering
\small
\caption{Effect of Random and learnable anchors.}
\label{tab:anchor}
\begin{tabular}{lccccccccc}
\toprule
Anchor & Room F1 & Corner F1 & Angle F1 \\
\midrule
Baseline & 94.1 & 91.1 & 82.0 \\
Random & 94.4 &	90.8 &	81.4 \\
\rowcolor{pink!40} Learnable & 99.6 & 98.3 & 92.7 \\
\bottomrule
\end{tabular}

\end{table}

\minihead{Sequence length.} \Cref{tab:seq_length} examines the impact of input sequence length on floorplan generation performance, comparing sequence lengths ranging from 256 to 1024. The results clearly show that increasing the sequence length significantly improves reconstruction quality across all metrics, with gains of 3–10 points when moving from a length of 256. The highlighted row for length 512 corresponds to the best-performing configuration, indicating that it strikes a sweet spot for capturing structural and geometric details in floorplans effectively.

\begin{table}[t]
\centering
\small
\caption{Effect of coordinate loss coefficient on floorplan reconstruction performance.}
\label{tab:coord_coeff}
\begin{tabular}{lccccccccc}
\toprule
Coord. Coeff & Room F1 & Corner F1 & Angle F1 \\
\midrule
10 & 93.2 & 89.2 & 74.3 \\
\rowcolor{pink!40} 20 & 96.3 & 93.7 & 82.6 \\
40 & 92.0 & 87.6 & 74.0 \\
\bottomrule
\end{tabular}

\end{table}

\minihead{Coordinate coefficient.} \Cref{tab:coord_coeff} presents an ablation study on the coordinate loss coefficient. In this experiment, we fix the token loss coefficient at 1 to isolate and evaluate the impact of varying the coordinate loss weight. Markedly, setting the coordinate loss coefficient to 20 yields the best overall performance, with Room F1 at 96.3, Corner F1 at 93.7, and Angle F1 at 82.6. Lower (10) and higher (40) values of the coefficient lead to a noticeable drop in all metrics, suggesting that an appropriately balanced coordinate loss is crucial for accurate geometric prediction.

\minihead{One-stage training.} In \cref{tab:onestage}, we find that training the semantic model in one stage achieves comparable scores to the two-stage model (with the same training duration), with negligible decrease in semantic metrics (RoomSemanticF1: 76.9 vs. 76.1). This further demonstrates the flexibility of our model across different training schemes. Here, we opt for a two-stage solution for the optimal performance. Additionally, we conduct an ablation study on generation order, comparing right-to-left versus left-to-right ordering. The flipped version yields similar performance.
\begin{table}[t]
\small
\centering
\caption{Performance comparison between 2-stages VS 1-stage training where 2-stages training is a vanilla option, including pretraining (no semantic) and finetuning (with semantic). \textit{*RoomSem and WD denote "Room Semantic" and "Window \& Door", respectively.}}
\label{tab:onestage}
\resizebox{\linewidth}{!}{%
    \begin{tabular}{lccc|cc}
    \toprule
     & Room F1 & Corner F1 & Angle F1 & RoomSem F1 & WD F1 \\
    \toprule
    1stage & 99.6 & \textbf{98.4} & \textbf{93.2} & 76.1 & 98.4 \\
    2stages (vanilla) & \text{99.7} & 98.3 & 92.7 & \textbf{76.9} & \textbf{98.5} \\
    \bottomrule
    \end{tabular}
}

\end{table}

\minihead{Rasterization loss.}
We augmented our method with rasterization loss \cite{lazarow2022instance} as done in RoomFormer (see \cref{tab:raster_loss}). On Structured3D-B, CornerF1 and AngleF1 gets an improvement of 0.2 and 1.0, respectively. On CubiCasa5K, the gains are minimal, with CornerF1 improving from 59.4 to 59.8 and AngleF1 from 37.4 to 37.9. Since the gain is considerably marginal, we omit this loss for simplicity.
\begin{table}[t]
\small
\centering
\caption{Ablation on Rasterization loss.}
\label{tab:raster_loss}
\resizebox{\linewidth}{!}{%
    \begin{tabular}{lccc}
    \toprule
     & Room F1 & Corner F1 & Angle F1 \\
    \toprule
    Structured3D-B \\
    w/o & 99.6 & 98.3 & 92.7 \\
    w/ & 99.6 & \textbf{98.5} & \textbf{93.7} \\
    \midrule
    CubiCasa5K \\
    w/o & 88.7 & 59.4 & 37.4 \\
    w/ & 88.7 & 59.8 & 37.9 \\
    \bottomrule
    \end{tabular}
}
\end{table}

\clearpage

\end{document}